\documentclass[12pt]{article}
\usepackage{pdflscape}
\usepackage{enumerate}
\usepackage{amsmath}
\usepackage{amssymb}
\usepackage{epic}
\usepackage{epic,eepic}
\usepackage{fancyheadings}
\usepackage[pdftex]{graphicx}
\usepackage{graphics}
\usepackage{graphicx}
\usepackage{latexsym}
\usepackage{ulem}
\usepackage{xspace}
\usepackage{lscape}
\usepackage{yhmath}
\usepackage[top=30truemm, bottom=30truemm, left=25truemm, right=25truemm]{geometry}
\usepackage{hyperref}
\usepackage{mathtools}
\usepackage{rotating}
\allowdisplaybreaks

\usepackage{delarray}

\def\IntKer2{R(K)}

\newtheorem{Lemma}{Lemma}
\newtheorem{Theorem}{Theorem}
\newtheorem{Assumption}{Assumption}

\newtheorem{Remark}{Remark}

\begin{document}


\vspace{5mm}
\centerline{\LARGE{{\bf{Kernel Density Estimation by Stagewise Algorithm}}}}
\vspace{5mm}
\centerline{\LARGE{{\bf{with a Simple Dictionary}}}}
\vspace{5mm}
\begin{center}
Kiheiji NISHIDA
\footnote{Lecturer, General Education Center, Hyogo University of Health Sciences. \\Address: 1-3-6, Minatojima, Chuo-ku, Kobe, Hyogo, 650-8530, JAPAN. E-mail: kiheiji.nishida@gmail.com},
Kanta NAITO \footnote{Professor, Graduate School of Science, Chiba University. \\Address: 1-33, Yayoicho, Inage-ku, Chiba-shi, Chiba, 263-8522, JAPAN. E-mail: naito@math.s.chiba-u.ac.jp}
\end{center}
\vskip 10mm
\noindent {\bf{ABSTRACT}}

This study proposes multivariate kernel density estimation by stagewise minimization algorithm based on $U$-divergence and a simple dictionary. The dictionary consists of an appropriate scalar bandwidth matrix and a part of the original data. The resulting estimator brings us data-adaptive weighting parameters and bandwidth matrices, and realizes a sparse representation of kernel density estimation. We develop the non-asymptotic error bound of estimator obtained via the proposed stagewise minimization algorithm. It is confirmed from simulation studies that the proposed estimator performs competitive to or sometime better than other well-known density estimators.
\vskip 10mm

\section{Introduction}
Let $\mathbf{X}_{i}^{T}=(X_{i1} X_{i2} ... X_{id})$, $i=1,2,...,N$, be $d$-dimensional i.i.d. sample generated from the true density function $f(\mathbf{x})$ on $\mathbb{R}^{d}$. General representation of multivariate Kernel Density Estimator (KDE) is written as
\begin{eqnarray} \label{def.KDE}
\widehat{f}_{\mathbf{H}}(\mathbf{x}) = \sum_{i=1}^{N} \alpha_{i} K_{\mathbf{H}_{i}}(\mathbf{x}-\mathbf{X}_{i}),
\end{eqnarray}
where $\mathbf{H}_{i}$, $i = 1,2,...,N$, is a symmetric and positive definite $d$-dimensional bandwidth matrix used for the data $\mathbf{X}_{i}$, $K_{\mathbf{H}_{i}}(\mathbf{x}) = |\mathbf{H}_{i}|^{-\frac{1}{2}}K(\mathbf{H}_{i}^{-\frac{1}{2}} \mathbf{x})$ is a non-negative real valued bounded kernel function, and $\alpha_{i}$, $i = 1,2,...,N$, is the weighting parameters assigned for the data $\mathbf{X}_{i}$.

One approach to implement \eqref{def.KDE} is setting $\alpha_{1} = \alpha_{2} = ... = \alpha_{N} = 1/N$ and estimating $\mathbf{H}_{1} = \mathbf{H}_{2} = \cdots = \mathbf{H}_{N} = \mathbf{H}$ efficiently. This approach emphasizes finding efficient full-bandwidth matrix, instead of putting simple assumptions on weighting parameters. Duong and Hazelton (2003) propose the Direct Plug-In (DPI) bandwidth matrix while setting a bivariate full-bandwidth matrix. We denote the estimator using the DPI bandwidth matrix to be $\widehat{f}_{KDE}(\mathbf{x})$.

Another approach is the Redused Set Density Estimator (RSDE) in Girolami and He (2003), which firstly employs the scalar bandwidth matrix $\mathbf{H}_{1} = \cdots = \mathbf{H}_{N} = h^{2} \mathbf{I}_{d}$, where $\mathbf{I}_{d}$ is the $d$-dimensional identity matrix and the constant $h$ is determined by cross-validation. Second, the parameters $\alpha_{i}$, $i = 1,2,...,N$, are estimated to minimize Integrated Squared Error (ISE) under the constraint $\alpha_{i} \ge 0$, $i = 1,2,...,N$, $\sum_{i=1}^{N} \alpha_{i} = 1$. RSDE imposes simple assumptions on the bandwidth matrix, but requires more efforts in calculating the weighting parameters. RSDE also allows $\alpha_{i}=0$ for some $i$'s, realizing the sparse representation of kernel density estimation because those data points are not used in the estimation. We denote the estimator using RSDE to be $\widehat{f}_{RSDE}(\mathbf{x})$.

Other than these approaches, algorithm-based methods have also been developed such as projection pursuit density estimation (Friedman et al. 1984) and boosting by Ridgeway (2002). In relation to boosting, Klemel\"a (2007) developed a density estimation using stagewise algorithm and its non-asymptotic error bounds. Naito and Eguchi (2013) developed the stagewise algorithm under the setting of $U$-divergence. The stagewise algorithm requires a dictionary beforehand where the words consist of density functions; it starts by choosing a density function from the dictionary which minimizes the empirical loss, and proceeds in a stage-wise manner, adding new simple functions to the convex combination.

In this study, we consider applying the stagewise algorithm in Klemel\"a (2007) and Naito and Eguchi (2013) for the kernel density estimator in \eqref{def.KDE}. We randomly split an i.i.d. sample into the two disjoint sets, one to be used for the means of the kernel functions in the dictionary and the other for calculating the criterion function, and implement the stagewise algorithm. The outcome is expressed in the form of \eqref{def.KDE} and brings us the data-adaptive weighting parameters $\alpha_{i}$, while virtually realizing the data-adaptive bandwidth matrix through a variation in the bandwidths in the dictionary. It also chooses the data points of no use for the estimation, to obtain a sparse representation of kernel density estimation just like RSDE. We are especially interested in ascertaining whether or not our estimators can outperform its competitors, KDE and RSDE, in terms of estimation error and the degree of data condensation, while making the dictionary as simple as possible in terms of its bandwidth matrix structure.

The remainder of this paper is organized as follows. In Section~\ref{criterion}, we introduce the evaluation criterion for our proposed method, $U$-divergence. Section~\ref{Method} describes our proposed method. Section~\ref{Theoretical results} shows the theoretical results of our estimator, the non-asymptotic error bound of the estimator. We show the simulation results and real data example used in our method in Section~\ref{Applications}. The discussion and conclusions are presented in Section~\ref{Discussion}. In Appendices~A and B, we provide the proofs of the theorems for the non-asymptotic error bounds of the proposed estimator and its normalized version in Section~\ref{Theoretical results} respectively. In Appendices~C and D, we show details of the related results in Section~\ref{Theoretical results}.
\section{$U$-divergence} \label{criterion}

To compose the algorithm, we employ $U$-divergence defined as the distance between the fixed $f$ and any density function $g$ written as
\begin{eqnarray} \label{def.U.loss}
D_{U}(f, g) = \int_{\mathbb{R}^{d}} \biggl[ U(\xi(g(\mathbf{x}))) - U(\xi(f(\mathbf{x}))) - f(\mathbf{x}) \bigl \{ \xi(g(\mathbf{x})) - \xi(f(\mathbf{x})) \bigr\} \biggr] d\mathbf{x} \ge 0, \ \ 
\end{eqnarray}
where $U(t)$ is a strictly convex function on $[0, \infty)$, $u(t) = U^{'}(t) = dU(t)/dt$ and $\xi(t) = u^{-1}(t)$. The equality of \eqref{def.U.loss} holds if and only if $f=g$ except the set of measure zero. The non-negative property of $D_{U}(f, g)$ is explained by the convex property of $U(t)$. The functional form of $U$-divergence is similar to that of the Bregman divergence (see Bregman 1967; Zhang et al. 2009, 2010).

Extracting the part relating to $g$ from \eqref{def.U.loss}, we obtain
\begin{eqnarray} \label{true.e.loss}
{L}_{U}(g) = - \int_{\mathbb{R}^{d}} f(\mathbf{x}) \xi(g(\mathbf{x})) d\mathbf{x} + \int_{\mathbb{R}^{d}} U(\xi(g(\mathbf{x}))) d\mathbf{x}.
\end{eqnarray}
Replacing the first term in the right-hand side of \eqref{true.e.loss} with its empirical form, we obtain the empirical $U$-loss function written as
\begin{eqnarray} \label{e.loss}
\widehat{L}_{U}(g) = - \frac{1}{N} \sum_{i=1}^{N} \xi(g(\mathbf{X}_{i})) + \int_{\mathbb{R}^{d}} U(\xi(g(\mathbf{x}))) d\mathbf{x}.
\end{eqnarray}
Minimizing \eqref{e.loss} with respect to $g$ is equivalent to minimizing the empirical form of \eqref{def.U.loss} for a fixed $f$.

If we specify the convex function $U(t)$ to be the following $\beta$-power function with a tuning parameter $\beta$:
\begin{eqnarray} \label{beta.pow.func}
U_{\beta}(t) = \frac{1}{\beta + 1} (1 + \beta t)^{\frac{\beta + 1}{\beta}}, \ \ \ 0 < \beta \le 1, \nonumber
\end{eqnarray}
we obtain the resulting divergence function
\begin{eqnarray} \label{beta.pow.div}
D_{\beta}(f, g) = \frac{1}{\beta + 1} \int_{\mathbb{R}^{d}} \bigl \{ g(\mathbf{x})^{\beta +1} - f(\mathbf{x})^{\beta +1} \bigr \} d\mathbf{x} - \frac{1}{\beta} \int_{\mathbb{R}^{d}} f(\mathbf{x}) \bigl \{ g(\mathbf{x})^{\beta} - f(\mathbf{x})^{\beta} \bigr \} d\mathbf{x},
\end{eqnarray}
which is called the $\beta$-power divergence (see Basu et al. 1998; Minami and Eguchi 2002). We notice that the limit of $D_{\beta}(f, g)$ is equivalent to Kullbuck-Leibler (KL) divergence as $\beta$ goes to zero because $\lim_{\beta \to 0} U_{\beta}(t) = \exp(t)$. Alternatively, when $\beta = 1$, $D_{\beta}(f, g)$ is equivalent to $L_{2}$ norm. We also notice that the $\beta$-power divergence with $0 < \beta < 1$ exhibits robustness property, judging from the functional form of \eqref{beta.pow.div}; employing $U$-divergence enables us to consider a variety of density estimators in one function.
\section{Method} \label{Method}

Supposed that we have i.i.d. sample $\mathbf{X}_{i}^{T}=(X_{i1} X_{i2} ..., X_{id})$, $i=1,2,..., m+n = N$, generated from $f$. For this i.i.d. sample, we define $\mathbf{X}_{i}^{*} \equiv \mathbf{X}_{i}$, $i=1,2,...,m$, and use it for the dictionary. For the rest of the i.i.d. sample, we define $\mathbf{X}_{j}^{\tiny{+}} \equiv \mathbf{X}_{i}$, $i= m + j$, $j=1, 2,...,n$, and use it for the algorithm to calculate empirical loss. Let $B$ be a set of $d$-dimensional scalar bandwidth matrices $h_{j}{\bf{I}}_{d}$, $j=1,2,..., |B|$,
\begin{eqnarray} \label{Dictionary}
B &=& \Biggl\{ h_{j}^2{\bf{I}}_{d} \Biggr | j=1,2,..., |{B}| \Biggr\}. \nonumber
\end{eqnarray}
Each element of $B$ is predetermined by users before starting the algorithm. Then, we define the dictionary,
\begin{eqnarray} \label{Dictionary}
D = \Biggl\{ \phi_{{h}_{j}} (\cdot - \mathbf{X}_{i}^{*}) \Biggr | h_{j}^2{\bf{I}}_{d} \in B, i=1,2,...,m, j=1,2,...,|B| \Biggr\},\ \  |D| = m \times |B|,
\end{eqnarray}
where $\phi_{h_{j}} (\cdot - \mathbf{X}_{i}^{*})$ is a density function with its mean and variance-covariance matrix respectively $\mathbf{X}_{i}^{*}$  and ${h}_{j}^{2}\mathbf{I}_{d}$. Each word in $D$ is denoted by $\phi_{s}(\mathbf{x}|\mathbf{X}^{*}), s = 1,2, ..., m|B|$, where each index number $s$ corresponds to a combination $(\mathbf{X}_{i}^{*}$, $h_{j})$, $i=1,2,...,m$, $j=1,2,...|B|$, one-to-one.
\subsubsection*{Stagewise minimization algorithm} 

Let $M$ ($\ge 2$) be the number of iterations for the algorithm and $\epsilon > 0$ be the approximation bound. We employ the mixing coefficients,
\begin{eqnarray} \label{def.pi}
\pi_{k} = 1\ \mbox{and}\  0 < \pi_{k} = \frac{\theta}{k+\theta} < 1, \ k=1, ..., M-1, \mbox{with}\ \  \theta \ge 2.
\end{eqnarray}
From \eqref{e.loss}, the empirical loss is calculated by
\begin{eqnarray} \label{e.loss.alg}
\widehat{L}_{U}(g(\cdot|\mathbf{X}^{*})) = - \frac{1}{n} \sum_{i=1}^{n} \xi(g(\mathbf{X}_{i}^{+}|\mathbf{X}^{*})) + \int_{\mathbb{R}^{d}} U(\xi(g(\mathbf{x}|\mathbf{X}^{*}))) d\mathbf{x}, \nonumber
\end{eqnarray}
where $g(\cdot|\mathbf{X}^{*})$ is a function on $\mathbb{R}^{d}$ given $\mathbf{X}_{i}^{*}$. Then, the algorithm for the stagewise minimization estimator consists of the following steps:\\\\
{\bf{Step1.}} At the initial stage $k=0$, choose $\tilde{f_{0}} \in D$ so that
\begin{eqnarray} \label{step1}
\widehat{L}_{U}(\tilde{f_{0}}(\cdot|\mathbf{X}^{*})) &\le& \inf_{\phi \in D} \widehat{L}_{U}(\phi(\cdot|\mathbf{X}^{*})) + \epsilon. \nonumber
\end{eqnarray}
{\bf{Step2.}} For $k=1, ..., M-1$, let
\begin{eqnarray} \label{step1}
\tilde{f_{k}}(\mathbf{x}|\mathbf{X}^{*}) &=& u\left((1-\pi_{k}) \xi(\tilde{f}_{k-1}(\mathbf{x}|\mathbf{X}^{*})) + \pi_{k} \xi({\tilde{\phi}(\mathbf{x}|\mathbf{X}^{*})}) \right), \nonumber
\end{eqnarray}
where ${\tilde{\phi}} \in D$ is chosen so that
\begin{eqnarray} \label{step1}
\widehat{L}_{U}(\tilde{f_{k}}(\cdot|\mathbf{X}^{*})) &\le& \inf_{\phi \in D} \widehat{L}_{U}\left(u((1-\pi_{k})\xi(\tilde{f}_{k-1}(\cdot|\mathbf{X}^{*})) + \pi_{k} \xi({\phi}(\cdot|\mathbf{X}^{*})))\right) + \pi_{k} \epsilon. \nonumber
\end{eqnarray}
{\bf{Step3.}} Let $\widehat{f}(\mathbf{x}|\mathbf{X}^{*}) = \tilde{f}_{M-1}(\mathbf{x}|\mathbf{X}^{*})$.
\\\\
At the final step $M-1$, we obtain the sequence of the words chosen at each stage, \\$\tilde{\phi}_{0}(\mathbf{x}|\mathbf{X}^{*})$,$\tilde{\phi}_{1}(\mathbf{x}|\mathbf{X}^{*})$,...,$\tilde{\phi}_{M-1}(\mathbf{x}|\mathbf{X}^{*})$, and the density estimator using the algorithm has the form of
\begin{eqnarray} \label{def.f_hat}
\widehat{f}(\mathbf{x}|\mathbf{X}^{*}) &=& u \left( \sum_{l=0}^{M-1} q_{l} \xi(\tilde{\phi}_{l}(\mathbf{x}|\mathbf{X}^{*})) \right), \ \ \ q_{l} = \pi_{l} \prod_{t=l+1}^{M-1} (1-\pi_{t}).
\end{eqnarray}
We can verify that $\sum_{l=0}^{M-1} q_{l} = 1$. When we employ the $\beta$-power divergence function with $\beta=1.0$, the estimator \eqref{def.f_hat} is rewritten as
\begin{eqnarray} \label{def.f_hat.kde}
\widehat{f}(\mathbf{x}|\mathbf{X}^{*}) &=& \sum_{l=0}^{M-1} q_{l} \tilde{\phi}_{l}(\mathbf{x}|\mathbf{X}^{*}).
\end{eqnarray}
Since \eqref{def.f_hat.kde} is a convex combination of the words $\tilde{\phi}_{l}(\mathbf{x}|\mathbf{X}^{*}), l = 1,2,...,M-1$, KDE is a sort of an estimator in the form of \eqref{def.f_hat}.
{\Remark{\hspace{-2.0mm}{\bf{.}} \hspace{-2.0mm}}\label{Rmk.1} The integral of the estimator $\widehat{f}(\mathbf{x}|\mathbf{X}^{*})$ is not always 1. Hence, we may consider its normalized form
\begin{eqnarray} \label{def.normal.f_hat}
\widehat{f_{c}}(\mathbf{x}|\mathbf{X}^{*}) &=& \gamma^{-1} \widehat{f}(\mathbf{x}|\mathbf{X}^{*}), \ \ \ \gamma = \gamma(\mathbf{X}^{*}) = \int_{\mathbb{R}^{d}} \widehat{f}(\mathbf{x}|\mathbf{X}^{*}) d\mathbf{x}. \nonumber
\end{eqnarray}}
{\Remark{\hspace{-2.0mm}{\bf{.}} \hspace{-2.0mm}}\label{Rmk.2}}
The proportion of the dictionary data points in the total sample size $m/N$ influences the performance of the density estimation, and parameter $m$ serves as a kind of smoothing parameter. Letting the problem of optimizing $m$ aside, we assume parameter $m$ is given before starting the algorithm.
\section{Theoretical results} \label{Theoretical results}

We show the theoretical results of our proposed estimator. The main result is to show the non-asymptotic error bound of the estimator in Theorem~\ref{Theorem.eb}. We also show the non-asymptotic error bound of the normalized version of the proposed estimator in Theorem~\ref{Theorem.eb.nrm}.

In the theorems, we use the following notations.
Let $co(\xi(D))$ be the set of convex hull composed by $\phi_{i}(\cdot|\mathbf{X}^{*})$,
\begin{eqnarray}
co(\xi(D)) &=& \Biggl\{ \sum_{i=1}^{|D|} \lambda_{i} \xi(\phi_{i}(\cdot|\mathbf{X}^{*})) \biggl| \phi_{i}(\cdot|\mathbf{X}^{*}) \in D, \sum_{i=1}^{|D|}\lambda_{i} = 1, \lambda_{i} \ge 0 \Biggr \}. \nonumber
\end{eqnarray}
We consider a triplet
\begin{eqnarray}
\Phi = \Bigl ( \sum_{m=1}^{T}q_{m}\xi(\tilde{\phi}_{m}), \phi, \bar{\phi} \Bigr), \sum_{m=1}^{T}q_{m}\xi(\tilde{\phi}_{m}) \in co(\xi(D)), \phi, \bar{\phi} \in D. \nonumber
\end{eqnarray}
The set of these triplets is denoted by
\begin{eqnarray}
H(D) \equiv co(\xi(D)) \times D \times D = \Biggl \{ \Phi \Biggl| \sum_{m=1}^{T}q_{m}\xi(\tilde{\phi}_{m}) \in co(\xi(D)), \phi, \bar{\phi} \in D \Biggr \}. \nonumber
\end{eqnarray}
For $\delta \in$ [0, 1] and $\Phi \in H(D)$, we define
\begin{eqnarray} \label{psi}
\lefteqn{\psi_{U}(\delta, \Phi| \mathbf{X}^{*})} \nonumber \\
&=& \int_{\mathbb{R}^{d}} U^{''} \biggl( (1-\delta) \sum_{m=1}^{T}q_{m} \xi(\tilde{\phi}_{m}(\mathbf{x}| \mathbf{X}^{*})) + \delta \xi(\phi(\mathbf{x}| \mathbf{X}^{*})) \biggr) \bigl \{\xi(\phi(\mathbf{x}| \mathbf{X}^{*})) - \xi(\bar{\phi}(\mathbf{x}| \mathbf{X}^{*}) \bigr \}^{2} d\mathbf{x}. \nonumber
\end{eqnarray}
\subsection{The non-asymptotic error bound of the estimator} \label{error.bound}
To obtain the non-asymptotic error bound of the estimator, we use Assumption~\ref{Ass1} as follows.
{\Assumption{\hspace{-2.0mm}{\bf{.}} \hspace{-2.0mm} \\(i) The convex function $U(t)$ is twice differentiable. \\(ii) There exists a constant $B_{U}(\mathbf{X}^{*})^{2} > 0$ such that
\begin{eqnarray} \label{f_hat}
\sup_{\delta \in [0,1]} \sup_{\Phi \in H(D)} \psi_{U}(\delta, \Phi| \mathbf{X}^{*}) &\le& B_{U}(\mathbf{X}^{*})^{2}, \mbox{almost surely}. \nonumber
\end{eqnarray}\label{Ass1}}}\\
{\bf{Example~1. (The case of KL divergence)}}\\
If we introduce a constant $B_{KL}(\mathbf{X}^{*})^{2}$ defined as
\begin{eqnarray} \label{B_KL}
B_{KL}(\mathbf{X}^{*})^{2} = \sup_{\phi, \bar{\phi}, \tilde{\phi} \in D} \int_{\mathbb{R}^{d}} \tilde{\phi}(\mathbf{x}|\mathbf{X}^{*}) \{ \log \phi(\mathbf{x}|\mathbf{X}^{*}) - \log \bar{\phi}(\mathbf{x}|\mathbf{X}^{*}) \}^2 d\mathbf{x}, \nonumber
\end{eqnarray}
in the case where KL divergence is employed for evaluating the algorithm, we see that $\Psi_{U}(\delta, \Phi | \mathbf{X}^{*}) \le B_{KL}(\mathbf{X}^{*})^{2}$ for any $\delta \in [0, 1]$ and any $\Phi \in H$. The proof is provided in Appendix~C. We also evaluate the constant $B_{KL}(\mathbf{X}^{*})^{2}$ and derive its upper bound \eqref{upb.BKL} in Appendix~C, employing Gaussian densities with scalar bandwidth matrix in the dictionary. If we consider the upper bound \eqref{upb.BKL} in the case of KL divergence, Assumption~\ref{Ass1} is justified.\\

Then, we obtain Theorem~\ref{Theorem.eb}. The proof is given in Appendix~A.
{\Theorem{\hspace{-2.0mm}{\bf{.}} \hspace{-2.0mm}
For the density estimator $\widehat{f}(\mathbf{x}|\mathbf{X}^{*})$ in \eqref{def.f_hat}, it holds under Assumption~\ref{Ass1} that
\begin{eqnarray} \label{error.bound}
\lefteqn{E_{\mathbf{X}^{*}} E_{\mathbf{X}^{+}} \bigl[ D_{U}(f(\cdot), \widehat{f}(\cdot|\mathbf{X}^{*})) \bigr]} \nonumber \\
& & \le E_{\mathbf{X}^{*}} [\inf_{g \in co(\xi(D))}D_{U}(f(\cdot), u(g(\cdot|\mathbf{X}^{*})))] + 2 E_{\mathbf{X}^{*}} [E_{\mathbf{X}^{+}} [\sup_{\phi \in D} |\nu_{n}(\xi(\phi( \cdot | \mathbf{X}^{*})))]] \nonumber \\
& & \ \ \ + \frac{\theta^2}{M+(\theta-1)} E_{\mathbf{X}^{*}} \bigl[ B_{U}(\mathbf{X}^{*})^{2} \bigr] + \epsilon, \ \ \ \ \ \ \ 
\end{eqnarray}
where $\nu_{n}(\cdot)$ is the centered operator,
\begin{eqnarray}
\nu_{n}(\xi(\phi(\cdot|\mathbf{X}^{*}))) = \frac{1}{n} \sum_{i=1}^{n} \xi(\phi(\mathbf{X}_{i}^{+}|\mathbf{X}^{*})) - \int_{\mathbb{R}^{d}} \xi(\phi(\mathbf{x}|\mathbf{X}^{*})) f(\mathbf{x}) d\mathbf{x}. \nonumber
\end{eqnarray}} \label{Theorem.eb}}

The symbol $E_{\mathbf{X}^{*}}[ \cdot ]$ represents the expectation regarding the sample used for the dictionary, $\mathbf{X}_{i}^{*}$, whereas $E_{\mathbf{X}^{+}}[ \cdot ]$ does the one used for the algorithm, $\mathbf{X}_{i}^{+}$. The error bound in \eqref{error.bound} diminishes as $M$ increases.
{\Remark{\hspace{-2.0mm}{\bf{.}} \hspace{-2.0mm}}\label{Rmk.3}} In the right-hand side of \eqref{error.bound}, the expected value $E_{\mathbf{X}^{*}}[B_{U}(\mathbf{X}^{*})^{2}]$ appears. In the case of KL divergence (Example~1), it suffices that the fourth moment of $\mathbf{X}_{i}^{*}$ exists to ensure the finiteness of $E_{\mathbf{X}^{*}}[B_{U}(\mathbf{X}^{*})^{2}]$. See Appendix~D in detail.
\subsection{Error bound of the normalized form of the estimator} \label{error.bound.normalized}

To obtain the non-asymptotic error bound of the normalized form of the estimator, we use Assumption~\ref{Ass2}.
{\Assumption{\hspace{-2.0mm}{\bf{.}} \hspace{-2.0mm} There exist two constants $C_{U} > 0$ and $0 < \alpha \le 1$ such that
\begin{eqnarray}
\inf_{\delta \in [0,1]} U^{''}\Bigl(\xi\bigl((1-\delta)\bar{f}(\mathbf{x}|\mathbf{X}^{*}) + \delta \bar{f}_{c}(\mathbf{x}|\mathbf{X}^{*})\bigr)\Bigr) \ge C_{U} \bar{f}^{\alpha} (\mathbf{x}|\mathbf{X}^{*}) > 0 \nonumber
\end{eqnarray}
for any $\mathbf{x} \in \mathbb{R}^{d}$ and for any $\bar{f} = u(\sum_{m=1}^{T} q_{m} \xi(\tilde{\phi}_{m}(\mathbf{x}|\mathbf{X}^{*})))$ with $\sum_{m=1}^{T} q_{m} \xi(\tilde{\phi}_{m}(\mathbf{x}|\mathbf{X}^{*})) \in co(\xi(D))$, where $\bar{f}_{c}(\mathbf{x}|\mathbf{X}^{*})$ is the normalized form of $\bar{f}(\mathbf{x}|\mathbf{X}^{*})$.}\label{Ass2}}\\

Subsequently, we obtain Theorem~\ref{Theorem.eb.nrm}. The proof is given in Appendix~B.
{\Theorem{\hspace{-2.0mm}{\bf{.}} \hspace{-2.0mm} For the normalized form $\widehat{f}_{c}(\mathbf{x}|\mathbf{X}^{*})$ of $\widehat{f}(\mathbf{x}|\mathbf{X}^{*})$, it follows from Assumptions~\ref{Ass1} and \ref{Ass2} that
\begin{eqnarray}
\lefteqn{E_{\mathbf{X}^{*}} \Bigl[ E_{\mathbf{X}^{+}} \bigl[ D_{U}(f(\cdot), \widehat{f_{c}}(\cdot|\mathbf{X}^{*})) \bigr] \Bigr]} \nonumber \\
& & \le E_{\mathbf{X}^{*}} \Bigl[ E_{\mathbf{X}^{+}} \bigl[ D_{U}(f(\cdot), \widehat{f}(\cdot|\mathbf{X}^{*}))\bigr] \Bigr] \nonumber \\
& & + C_{U}^{-1} E_{\mathbf{X}^{*}} \Biggl[ E_{\mathbf{X}^{+}} \biggl[ \bigl|1-v_{\hat{f}}(\mathbf{X}^{*})^{-1} \bigr| \int_{\mathbb{R}^{d}} \bigl| \widehat{f_{c}}(\mathbf{x}|\mathbf{X}^{*}) - f(\mathbf{x}) \bigr| \widehat{f}(\mathbf{x}|\mathbf{X}^{*})^{1-\alpha} d\mathbf{x}\biggr] \Biggr], \ \ \ \nonumber \\
& & where\ \ v_{\hat{f}}(\mathbf{X}^{*}) = \int_{\mathbb{R}^{d}} \widehat{f}(\mathbf{x}|\mathbf{X}^{*})d\mathbf{x}. \nonumber
\end{eqnarray}}\label{Theorem.eb.nrm}}
{\Remark{\hspace{-2.0mm}{\bf{.}} \hspace{-2.0mm}}\label{Rmk.4}} Theorem~\ref{Theorem.eb.nrm} reveals that the bound for the normalized estimator $\widehat{f}_{c}(\mathbf{x}|\mathbf{X}^{*})$ corresponds to that for $\widehat{f}(\mathbf{x}|\mathbf{X}^{*})$ given in Theorem~\ref{Theorem.eb} along with an extra term.
{\Remark{\hspace{-2.0mm}{\bf{.}} \hspace{-2.0mm}}\label{Rmk.5}} We obtain $v_{f} = 1$, when the $\beta$ power divergence with $\beta=1.0$ is employed. In such a situation, the result of {{Theorem~\ref{Theorem.eb.nrm}}} coincides with that of Theorem~\ref{Theorem.eb}.
\section{Applications} \label{Applications}

\subsection{Practical setting}

{For the sake of practical use, we consider the dictionaries 1 and 2, which are denoted as $D_{1}$ and $D_{2}$, respectively. In dictionary~1, we use the following set of scalar bandwidth matrices:
\begin{eqnarray} \label{B1_Dictionary}
B_{1} = \Biggl\{ h^2 \mathbf{I}_{2} \Biggl| h=\widehat{h} \cdot \Biggl(\frac{m}{j} \Biggr)^{\frac{1}{6}}, \widehat{h} = \sqrt{\widehat{h}_{DPI, 11} \widehat{h}_{DPI, 22}}, j=1,2,...,5 \Biggr \},
\end{eqnarray}
where $\widehat{h}_{DPI, 11}$ and $\widehat{h}_{DPI, 22}$ are the DPI estimators in Duong and Hazelton (2003) of the bivariate diagonal bandwidth matrix, $\mbox{diag}(h_{11}^{2}, h_{22}^{2})$, calculated by the dictionary data $\mathbf{X}^{*}_{i}$, $i=1,2,...,m$. To obtain $\widehat{h}_{DPI, 11}^2$ and $\widehat{h}_{DPI, 22}^2$, we employ {\texttt{Hpi.diag}} function in '{\texttt{ks}}' library in R. 
The bandwidth that should be used for $\tilde{\phi}_{k}(\cdot)$ is larger in size than $\widehat{h}$, which is calculated by the number of $m$ data points, because the resulting estimator entails the convex combination of not more than $M$ kernel functions. In this sense, each word in $B_{1}$ is augmented by multiplying $\widehat{h}$ by the factor $(m/j)^{1/6}, j=1,2,...,5$.

In dictionary~2, we consider the following set of scalar bandwidth matrices:
\begin{eqnarray} \label{B2_Dictionary}
B_{2} &=& \Biggl\{ h^{2} \mathbf{I}_{2} \ \Biggl | h = [SD({X}_{1}^{*}) SD({X}_{2}^{*})]^{\frac{1}{2}} \cdot \left(\frac{2}{1 + \eta (j-1)}\right)^{\frac{1}{6}}, \ \ j=1 , 2, \cdots,10 \Biggr\},\ \
\end{eqnarray}
where $SD({X}_{p}^{*})$ is the standard deviation of $X_{ip}$, $i = 1, 2, ..., m$ and $p = 1, 2$. Parameter $\eta \ge 0$ is a tuning parameter, determined according to the sample size and/or the curvature of the true functions. We normally set $\eta = 1.0$, but we set $\eta=10.0$ in estimating Type J. If we assume parameter $\eta$ to be an increasing function of the sample size, then $h_{j}$ in \eqref{B2_Dictionary} is similar to the geometric mean of $h_{p} = SD({X}_{p}^{*}) N^{-1/6}, p=1,2$, which is Scott's rule in $\mathbb{R}^{d}$ (Scott 2017, p.164).
\subsection{Simulation}

We consider simulations~1 and 2 for the dictionaries $D_{1}$ and $D_{2}$ respectively. In each simulation case, we examine the behaviors of the proposed density estimator in terms of Mean Integrated Squared Error (MISE) when the proportion of the dictionary data points in the total sample size $m/N$ changes. We design the following five simulation cases for that purpose:
\begin{enumerate}[(a)]
\item
$\displaystyle{m=\frac{N}{4}}$, $\displaystyle{n=\frac{3N}{4}}$.
\item
$\displaystyle{m=\frac{N}{2}}$, $\displaystyle{n=\frac{N}{2}}$; however $\mathbf{X}_{i}^{*} \neq \mathbf{X}_{i}^{\tiny{+}}$, $i=1,2,..., \displaystyle{\frac{N}{2}}$.
\item
$m=\displaystyle{\frac{3N}{4}}$, $n=\displaystyle{\frac{N}{4}}$.
\item
$\mathbf{X}_{i}^{*} = \mathbf{X}_{i}^{\tiny{+}}$, $i=1,2,..., \displaystyle{\frac{N}{2}}$.
\item
$\mathbf{X}_{i}^{*} = \mathbf{X}_{i}^{\tiny{+}}$, $i=1,2,..., N$.
\end{enumerate}
Cases (a), (b), and (c) examine the impact of the ratio $m/N$ to the behaviors of the proposed density estimators. Cases (d) and (e) are designed for comparison. In case (d), half of the original i.i.d. sample $\mathbf{X}_{i}$, $i=(N/2)+1,(N/2)+2,..., N$, is discarded and the remainder $\mathbf{X}_{i}$, $i=1,2,..., N/2$, is used for both the dictionary data $\mathbf{X}_{i}^{*}$ and the algorithm data $\mathbf{X}_{i}^{+}$. In case (e), the original i.i.d. sample $\mathbf{X}_{i}$, $i=1,2,..., N$, is used in common for the dictionary and the algorithm. 

In each simulation case, we use the bivariate simulation settings of Wand and Jones (1993), Type C, J and L, whose contour plots are shown in Figure~\ref{true.C.J.L}.
For each simulation setting, we generate a sample of size $N$; we retain one part of it for the dictionary and use the remainder for calculating empirical loss, and run the algorithm. We repeat this process 10 times and obtain MISE by averaging the ISEs calculated for each process. We consider three alternatives to our estimator, KDE1 and KDE2 with Duong and Hazelton's (2003) DPI full bandwidth matrix and DPI diagonal bandwidth matrix, respectively, as well as RSDE. For the divergence function, we employ the $\beta$-power divergence function in \eqref{beta.pow.div} and set the tuning parameter $\beta$ to be $0.5$ and $1.0$. We denote our estimators minimizing the $\beta$-power divergence with $\beta= 0.5$ and $\beta = 1.0$ to be $\widehat{f}_{0.5}$ and $\widehat{f}_{1.0}$, respectively. For the parameter $\theta$ in the mixing coefficient in \eqref{def.pi}, we set $\theta=2$, following Klemel\"{a} (2007). The total iterations of the algorithm are $M=100$.
\begin{figure}[h]
\begin{center}
\includegraphics*[width=0.3\linewidth]{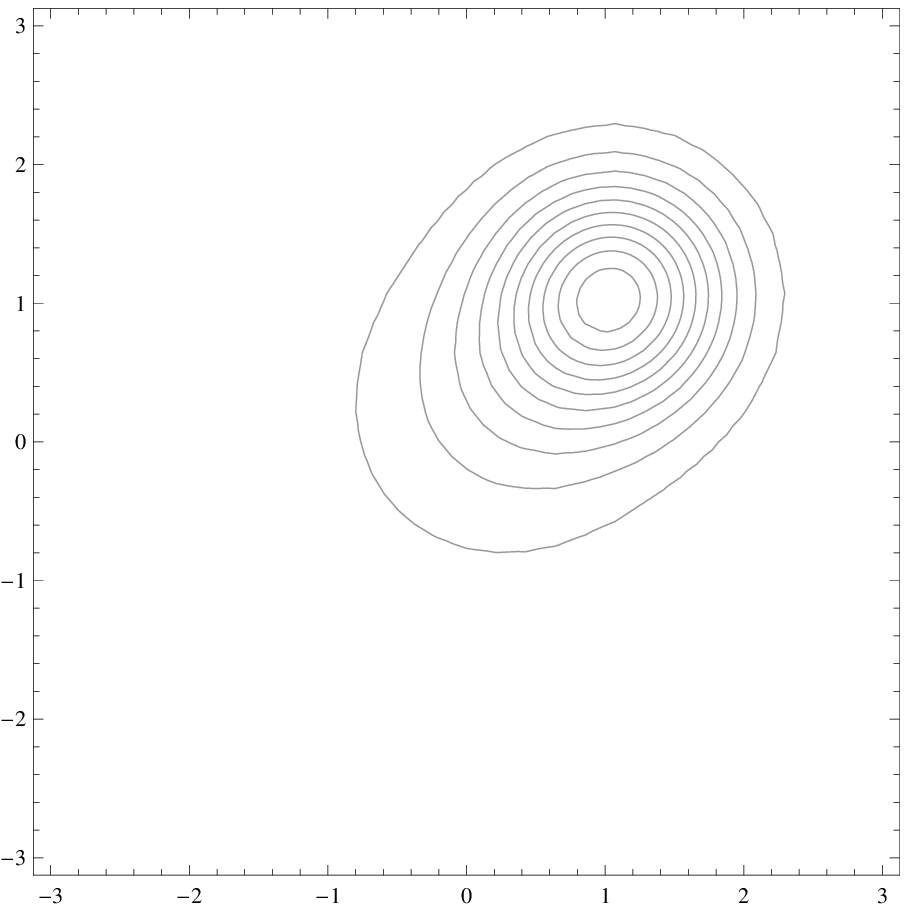}
\includegraphics*[width=0.3\linewidth]{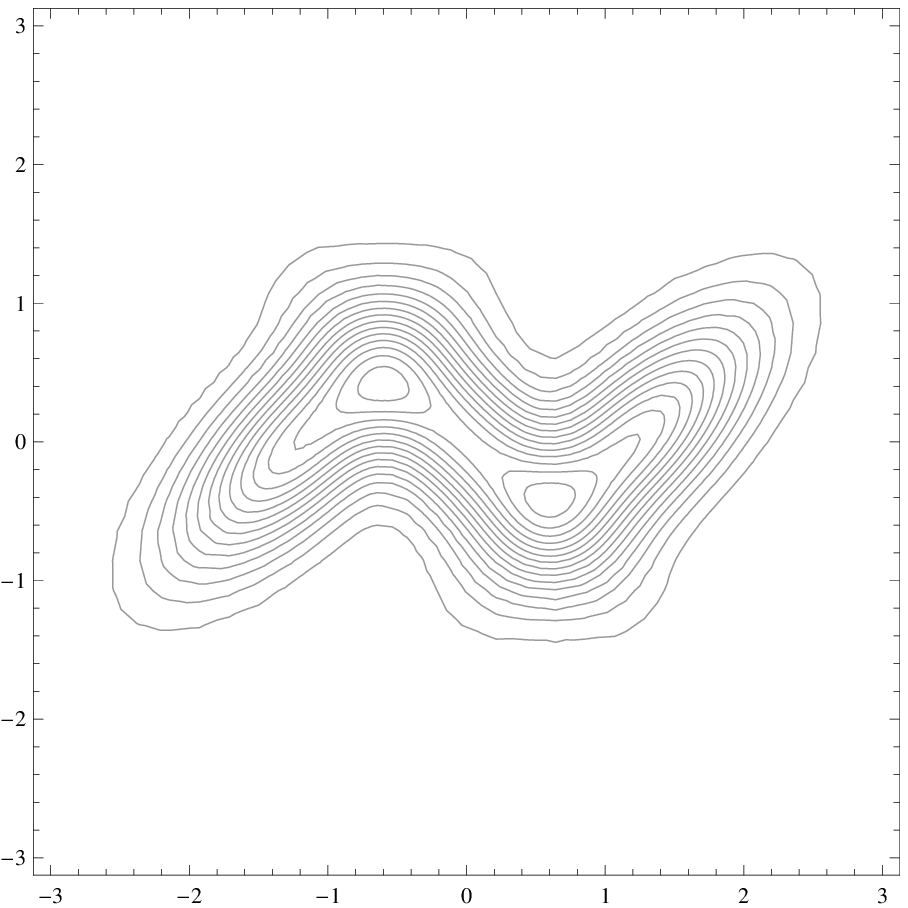}
\includegraphics*[width=0.3\linewidth]{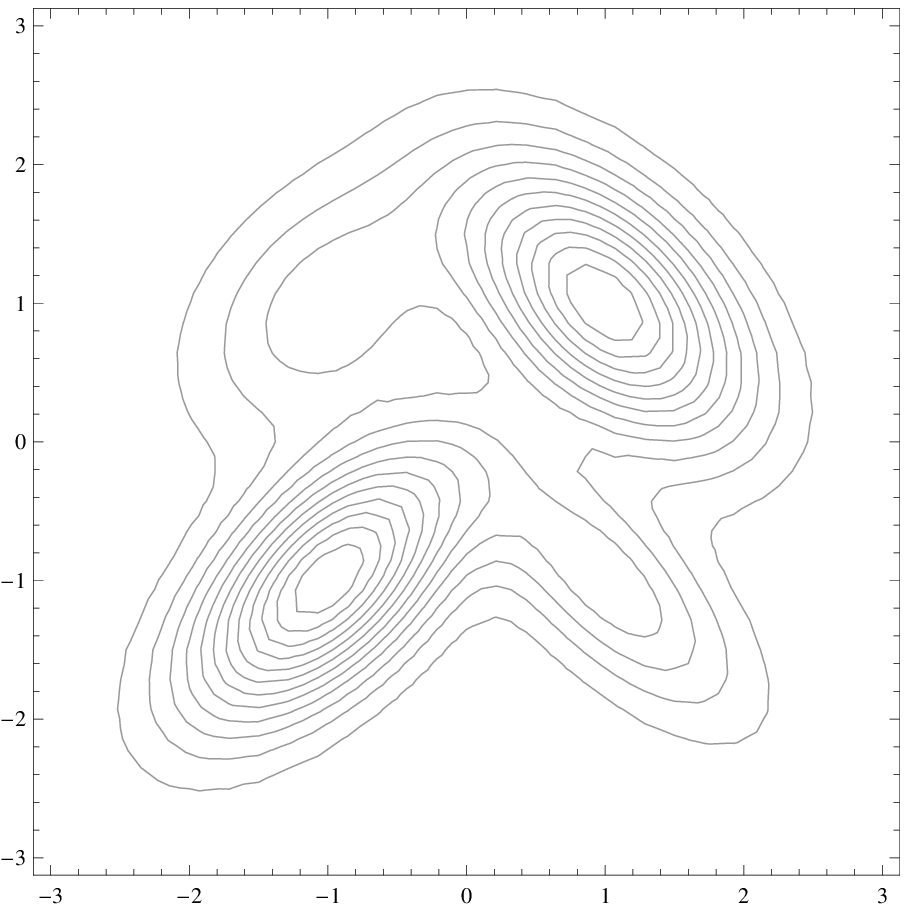}
\end{center}
\caption[]{True density functions: Left=Type C. Center=Type J. Right=Type L.} \label{true.C.J.L}
\end{figure}
\subsubsection{Simulation~1}

We present the numerical results of $\widehat{f}_{0.5}$ and $\widehat{f}_{1.0}$ in Tables~\ref{tab.results.sim1.beta0.5} and~\ref{tab.results.sim1.beta1.0}, respectively, in terms of MISE. The visual presentation of the results for $\widehat{f}_{0.5}$ in Type~L and for $\widehat{f}_{1.0}$ in Types C, J and L is given in Figure~\ref{MISE.for.m.sim1}. We visually present the results of (b) in simulation~1 for Type~C in Figure~\ref{results.C.sim1}. In the figure, the two upper panels represent the plot of the value of MISE for each iteration step $k=1,...,M$ of the algorithm. The middle and bottom panels in Figure~\ref{results.C.sim1} are the contour plots of the estimators. The red points in the contour plots are the data points used for the dictionary, while the blue ones are those chosen for estimation by the algorithm. The number of blue points is less than $M$ because the algorithm chooses the same data points more than once. 

We see the following findings of $\widehat{f}_{1.0}$ in terms of MISE. In the case of Type C, we observe the cases (a) and (e) for $N=200$ and $400$ can outperform KDE1, KDE2, and RSDE; the cases (b) and (d) for $N=400$ can do those (see Table~\ref{tab.results.sim1.beta1.0} and the two panels in the second column of Figure~\ref{MISE.for.m.sim1}). This result is important in that our estimator can be superior to the three alternatives with the help of DPI bandwidth matrix estimator. In the case of Type J, we observe the case (e) for $N=200$ can outperform RSDE; the case (e) for $N=400$ can do KDE2 and RSDE (see Table~\ref{tab.results.sim1.beta1.0} and the two panels in the third column of Figure~\ref{MISE.for.m.sim1}). In the case of Type L, we observe case (e) for $N=200$ and cases (b) and (d) for $N=400$ can outperform RSDE; case (e) for $N=400$ can do KDE2 and RSDE (see Table~\ref{tab.results.sim1.beta1.0} and the two rightmost panels in Figure~\ref{MISE.for.m.sim1}). The reason that the estimator in Type~C performs better than those in Type J and L is that the true function of Type C is a symmetric and is compatible with a scalar bandwidth matrix. We also observe a general trend that case (e) performs better than cases (a)-(d) in terms of MISE except for $\widehat{f}_{0.5}$ in the case of Type L (see Table~\ref{tab.results.sim1.beta0.5} and the two leftmost panels in Figure~\ref{MISE.for.m.sim1}).

We compare our estimators with RSDE by the degree of data condensation. In Table~\ref{tab.c.ratio}, we show the data condensation ratios of our estimators and RSDE. In the columns of RSDE and (I), we show the ratios of the actual data points used for estimating the density function in the number of total data points $N$. In columns (II), we show the ratios of the actual number of words in $D_{1}$ used for the estimations in the number of total words $|D_{1}| = m \times |B_{1}|$. We observe four results. First, our method yields lower data condensation ratios in terms of (I) and (II) than RSDE in all situations. Second, we observe that the case (a) yields the smallest data condensation ratios (I) and (II) in all situations. Third, the ratio (II) decreases as $m$ increases. Fourth, the ratios (I) and (II) in the case of $\beta=1.0$ are greater than those of $\beta=0.5$ in each simulation setting. The case of $\beta=0.5$ uses less data points and words for estimation than that of $\beta=1.0$.
\begin{table}
\begin{center}
{\scriptsize{
\begin{tabular}{clllllllll}
\hline
\hline
$k+1$ & 1 & 25 & 50 & 75 & 100 & KDE1 & KDE2 & RSDE \\
\hline
Type C & & & & & & & \\
$\underline{N=200}$ & --- & --- & --- & --- & --- &  84(26) & 81(28) & 106(31) \\
$(a)$ & 267(333) & 89(61) & 85(59) & 84(57) & 85(56) \\

$(b)$ & 318(598) & 106(122) & 106(132) & 101(118) & 105(129) \\
$(c)$ & 278(387) & 132(119) & 141(145) & 147(146) & 146(153) \\
$(d)$ & 274(550) & 100(134) & 111(160) & 108(155) & 112(150) \\
$(e)$ & 115(78) & 57(35) & 57(35) & 57(37) & 58(39) \\
$\underline{N=400}$ & --- & --- & --- & --- & --- & 53(14) & 54(11) & 84(18) \\
$(a)$ & 272(501)  & 72(76)  & 73(85)  & 73(82) & 75(88) \\
$(b)$ & 177(203)  & 76(73)  & 71(61)  & 76(70)  & 76(70)  \\
$(c)$ & 247(322)  & 145(159)  & 142(151)  & 142(148)  & 140(153)  \\
$(d)$ & 176(207)  & 69(48) & 72(56) & 75(60) & 75(60)  \\
$(e)$ & 151(129)  & 60(47) & 61(42)  & 60(41)  & 58(40)  \\
\hline
Type J & & & & & & & \\
$\underline{N=200}$ & --- & --- & --- & --- & --- & 108(17) & 118(31) & 138(33) \\
$(a)$ & 1900(3402)  & 404(327)  & 415(388)  & 409(343)  & 399(340)  \\
$(b)$ & 1096(594)  & 301(73)  & 302(90)  & 298(86)  & 291(75)  \\
$(c)$ & 1146(530)  & 377(132)  & 376(142)  & 356(127)  & 354(126)  \\
$(d)$ & 1093(587)  & 335(49)  & 350(50)  & 336(44)  & 342(54)  \\
$(e)$ & 1150(327)  & 293(43)  & 274(46)  & 272(37)  & 271(34)  \\
$\underline{N=400}$ & --- & --- & --- & --- & --- & 74(10) & 80(19) & 111(19) \\
$(a)$ & 983(611)  & 273(32)  & 271(31)  & 269(27)  & 268(28)  \\
$(b)$ & 1213(520)  & 311(52)  & 297(60)  & 302(54)  & 296(54)  \\
$(c)$& 1266(481)  & 322(80)  & 308(78)  & 314(71)  & 314(75)  \\
$(d)$ & 1231(555) & 284(56)  & 278(54)  & 277(53)  & 277(51)  \\
$(e)$ & 1309(342)  & 267(34)  & 265(29)  & 264(32) & 260(28)  \\
\hline
Type L & & & & & & \\
$\underline{N=200}$ & --- & --- & --- & --- & --- & 67(14) & 77(14) & 131(87) \\
$(a)$ & 574(223)  & 185(44)  & 183(42)  & 180(41)   & 177(40)   \\
$(b)$ & 748(324)  & 199(92)  & 183(76)   & 190(85)  & 183(71)   \\
$(c)$ & 1157(876)  & 314(248) & 319(232) & 332(254)  & 329(250)  \\
$(d)$ & 777(352)  & 226(56) & 235(78)  & 225(72) & 225(69)   \\
$(e)$ & 889(321) & 191(67)  & 177(51)  & 180(56)  & 180(61)  \\
$\underline{N=400}$ & --- & --- & --- & --- & --- & 45(6) & 54(18) & 98(26) \\
$(a)$ & 630(237)  & 171(23)  & 160(27)  & 158(26)  & 158(25)  \\
$(b)$ & 968(349)  & 185(33)  & 173(21)  & 180(22)  & 178(22)   \\
$(c)$ & 1022(420)  & 232(92) & 214(74)  & 213(66)  & 214(76)   \\
$(d)$ & 933(335) & 170(56)  & 179(80)  & 181(67)  & 182(79)   \\
$(e)$ & 1062(321) & 173(39) & 164(37)  & 165(32)  & 165(33)   \\
\hline
\hline
\end{tabular}}}
\caption[]
{Simulation~1: Result of MISE $\times 10^{4}$ (standard deviation $\times 10^{4}$). ($\beta=0.5$)} \label{tab.results.sim1.beta0.5}
\end{center}
\end{table}
\begin{table}
\begin{center}
{\scriptsize{
\begin{tabular}{cllllllll}
\hline
\hline
$k+1$ & 1 & 25 & 50 & 75 & 100 & KDE1 & KDE2 & RSDE \\
\hline
Type C & & & & & & & \\
$\underline{N=200}$ & --- & --- & --- & --- & --- &  84(26) & 81(28) & 106(31) \\
$(a)$ & 270(337)  & 56(25) & 55(24) & 56(27) & 59(27) \\
$(b)$ & 341(590)  & 79(40) & 84(41) & 85(44) & 86(47) \\
$(c)$ & 291(436) & 130(94) & 125(84) & 120(81) & 118(76) \\
$(d)$ & 297(603) & 79(91)  & 79(83) & 81(88) & 85(91) \\
$(e)$ & 119(85) & 42(22) & 43(25) & 46(29) & 46(29) \\
$\underline{N=400}$ & --- & --- & --- & --- & --- & 53(14) & 54(11) & 84(18) \\
$(a)$ & 297(552) & 38(21) & 34(19) & 34(15) & 33(14) \\
$(b)$ & 173(204) & 48(28) & 50(30) & 49(30) & 50(32) \\
$(c)$ & 265(321) & 84(47) & 84(48) & 87(46) & 89(49) \\
$(d)$ & 174(209) & 48(27) & 46(30) & 47(26) & 50(31) \\
$(e)$ & 156(139) & 24(9)  & 24(9)  & 26(12) & 26(10) \\
\hline
Type J & & & & & & & \\
$\underline{N=200}$ & --- & --- & --- & --- & --- &  108(17) & 118(30) & 138(33) \\
$(a)$ & 1932(3390)  & 209(73)  & 179(43)  & 178(43)  & 180(44)  \\
$(b)$ & 1146(567)  & 196(35)  & 190(41)  & 187(39)  & 179(40)  \\
$(c)$ & 1210(546)  & 249(68)  & 241(70)  & 241(72)  & 242(66)  \\
$(d)$ & 1130(596)  & 219(56)  & 205(52)  & 206(49)  & 202(50)  \\
$(e)$ & 1160(357)  & 139(24)  & 122(30)  & 126(30)  & 124(29)  \\
$\underline{N=400}$ & --- & --- & --- & --- & --- & 74(10) & 80(19) & 111(19) \\
$(a)$ & 1004(606)  & 163(44)  & 145(50)  & 140(50)  & 140(53)  \\
$(b)$ & 1248(514)  & 146(36)  & 135(30)  & 131(33)  & 130(32)  \\
$(c)$& 1297(455)  & 158(28)  & 152(42)  & 146(35)  & 152(36)  \\
$(d)$ & 1219(548)  & 136(29)  & 119(31)  & 124(33)  & 117(32)   \\
$(e)$ & 1302(324)  & 100(13)  & 84(16)  & 83(13)  & 79(15)  \\
\hline
Type L & & & & & & \\
$\underline{N=200}$ & --- & --- & --- & --- & --- & 67(14) & 77(14) & 131(87) \\
$(a)$ & 574(190)  & 111(24)  & 108(28)  & 106(31)  & 104(30)   \\
$(b)$ & 740(279)  & 118(25)  & 113(23)   & 113(24)  & 111(25)   \\
$(c)$ & 1177(861)  & 193(72)  & 186(64)  & 180(53)  & 179(57)   \\
$(d)$ & 760(273)  & 138(25)  & 130(33)   & 126(31)  & 129(30)   \\
$(e)$ & 851(281)  & 92(17)  & 86(19)  & 83(17)  & 83(17)  \\
$\underline{N=400}$ & --- & --- & --- & --- & --- & 45(6) & 54(18) & 98(26) \\
$(a)$ & 610(208)  & 92(17)  & 84(20) & 82(21) & 80(20)   \\
$(b)$ & 911(340)  & 81(17)  & 81(26) & 78(24) & 75(21)   \\
$(c)$ & 971(424)  & 119(26)  & 107(24) & 107(22) & 105(20)   \\
$(d)$ & 919(336)  & 84(23)  & 77(21) & 74(20) & 75(19)   \\
$(e)$ & 988(315)  & 66(10)  & 51(8) & 47(7) & 48(8)   \\
\hline
\hline
\end{tabular}}}
\caption[]
{Simulation~1: Result of MISE $\times 10^{4}$ (standard deviation $\times 10^{4}$). ($\beta=1.0$)} \label{tab.results.sim1.beta1.0}
\end{center}
\end{table}
\begin{sidewaysfigure}
\begin{center}
\includegraphics*[width=0.24\linewidth]{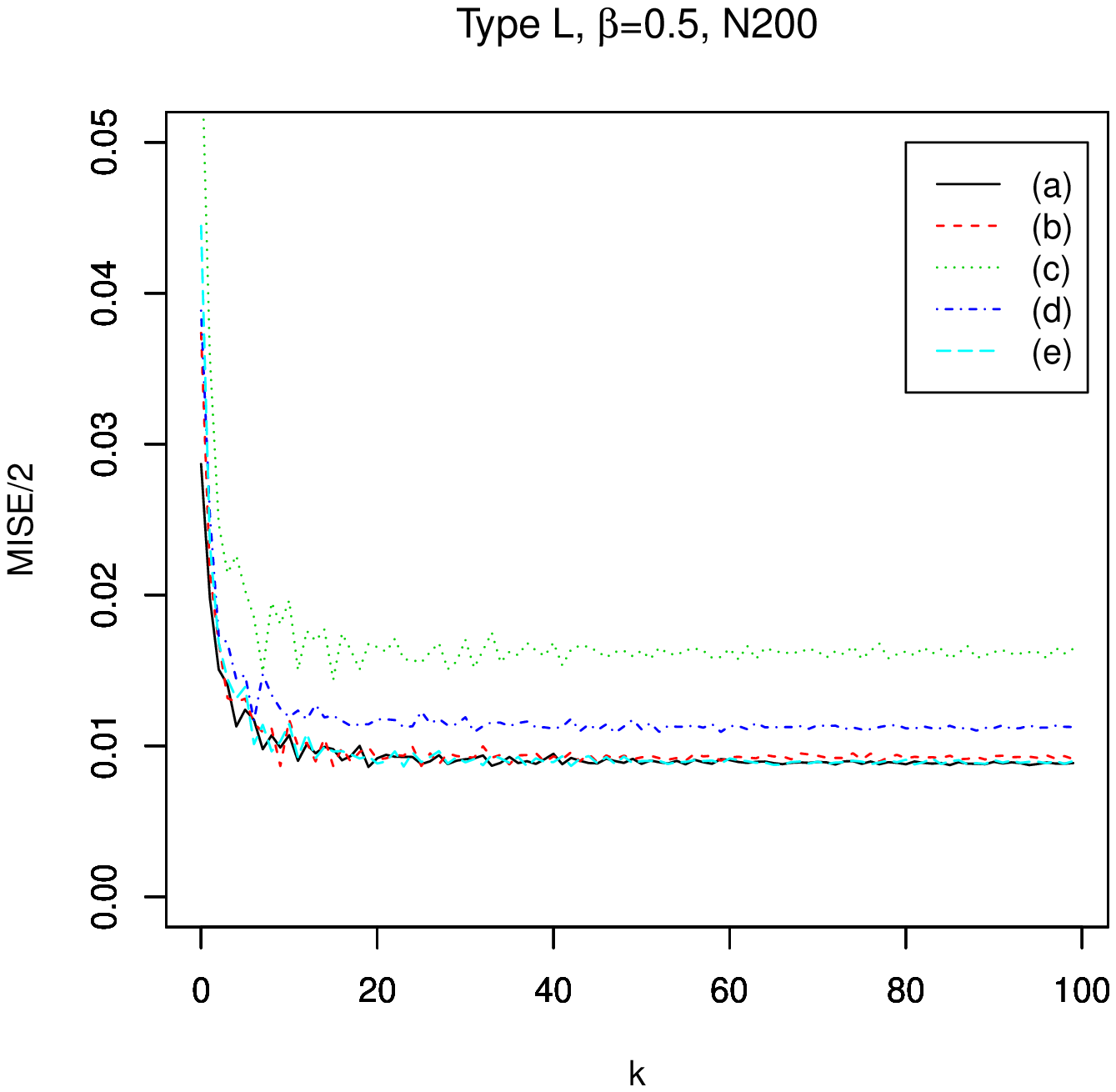}
\includegraphics*[width=0.24\linewidth]{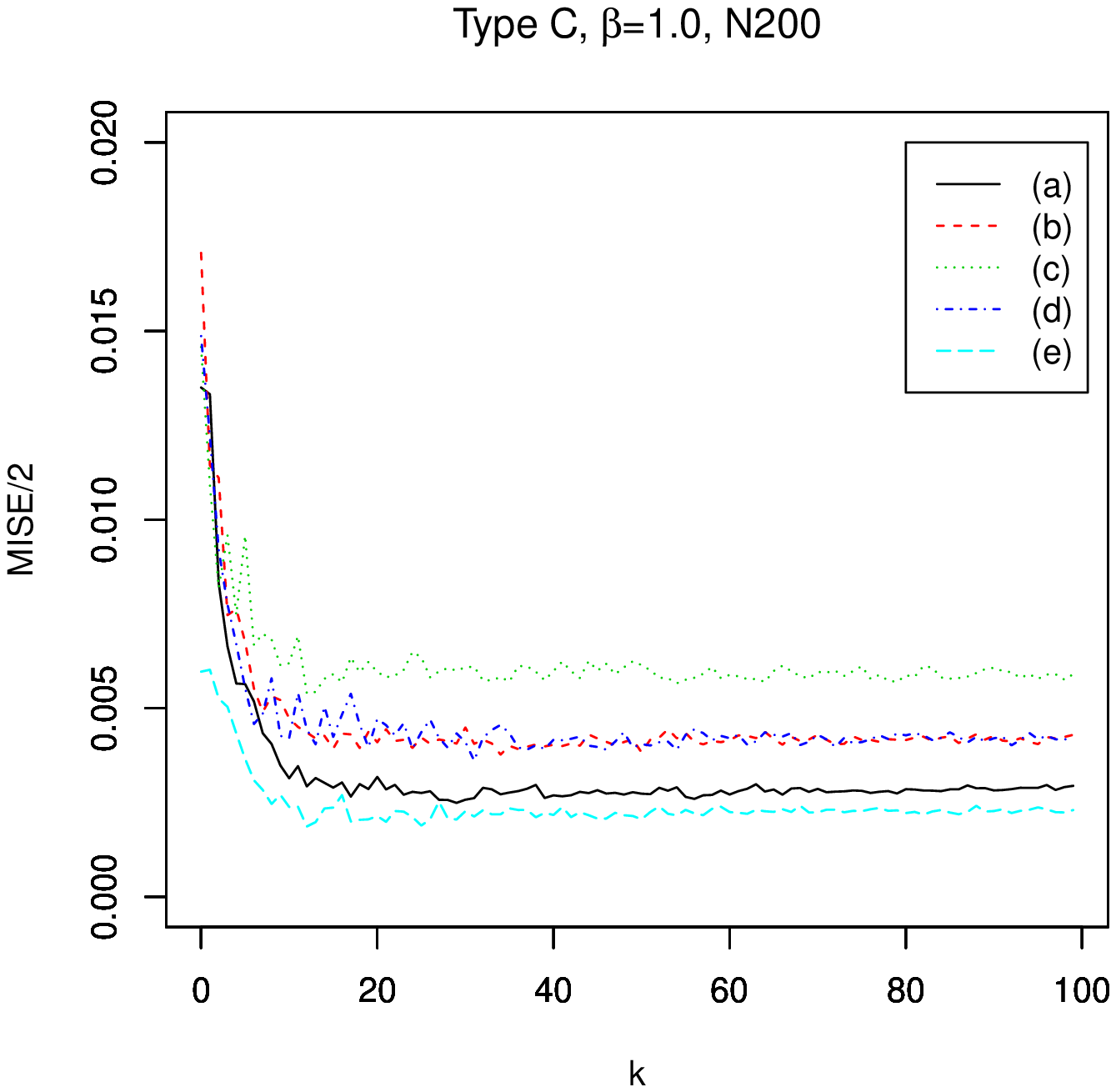}
\includegraphics*[width=0.24\linewidth]{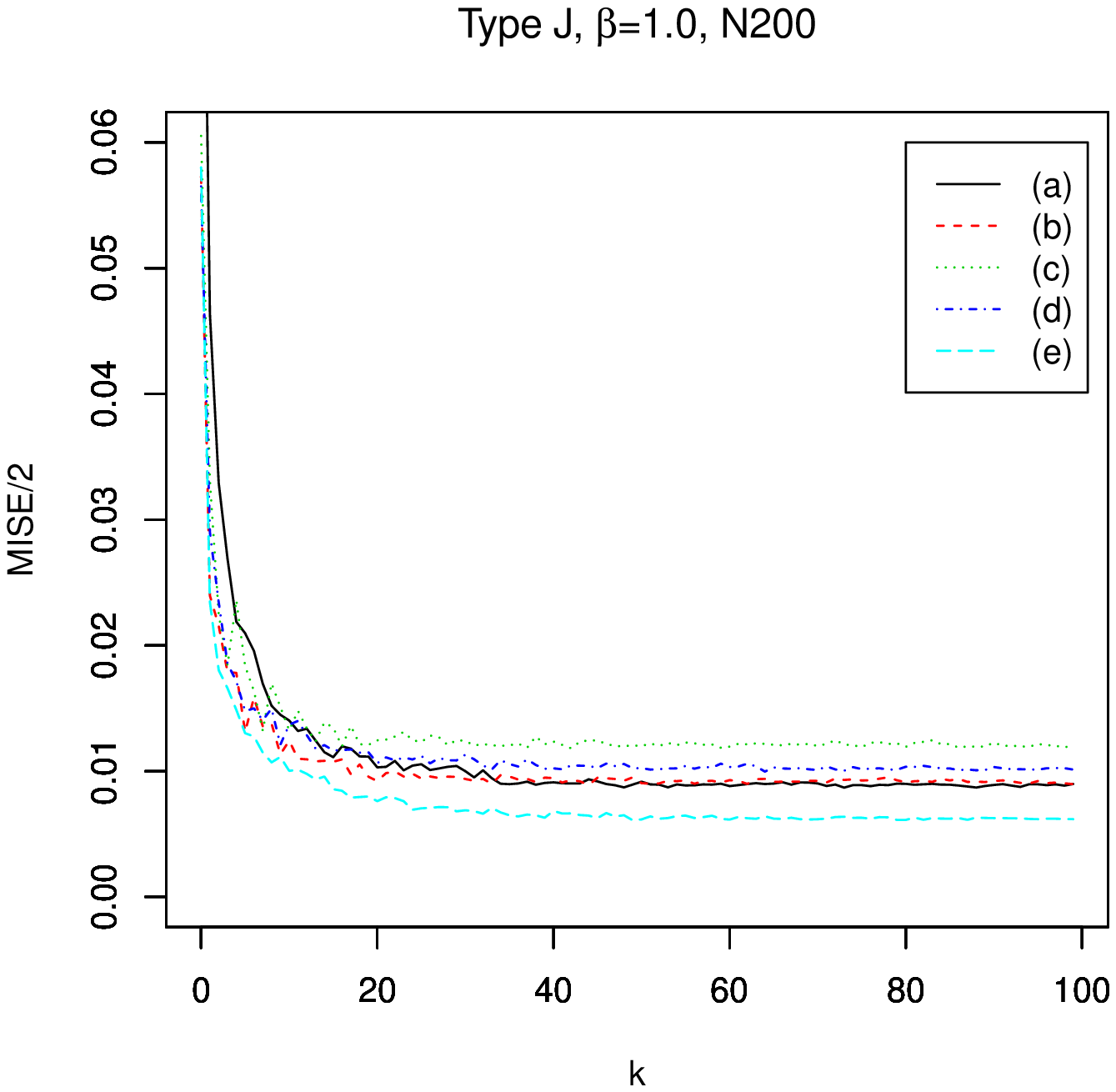}
\includegraphics*[width=0.24\linewidth]{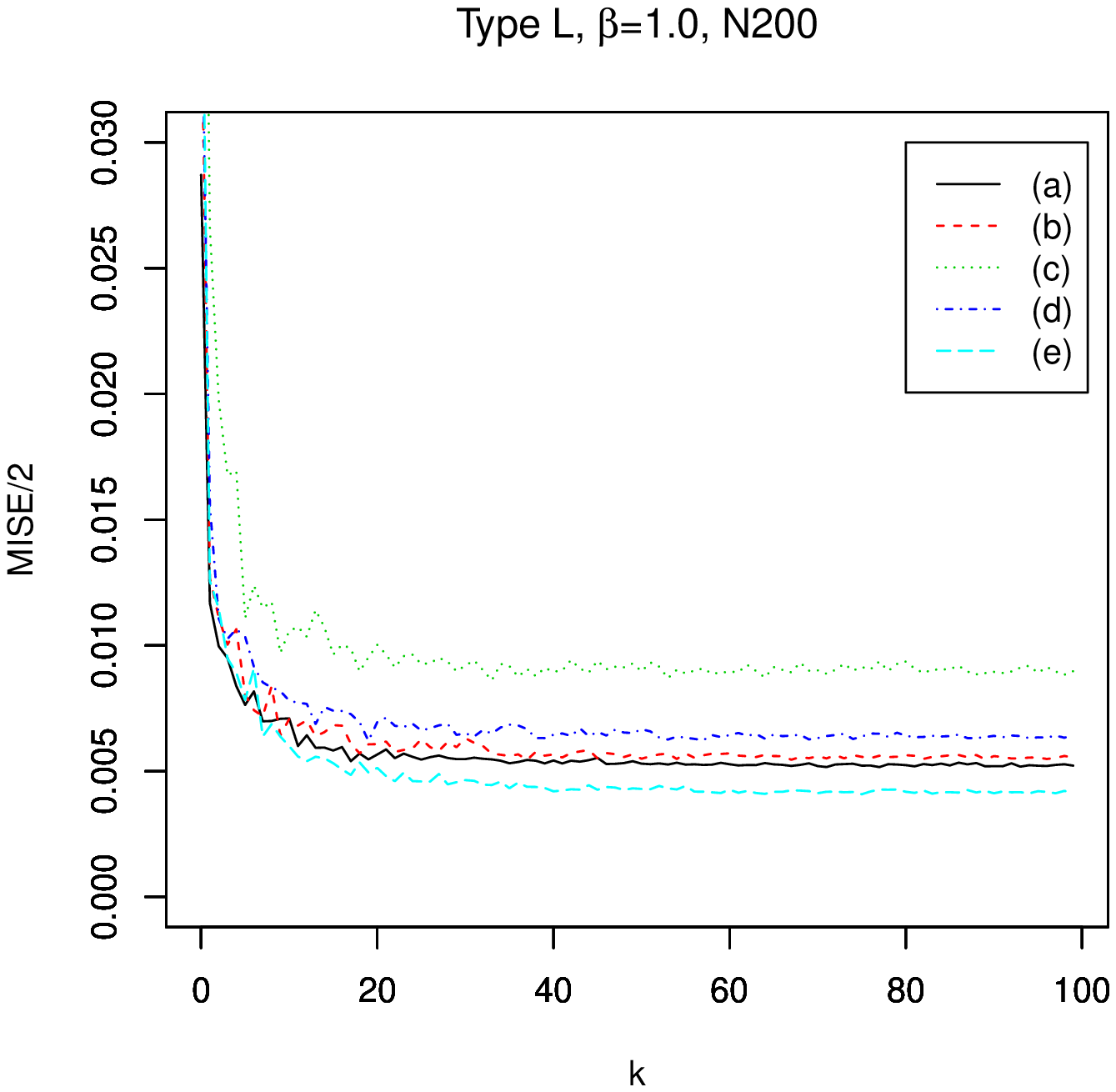}
\\
\includegraphics*[width=0.24\linewidth]{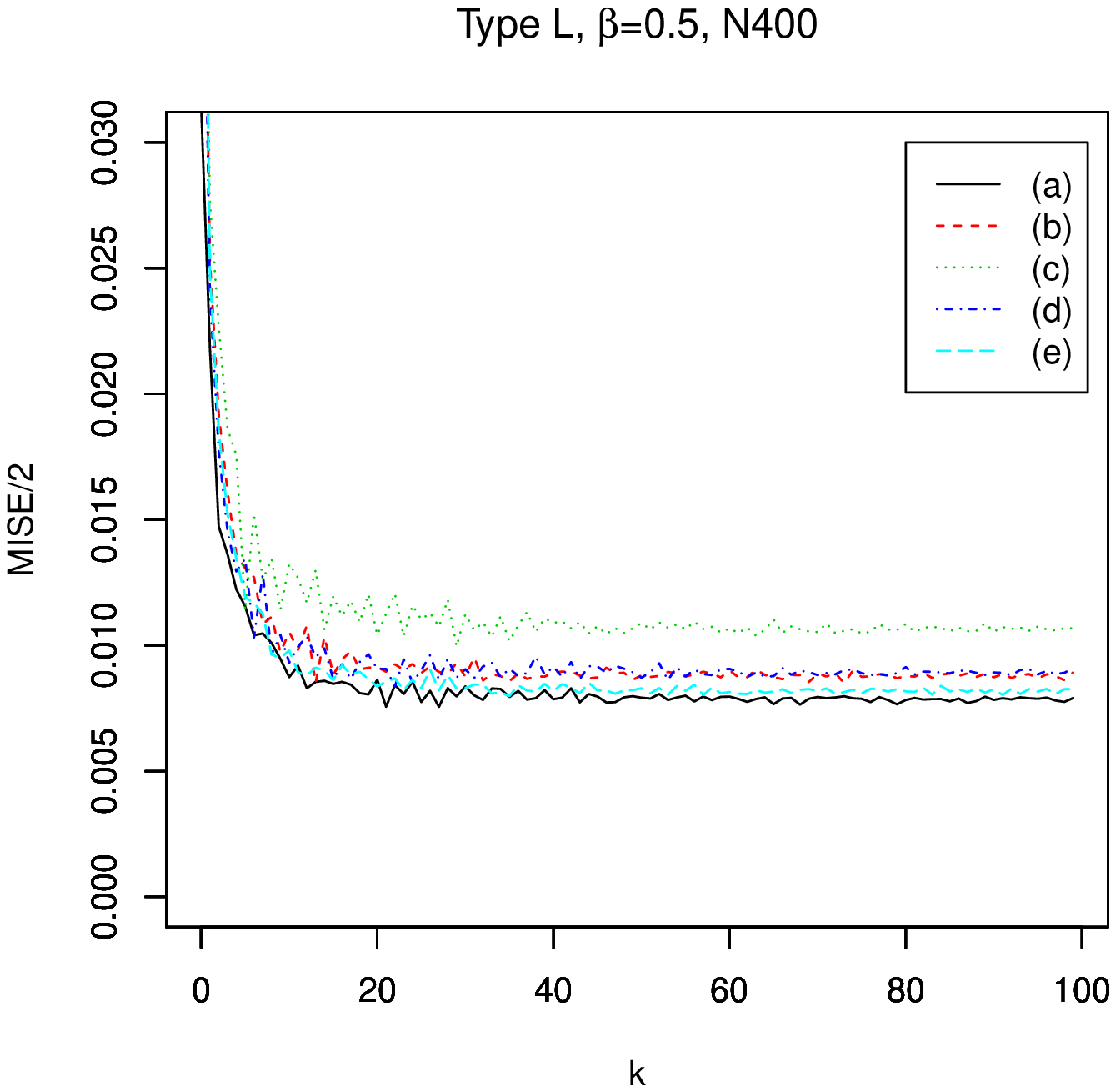}
\includegraphics*[width=0.24\linewidth]{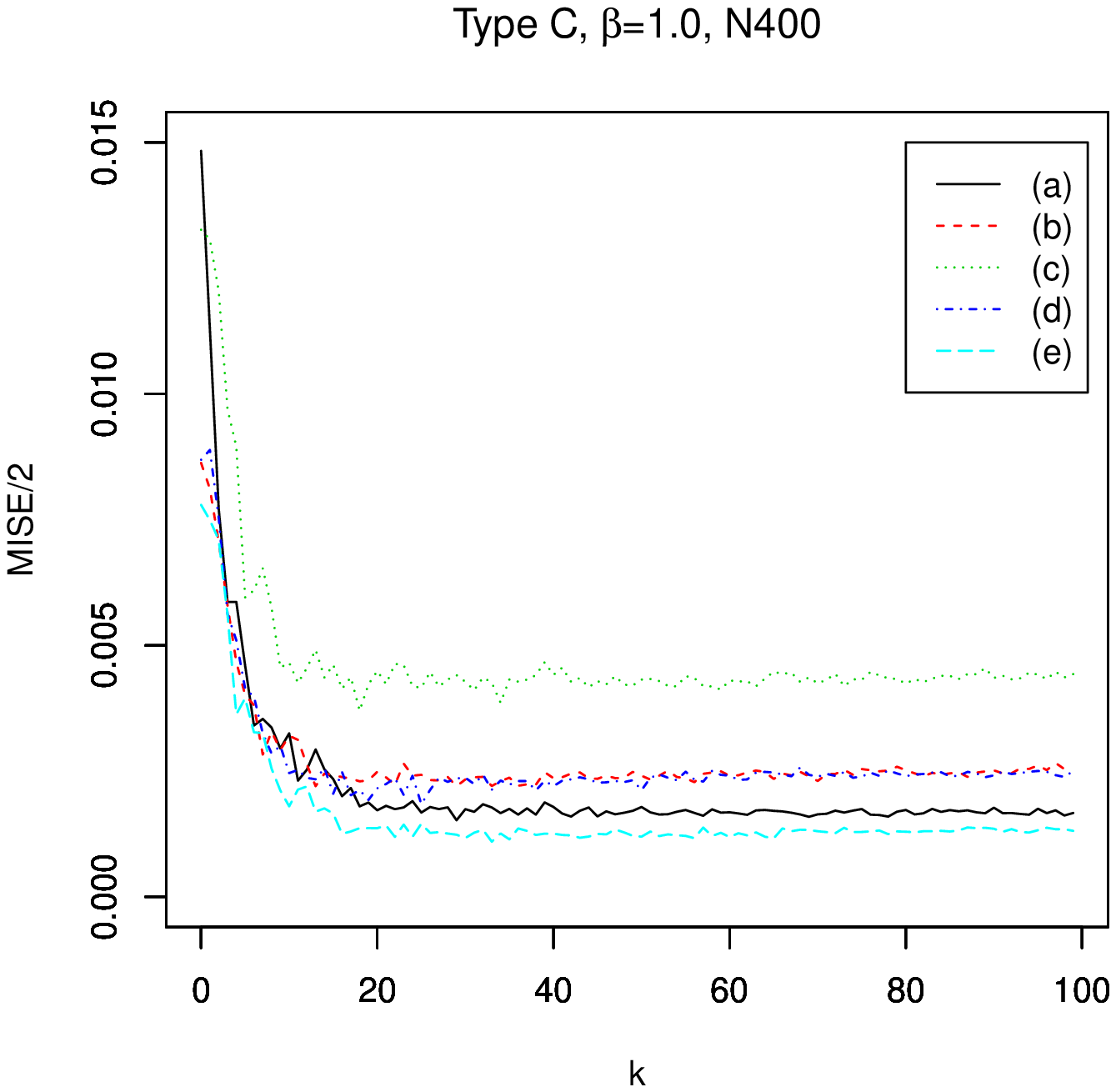}
\includegraphics*[width=0.24\linewidth]{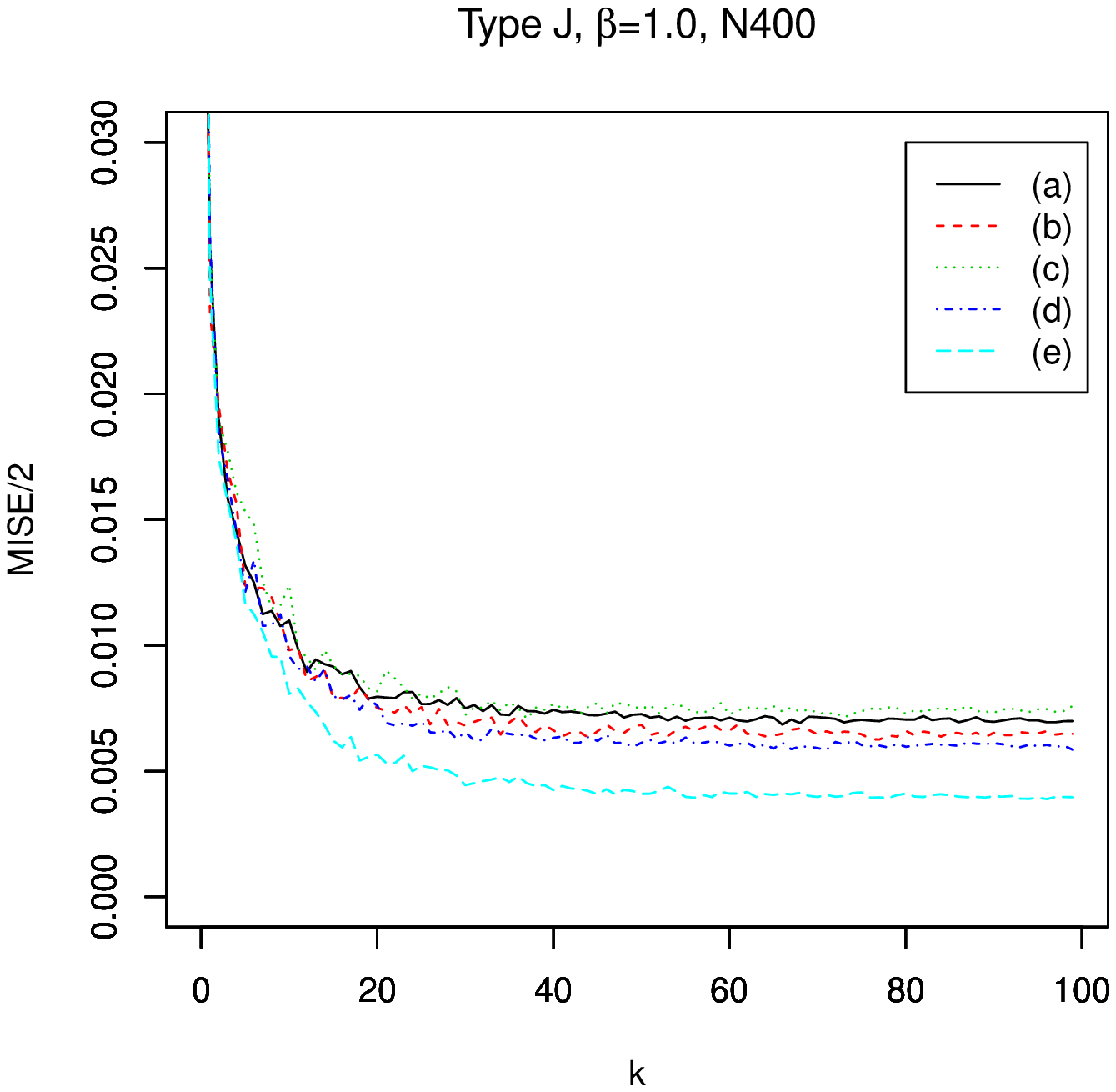}
\includegraphics*[width=0.24\linewidth]{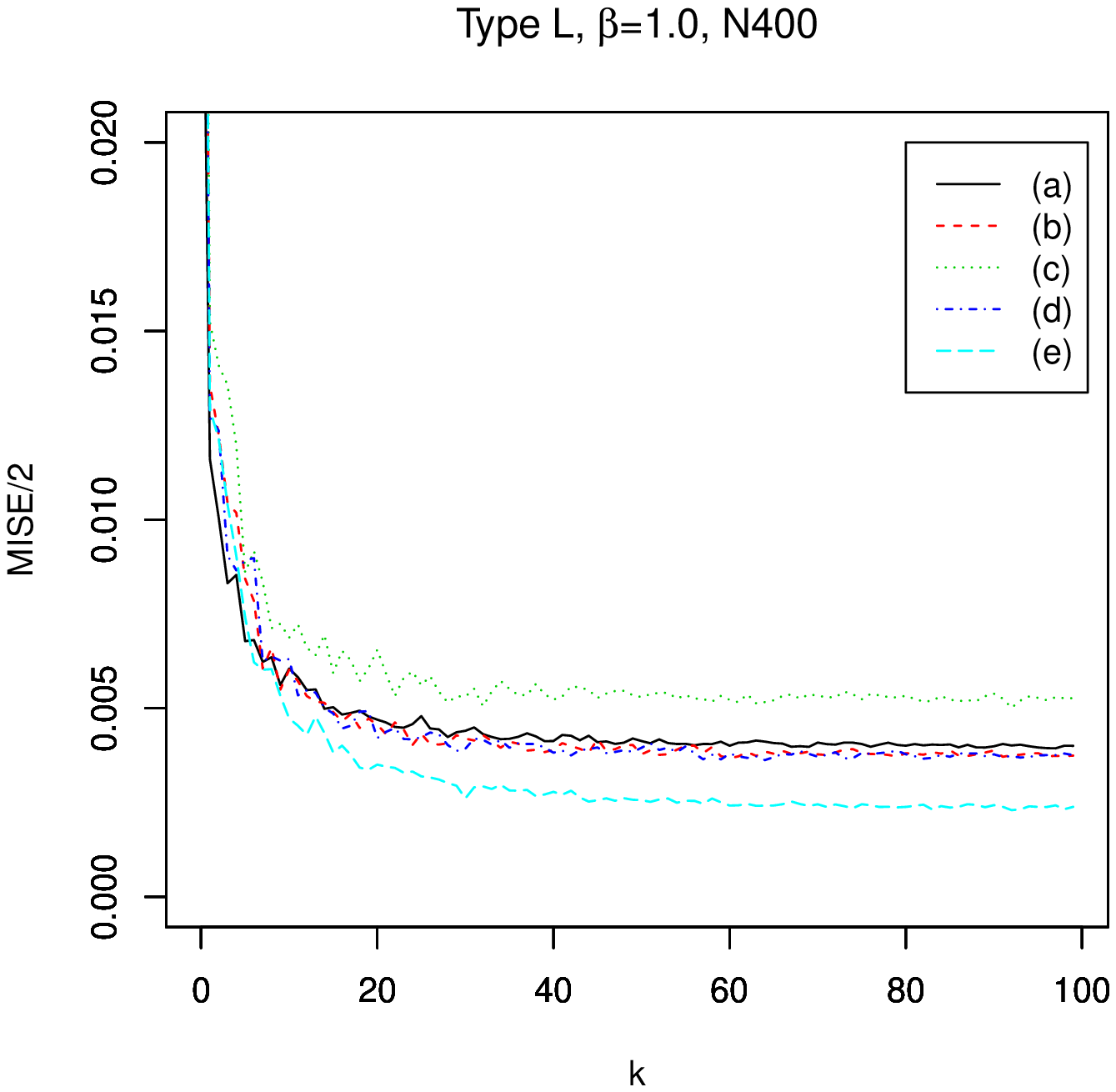}
\end{center}
\caption[]{Simulation~1. Plot of MISEs at each stage of algorithm for different $m/N$.} \label{MISE.for.m.sim1}
\end{sidewaysfigure}
\begin{figure}[htpb]
\begin{center}
\includegraphics*[width=0.4\linewidth]{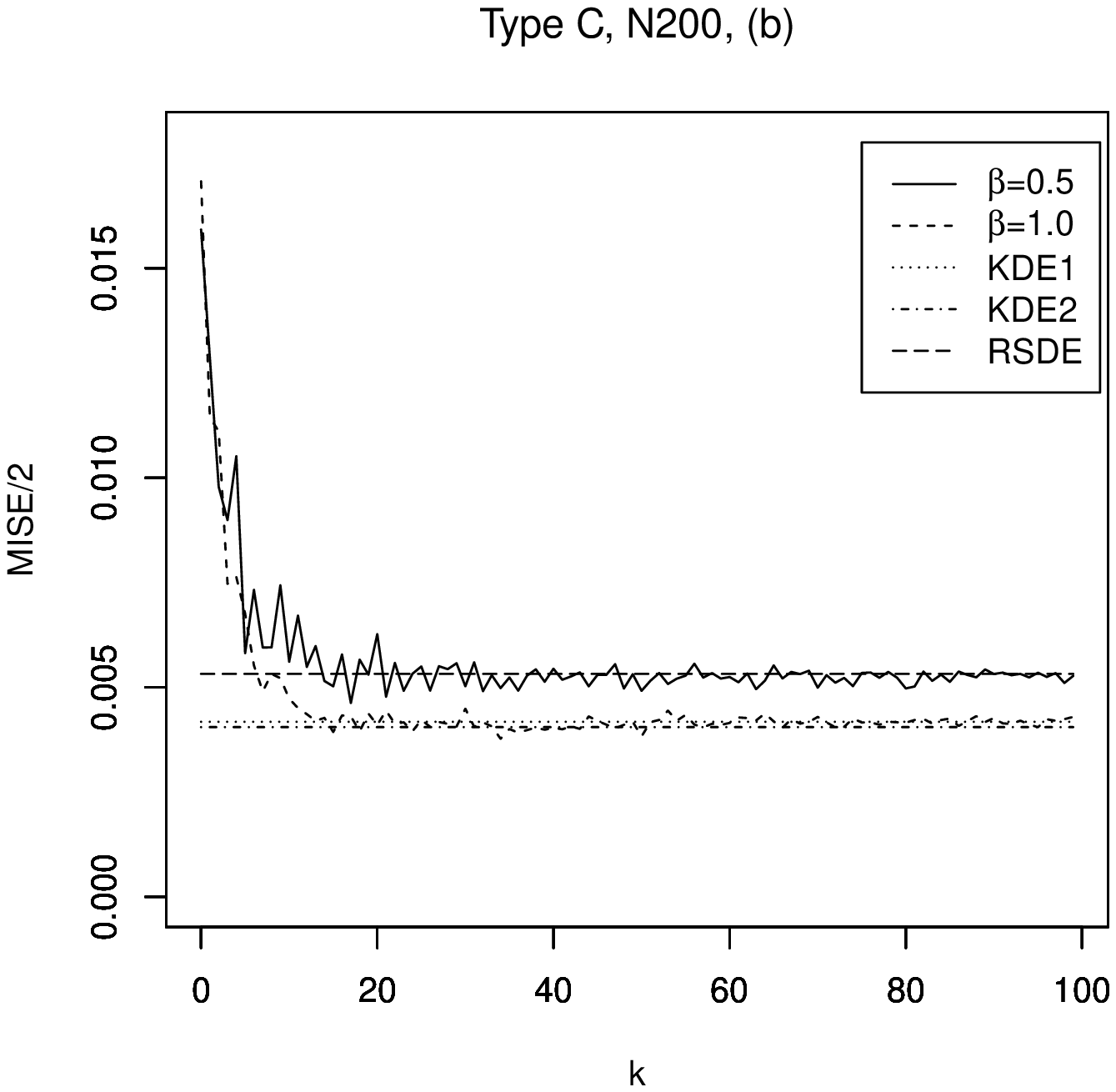}
\includegraphics*[width=0.4\linewidth]{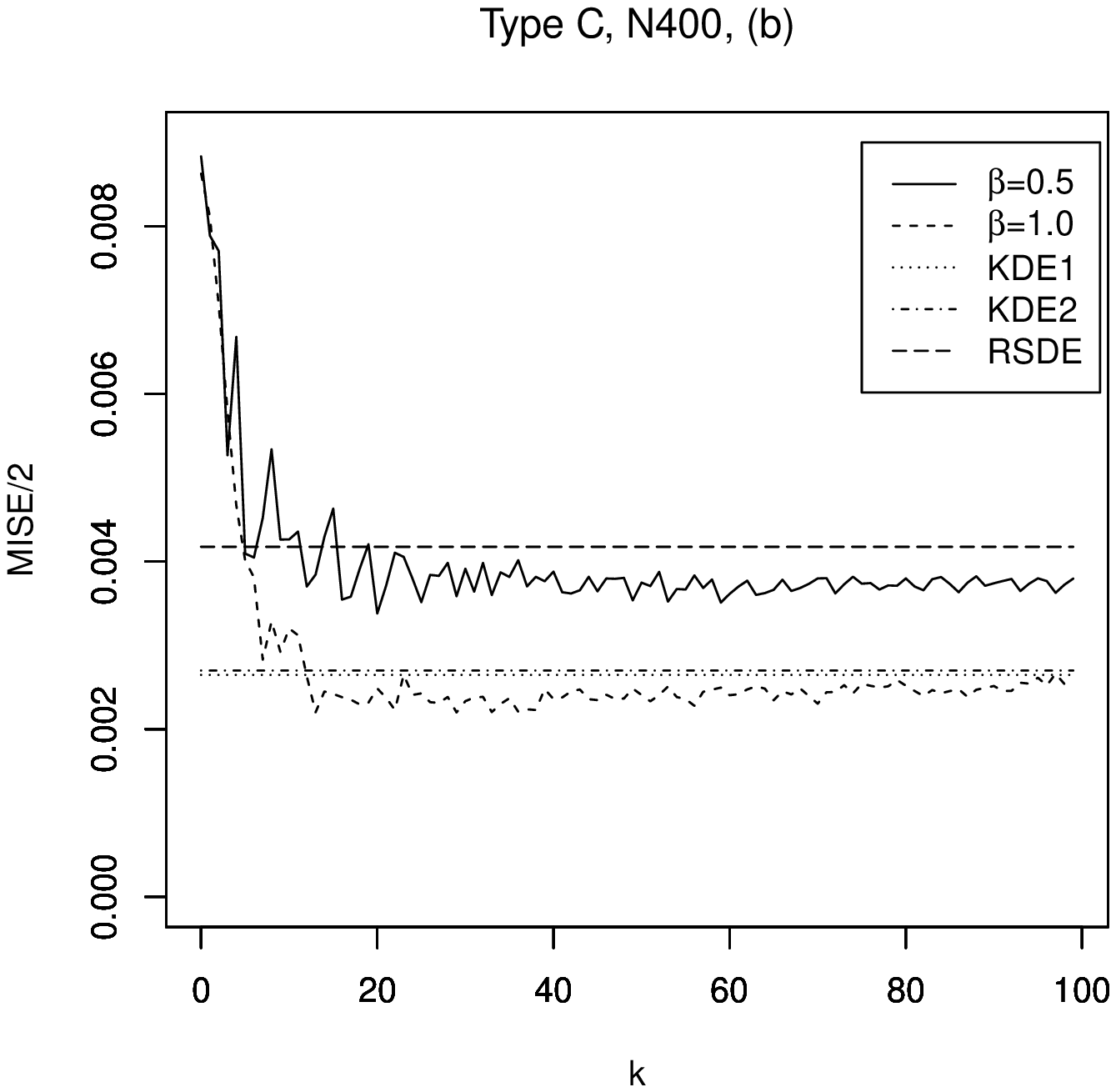}
\includegraphics*[width=0.4\linewidth]{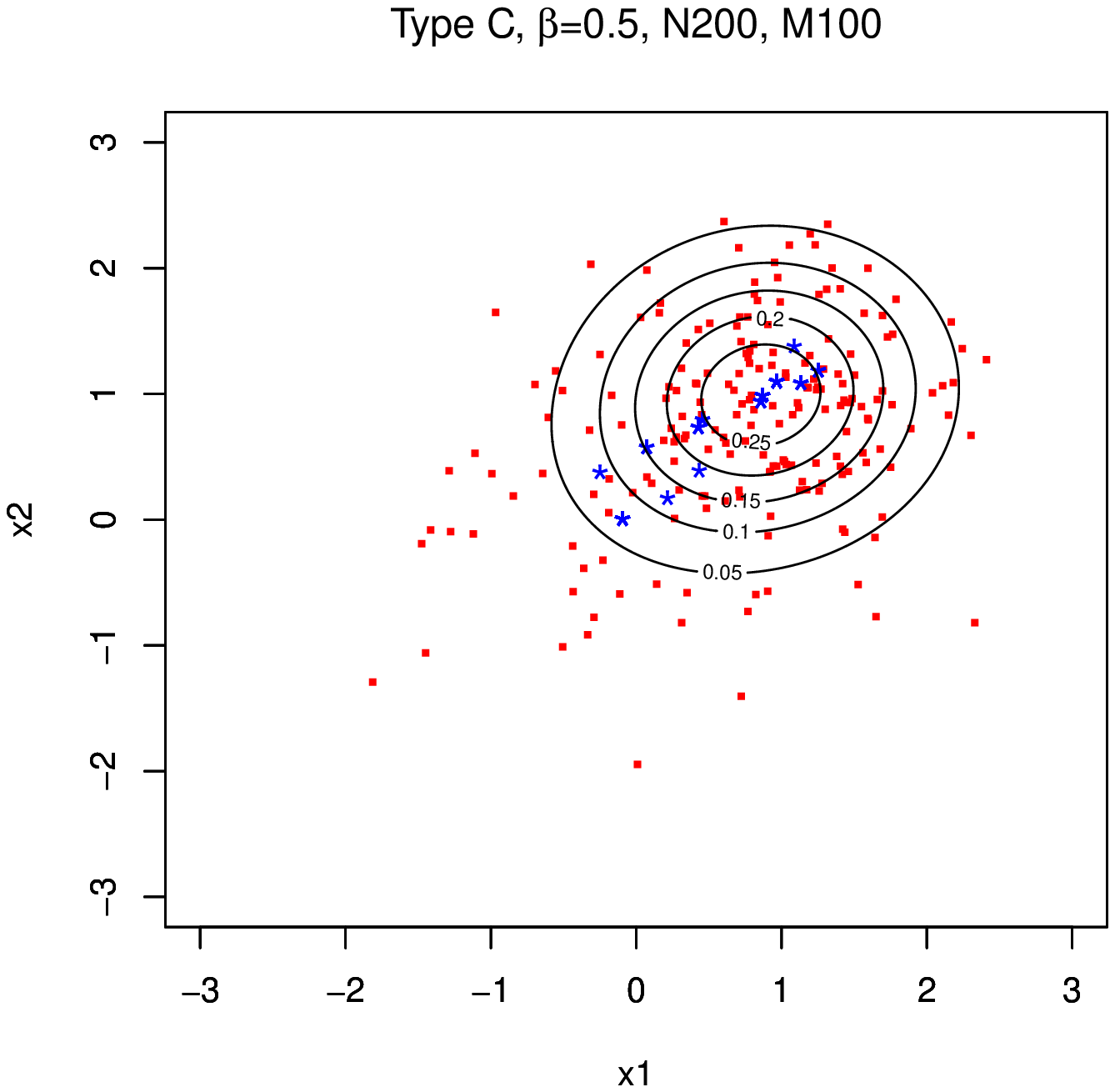}
\includegraphics*[width=0.4\linewidth]{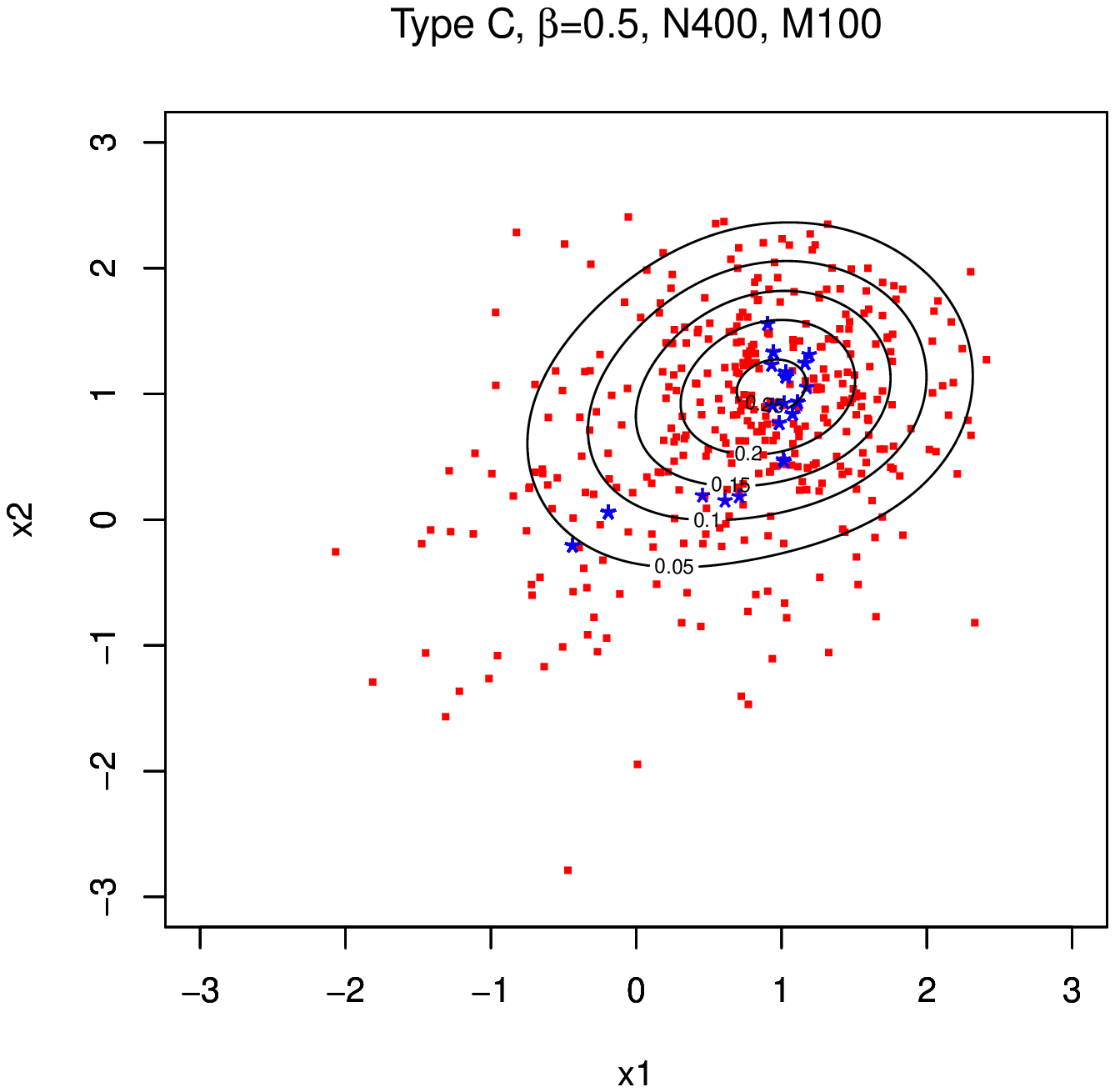}
\includegraphics*[width=0.4\linewidth]{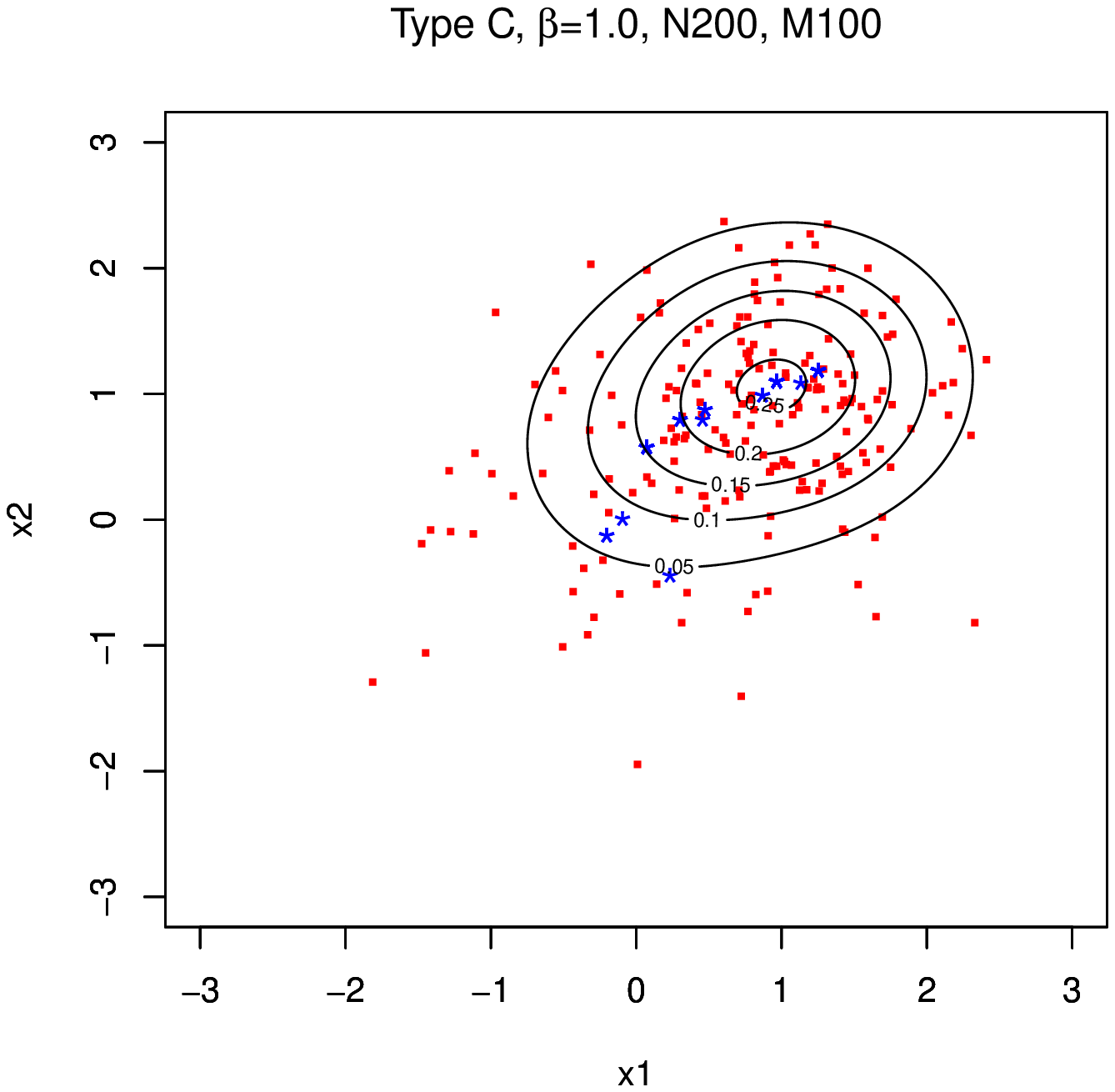}
\includegraphics*[width=0.4\linewidth]{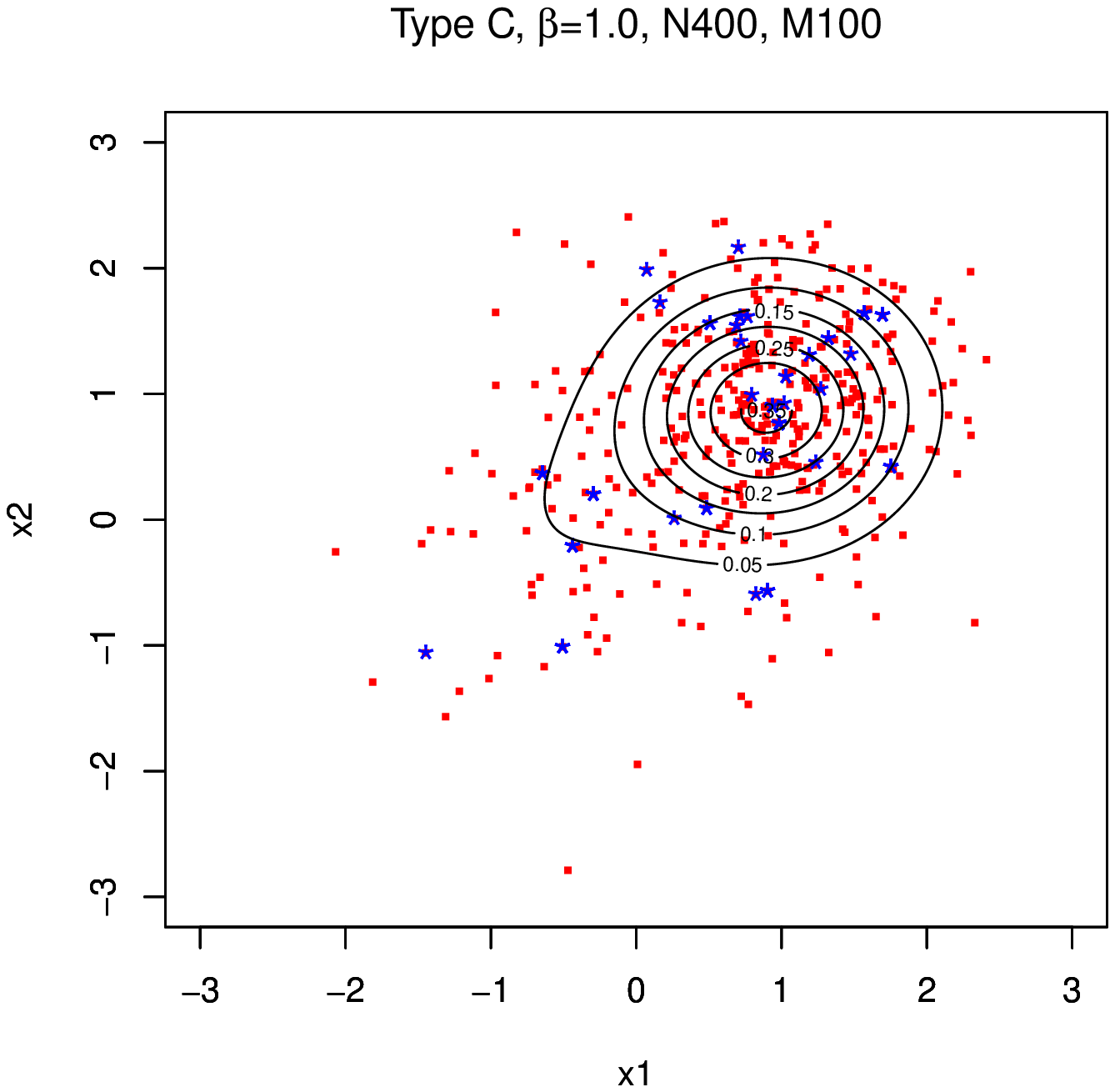}
\end{center}
\caption[]{Simulation~1: Upper: Plots of MISE vs $k$.\ \ \  Middle: Contour plots of $\widehat{f}_{0.5}$.\ \ \ Bottom: Contour plots of $\widehat{f}_{1.0}$} \label{results.C.sim1}
\end{figure}
\begin{table}
\begin{center}
{\scriptsize{
\begin{tabular}{c|c|c|cc|cc}
\hline
\hline
Type  & & RSDE & \multicolumn{2}{c|}{$\beta = 0.5$} & \multicolumn{2}{c}{$\beta = 1.0$} \\
 & & & (I) & (II) & (I) & (II) \\
\hline
C & $N=200$ & 1920 (219) & ---  & ---  & --- & --- \\
  & (a) & ---  & 495 (155) & 484 (135) & 830 (368) & 984 (429) \\
  & (b) & ---  & 565 (180)  & 284 (\ 64) & 905 (370) & 500 (213) \\
  & (c) & ---  & 660 (185)  & 190 (\ 49)  & 890 (321) & 291 (\ 98) \\
  & $N=400$ & 1640 (104)  & ---  & ---  & --- & --- \\
  & (a) & ---  & 268 (\ 81) & 254 (\ 65) & 503 (348) & 522 (342) \\
  & (b) & ---  & 350 (\ 68) & 164 (\ 33) & 720 (328) & 358 (135) \\
  & (c) & ---  & 558 (338)  & 161 (\ 85) & 733 (329) & 242 (113) \\
\hline
J & $N=200$ & 1695 (206) & ---  & ---  & --- & --- \\
  & (a) & ---  & 430 (\ 79) & 460 (115) & 760 (284) & 828 (302) \\
  & (b) & ---  & 545 (172) & 272 (104) & 860 (360) & 458 (211) \\
  & (c) & ---  & 535 (155) & 171 (\ 43) & 780 (307) & 264 (110) \\
  & $N=400$ & 1470 (220)  & ---  & ---  & --- & --- \\
  & (a) & ---  & 215 (\ 39) & 230 (\ 58) & 455 (118) & 516 (194) \\
  & (b) & ---  & 305 (107) & 146 (\ 42) & 628 (163) & 317 (\ 86) \\
  & (c) & ---  & 308 (\ 55) & 141 (\ 26) & 738 (202) & 232 (\ 73) \\
\hline
L & $N=200$ & 1925 (134) & ---  & ---  & --- & --- \\
  & (a) & ---  & 405 (\ 64) & 384 (\ 54) & 600 (139) & 776 (280) \\
  & (b) & ---  & 475 (125)  & 202 (\ 50) & 795 (215) & 476 (173) \\
  & (c) & ---  & 580 (149) & 168 (\ 41) & 880 (261) & 329 (109) \\
  & $N=400$ & 1730 (250) & ---  & ---  & --- & --- \\
  & (a) & ---  & 270 (\ 55) & 230 (\ 45) & 445 (158) & 518 (230) \\
  & (b) & ---  & 350 (131) & 144 (\ 51) & 685 (230) & 364 (126) \\
  & (c) & ---  & 350 (\ 91) & \ 97 (\ 25) & 693 (203) & 219 (\ 70) \\
\hline
\hline
\end{tabular}}}
\caption[]
{Simulation~1: Data condensation ratios $\times 10^4$ (standard deviation $\times 10^{4}$). The RSDE column contains the data condensation ratios in the case of RSDE. The numbers in column (I) are the actual number of data points chosen by the algorithm divided by $N$. The numbers in column (II) are the actual number of words in $D_{1}$ chosen by the algorithm divided by $m|B_{1}|$.} \label{tab.c.ratio}
\end{center}
\end{table}
\subsubsection{Simulation~2}

We present the numerical results of $\widehat{f}_{0.5}$ and $\widehat{f}_{1.0}$ in Tables~\ref{tab.results.sim2.beta0.5} and~\ref{tab.results.sim2.beta1.0}, respectively. The visual presentation of the tables is given in Figure~\ref{MISE.for.m.sim2} for Type~C. We observe two general features from the results in Tables~\ref{tab.results.sim2.beta0.5} and~\ref{tab.results.sim2.beta1.0}. One is that (c) $\ge$ (d) $\ge$ (a) in terms of MISE. (Compare each number in Table~\ref{tab.results.sim2.beta0.5} with its counterpart in Table~\ref{tab.results.sim2.beta1.0}. In the case of Type~C of $\widehat{f}_{0.5}$ for $N=400$, observe the two line graphs of (a) and (d) in the upper right panel of Figure~\ref{MISE.for.m.sim2}). This indicates that (d) lies between the best and the worst cases of the proposed estimators in terms of MISE. The second is that (c) $\ge$ (b) $\ge$ (a) in terms of MISE. (Compare each number in Table~\ref{tab.results.sim2.beta0.5} with its counterpart in Table~\ref{tab.results.sim2.beta1.0}. See also the line graphs (a), (b), and (c) in each panel of Figure~\ref{MISE.for.m.sim2} to cite an example of Type~C.) This indicates that MISE can be improved as the percentage of the dictionary data points in the sample size decreases. However, it should be noted that the improvement of MISE caused by a decrease in the ratio of $m/N$ occurs as far as the number of the dictionary data points $m$ is not small, because the algorithm can no longer be executable in such a situation. Our experiments suggest that MISE can deteriorate when $m$ is less than $N/8$. The reason that the two features, (c) $\ge$ (d) $\ge$ (a) and (c) $\ge$ (b) $\ge$ (a) in terms of MISE, are not observed in simulation~1 is that each word in \eqref{B1_Dictionary} has larger inter-sample variance than that in \eqref{B2_Dictionary}. (Compare each SD of MISE in Tables \ref{tab.results.sim1.beta0.5} and \ref{tab.results.sim1.beta1.0} with its counterpart in Tables~\ref{tab.results.sim2.beta0.5} and \ref{tab.results.sim2.beta1.0}, respectively.)


In the same manner of Figure~\ref{results.C.sim1}, we visually present the results of (b) in simulation~2 for Types C, J, and L in Figures~\ref{results.C.sim2}, ~\ref{results.J.sim2}, and \ref{results.L.sim2}, respectively. We find $\widehat{f}_{0.5}$ and $\widehat{f}_{1.0}$ outperform KDE1, KDE2, and RSDE in terms of MISE as $k$ increases in Type C. (See the upper two panels of Figure~\ref{results.C.sim2}.) In comparison with simulation~1, simulation~2 yields the smaller MISE for Types C and L. (Compare each number in Tables~\ref{tab.results.sim1.beta0.5} and \ref{tab.results.sim1.beta1.0} with its counterpart in Tables~\ref{tab.results.sim2.beta0.5} and \ref{tab.results.sim2.beta1.0}, respectively.)
Observing the contour plots in the same figures, we find in Types J and L that $\widehat{f}_{1.0}$ captures the shape of the true contour plot better than $\widehat{f}_{0.5}$. (See the middle and bottom panels of Figures~\ref{results.J.sim2} and \ref{results.L.sim2}.) We consider this difference could be the result of the robustness property of the $\beta$-power divergence function in the case of $\beta = 0.5$.

We describe the data points chosen by our method for estimation from the dictionary. From the contour plots of $\widehat{f}_{1.0}$ in the lower two panels of Figure~\ref{results.J.sim2}, we find that our algorithm generally chooses data points in the dictionary along with the mountain ridges of the contour plots. This tendency is also observed in RSDE (see Girolami and He 2003, p.1256).
\begin{table}
\begin{center}
{\scriptsize{
\begin{tabular}{cllllllll}
\hline
\hline
$k+1$ & 1 & 25 & 50 & 75 & 100 & KDE1 & KDE2 & RSDE \\
\hline
Type C & & & & & & & \\
$\underline{N=200}$ & --- & --- & --- & --- & --- &  84(26) & 81(28) & 106(31) \\
$(a)$ & 89 (37) & 47 (19) & 45 (18) & 44 (18) & 42 (17) \\
$(b)$ & 114 (56) &  58 (22) & 54 (24) & 52 (23) & 52 (24) \\
$(c)$ & 108 (55) & 70 (45) & 63 (39) & 67 (41) & 64 (37) \\
$(d)$ & 90 (33) & 57 (26) & 53 (31) & 52 (29) & 53 (30) \\
$(e)$ & 80 (19) & 43 (14) & 39 (15) & 38 (15) & 37 (16) \\
$\underline{N=400}$ & --- & --- & --- & --- & --- & 53(14) & 54(11) & 84(18) \\
$(a)$ & 82 (26) & 40 (13) & 38 (11) & 39 (13) & 38 (12) \\
$(b)$ & 79 (23) & 46 (19) & 43 (19) & 42 (20) & 41 (19) \\
$(c)$ & 95 (41) & 63 (29) & 62 (33) & 62 (35) & 62 (34) \\
$(d)$ & 80 (19) & 43 (14) & 39 (15) & 38 (15) & 37 (16) \\
$(e)$ & 65 (9) & 38 (12) & 35 (12) & 34 (12) & 34 (12) \\
\hline
Type J & & & & & & & \\
$\underline{N=200}$ & --- & --- & --- & --- & --- &  108(17) & 118(30) & 138(33) \\
$(a)$ & 322 (14) & 299 (12) & 300 (16) & 299 (17) & 299 (17) \\
$(b)$ & 333 (35) & 310 (25) & 311 (34) & 311 (34) & 310 (33) \\
$(c)$ & 353 (74) & 343 (69) & 339 (62) & 342 (65) & 341 (61) \\
$(d)$ & 341 (38) & 318 (26) & 319 (26) & 321 (28) & 320 (28) \\
$(e)$ & 308 (11) & 301 (15) & 297 (16) & 297 (18) & 296 (17) \\
$\underline{N=400}$ & --- & --- & --- & --- & --- & 74(10) & 80(19) & 111(19) \\
$(a)$ & 311 (14) & 298 (14) & 295 (17) & 294 (17) & 293 (17) \\
$(b)$ & 312 (12) & 299 (16) & 296 (15) & 295 (17) & 294 (16) \\
$(c)$& 315 (25) & 310 (31) & 306 (27) & 307 (27) & 307 (27) \\
$(d)$ & 308 (11) & 301 (15) & 297 (16) & 297 (18) & 296 (17)  \\
$(e)$ & 304 (11) & 294 (12) & 291 (12) & 289 (12) & 288 (13) \\
\hline
Type L & & & & & & \\
$\underline{N=200}$ & --- & --- & --- & --- & --- & 67(14) & 77(14) & 131(87) \\
$(a)$ & 968 (111) & 165 (53) & 175 (47) & 174 (50)  & 174 (52)  \\
$(b)$ & 1247 (115) & 237 (89) & 225 (59)  & 229 (70) & 225 (66)  \\
$(c)$ & 1452 (99) & 397 (211)  & 398 (183) & 400 (193)  & 412 (198)  \\
$(d)$ & 1218 (98) & 233 (70) & 242 (67)  & 259 (82)  & 245 (68)  \\
$(e)$ & 1567 (112)  & 244 (83) & 243 (79) & 236 (82) & 237 (86) \\
$\underline{N=400}$ & --- & --- & --- & --- & --- & 45(6) & 54(18) & 98(26) \\
$(a)$ & 1212 (106) & 180 (34) & 188 (45) & 178 (32) & 182 (29)  \\
$(b)$ & 1565 (111) & 232 (53) & 247 (71) & 240 (75) & 236 (75)  \\
$(c)$ & 1857 (139) & 350 (84) & 345 (91) & 342 (92) & 343 (89)  \\
$(d)$ & 1567 (112) & 244 (83) & 243 (79) & 236 (82) & 237 (86)  \\
$(e)$ & 2067 (85)  & 256 (40) & 270 (55) & 266 (57) & 260 (60)  \\
\hline
\hline
\end{tabular}}}
\caption[]
{Simulation~2: Result of MISE $\times 10^{4}$ (standard deviation $\times 10^{4}$). ($\beta=0.5$)} \label{tab.results.sim2.beta0.5}
\end{center}
\end{table}
\begin{table}
\begin{center}
{\scriptsize{
\begin{tabular}{cllllllll}
\hline
\hline
$k+1$ & 1 & 25 & 50 & 75 & 100 & KDE1 & KDE2 & RSDE \\
\hline
Type C &  &  &  &  &  &  \\
$\underline{N=200}$ & --- & --- & --- & --- & --- & 84(26) & 81(28) & 106(31) \\
$(a)$ & 92 (31) & 43 (12) & 42 (13) & 42 (13) & 42 (13) \\
$(b)$ & 125 (58) & 66 (26) & 66 (22) & 64 (21) & 64 (22) \\
$(c)$ & 120 (53) & 68 (40) & 68 (35) & 72 (34) & 71 (37) \\
$(d)$ & 101 (33) & 55 (26) & 56 (29) & 55 (29) & 55 (28) \\
$(e)$ & 73 (12) & 35 (15) & 33 (15) & 34 (15) & 34 (14) \\
$\underline{N=400}$ & --- & --- & --- & --- & --- & 53(14) & 54(11) & 84(18) \\
$(a)$ & 88 (27) & 32 (13) & 27 (10) & 28 (9) & 29 (10) \\
$(b)$ & 78 (27) & 35 (18) & 30 (12) & 30 (14) & 30 (13) \\
$(c)$ & 104 (40) & 56 (32) & 58 (31) & 57 (31) & 55 (30) \\
$(d)$ & 73 (12) & 35 (15) & 33 (15) & 34 (15) & 34 (14) \\
$(e)$ & 66 (18) & 25 (10) & 23 (9) & 23 (8) & 23 (8) \\
\hline
Type J & & & & & & \\
$\underline{N=200}$ & --- & --- & --- & --- & --- & 108(17) & 118(31) & 138(33) \\
$(a)$ & 331 (39) & 259 (22) & 255 (20) & 2541 (20) & 254 (22) \\
$(b)$ & 331 (37) & 266 (18) & 266 (28) & 262 (19) & 264 (22) \\
$(c)$& 381 (1431) & 302 (22) & 305 (30) & 304 (33) & 305 (33) \\
$(d)$ & 349 (60) & 280 (27) & 270 (28) & 265 (27) & 267 (29) \\
$(e)$ & 307 (12) & 2508 (23) & 244 (22) & 242 (23) & 242 (25) \\
$\underline{N=400}$ & --- & --- & --- & --- & --- & 74(10) & 80(19) & 111(19) \\
$(a)$ & 322 (21) & 242 (15) & 233 (14) & 233 (17) & 231 (17) \\
$(b)$ & 317 (18) & 250 (20) & 242 (19) & 241 (21) & 241 (20) \\
$(c)$ & 328 (49) & 260 (22) & 252 (21) & 254 (21) & 252 (20) \\
$(d)$ & 309 (12) & 251 (23) & 244 (22) & 242 (24) & 242 (25) \\
$(e)$ & 303 (11) & 231 (8) & 222 (9) & 221 (10) & 220 (10) \\
\hline
Type L & & & & & & \\
$\underline{N=200}$ & --- & --- & --- & --- & --- & 67(14) & 77(14) & 131(87) \\
$(a)$ & 925 (134) & 91 (18) & 82 (28) & 82 (24) & 81 (22) \\
$(b)$ & 1210 (105) & 129 (29) & 118 (30) & 118 (30) & 116 (30) \\
$(c)$ & 1508 (194) & 208 (86) & 220 (91) & 216 (87) & 212 (82) \\
$(d)$ & 1172 (80) & 139 (39) & 136 (42) & 124 (35) & 127 (36) \\
$(e)$ & 1558 (115) & 94 (23) & 90 (25) & 93 (25) & 94 (23) \\
$\underline{N=400}$ & --- & --- & --- & --- & --- & 45(6) & 54(18) & 98(26) \\
$(a)$ & 1144 (51) & 77 (20) & 61 (16) & 54 (14) & 54 (12) \\
$(b)$ & 1536 (126) & 91 (32) & 89 (26) & 83 (22) & 85 (26) \\
$(c)$ & 1823 (109) & 161 (55) & 150 (30) & 157 (32) & 158 (29) \\
$(d)$ & 1558 (115) & 94 (23) & 90 (25) & 93 (25) & 94 (23) \\
$(e)$ & 2047 (72) & 59 (13) & 49 (12) & 56 (12) & 57 (15) \\
\hline
\hline
\end{tabular}}}
\caption[]
{Simulation~2: Result of MISE $\times 10^{4}$ (standard deviation $\times 10^{4}$).($\beta=1.0$)} \label{tab.results.sim2.beta1.0}
\end{center}
\end{table}
\begin{figure}
\includegraphics*[width=0.49\linewidth]{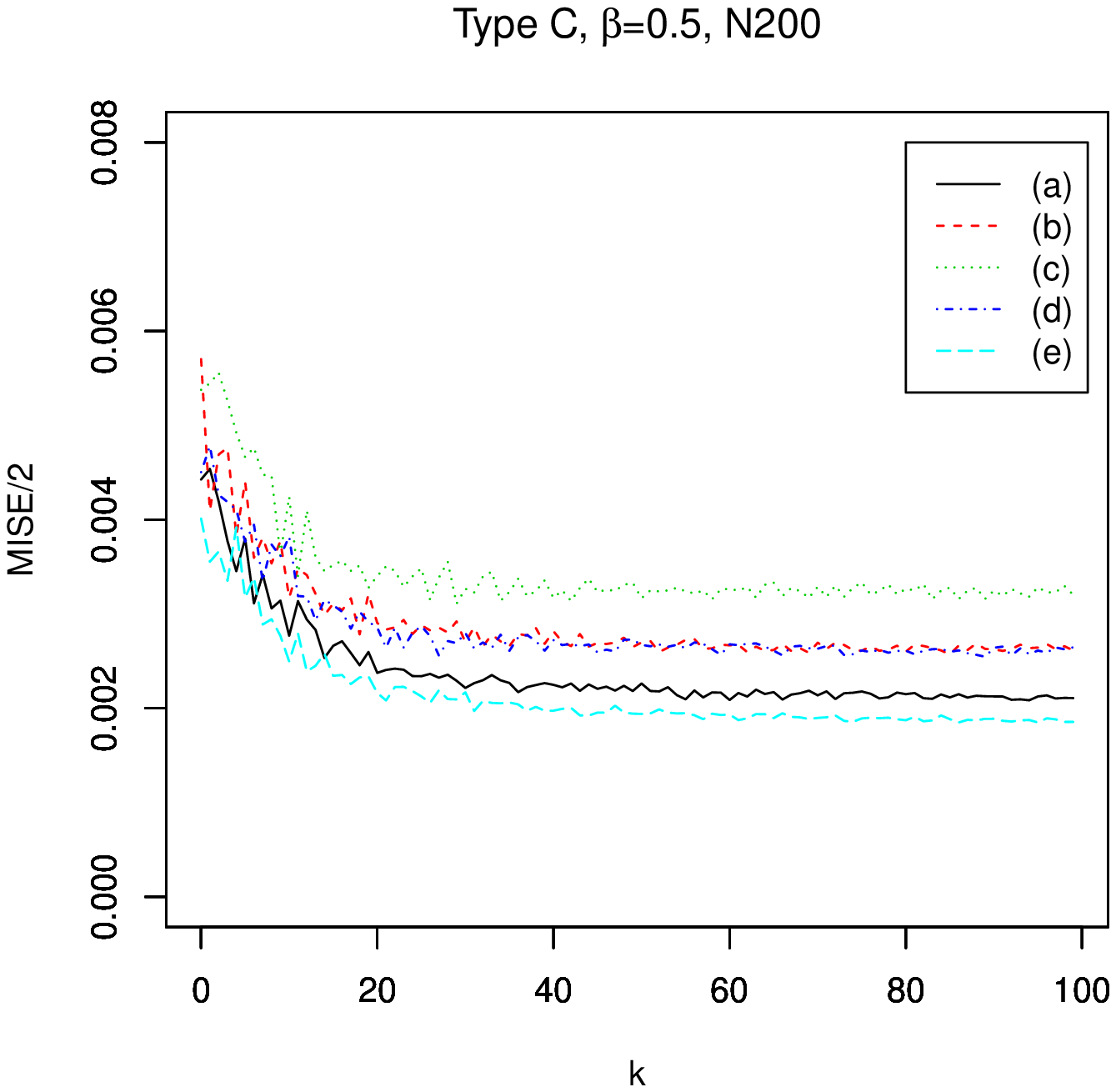}
\includegraphics*[width=0.49\linewidth]{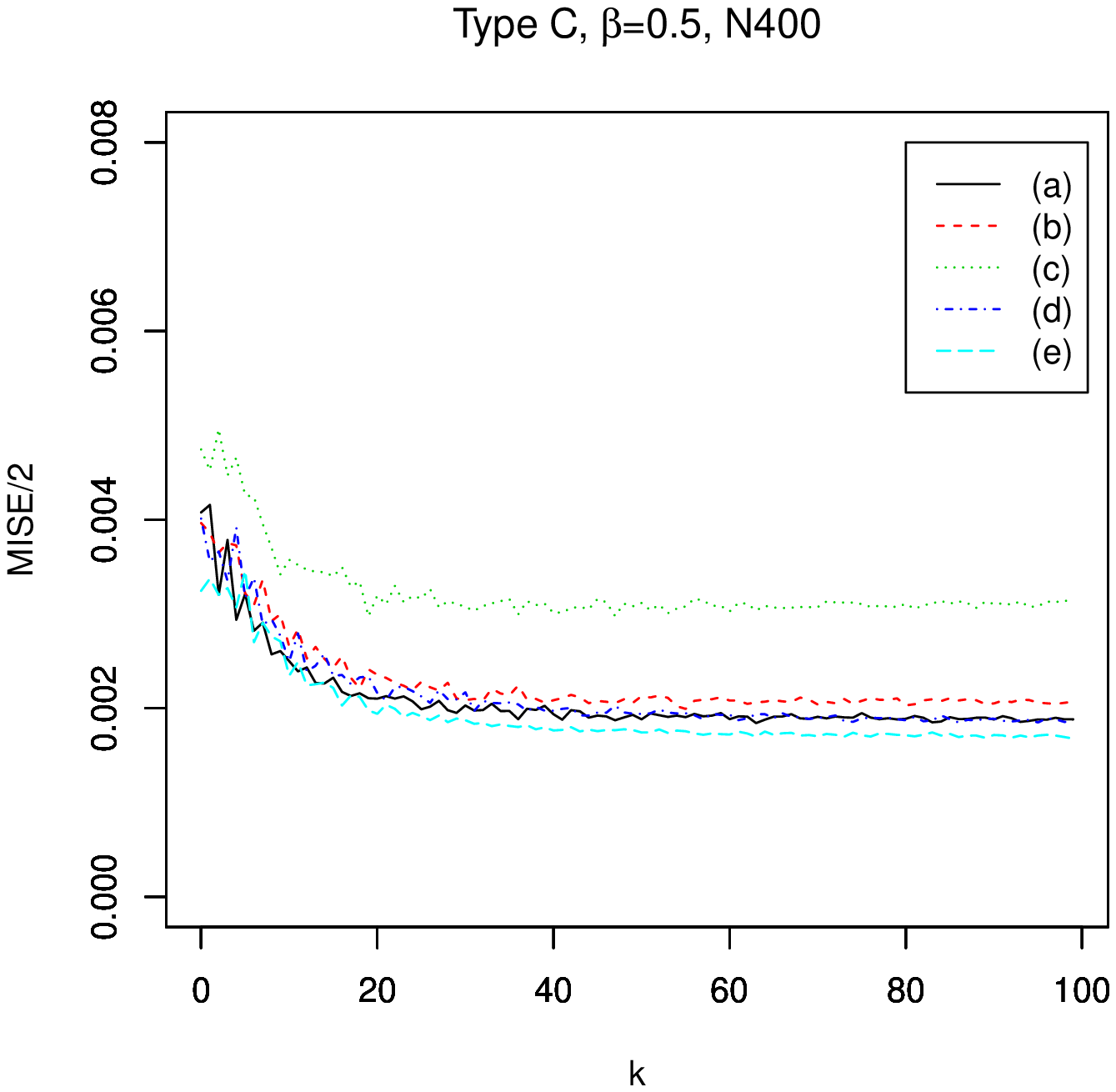}\\
\includegraphics*[width=0.49\linewidth]{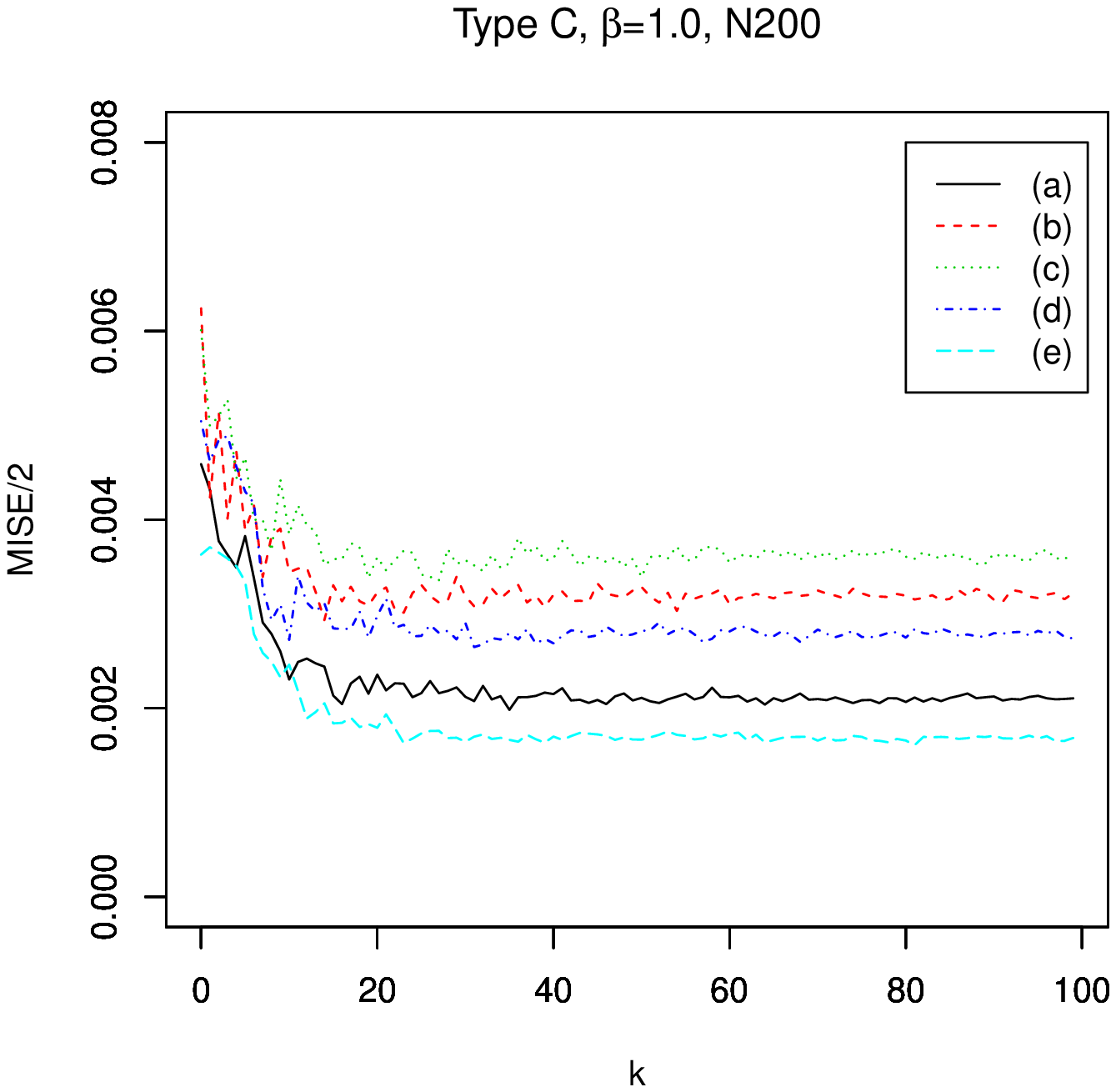}
\includegraphics*[width=0.49\linewidth]{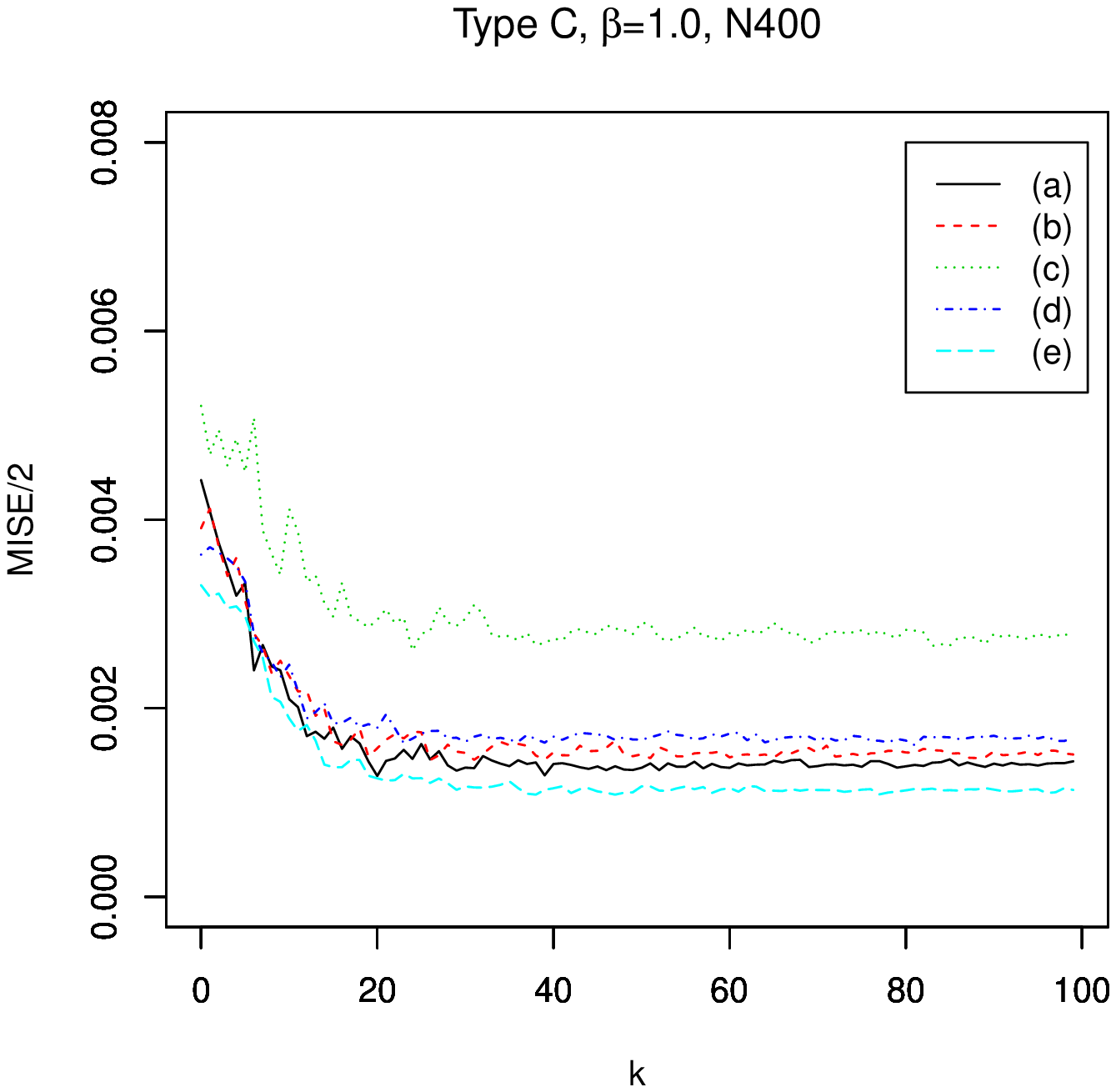}
\caption[]{Simulation~2: Plot of MISEs at each stage of the algorithm in the case of Type~C for different $m/N$.} \label{MISE.for.m.sim2}
\end{figure}
\begin{figure}[htpb]
\begin{center}
\includegraphics*[width=0.4\linewidth]{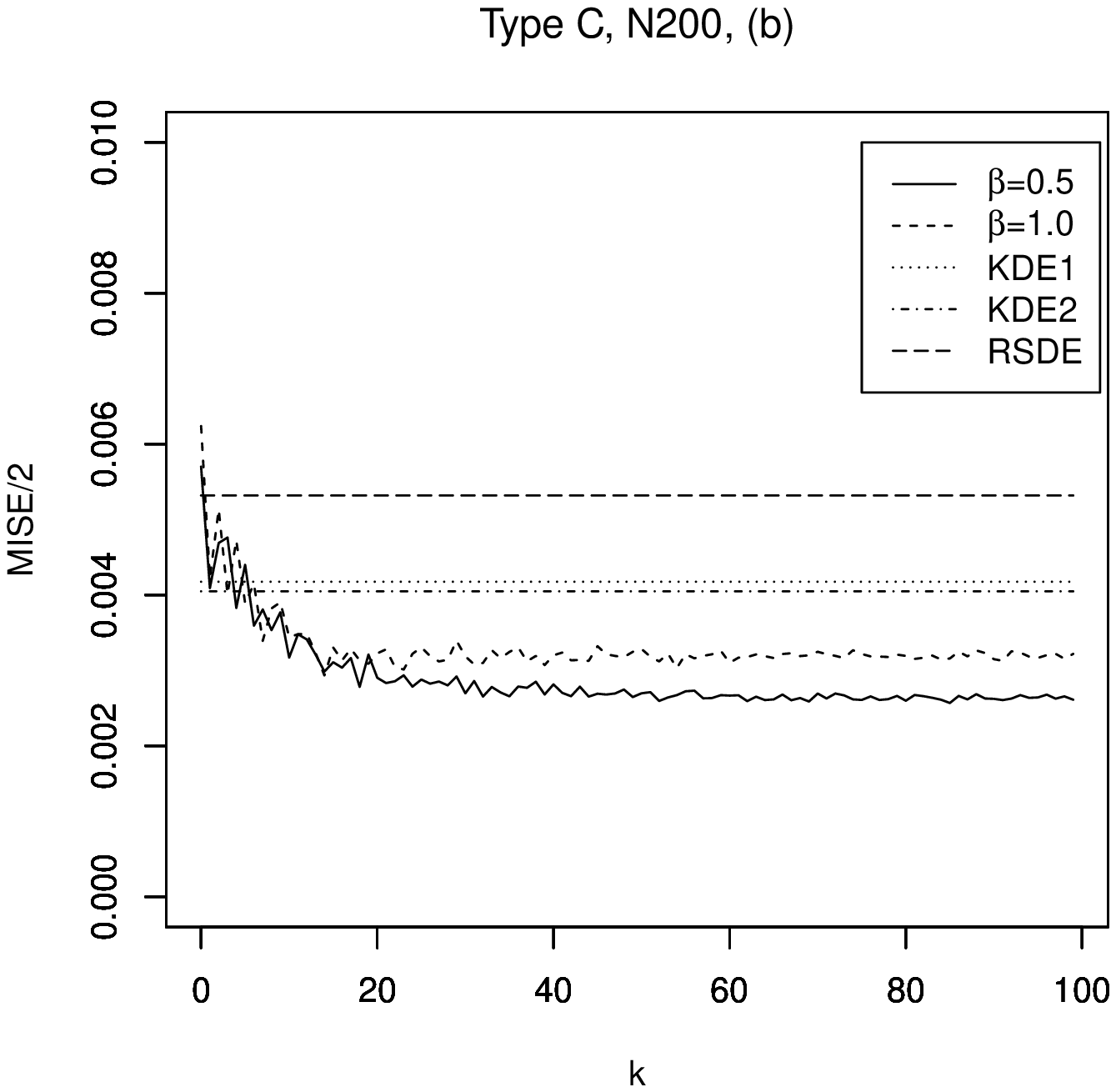}
\includegraphics*[width=0.4\linewidth]{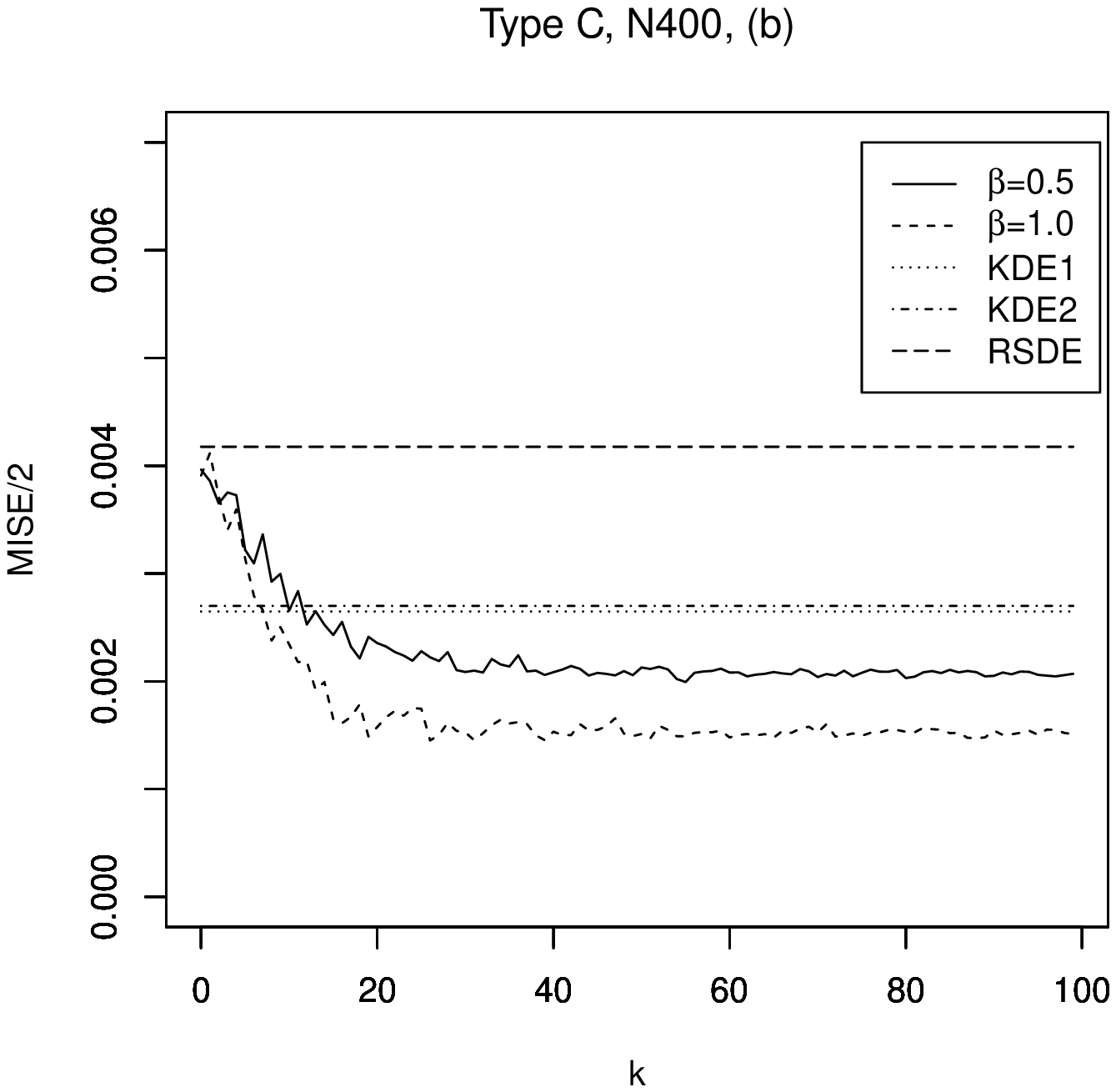}
\includegraphics*[width=0.4\linewidth]{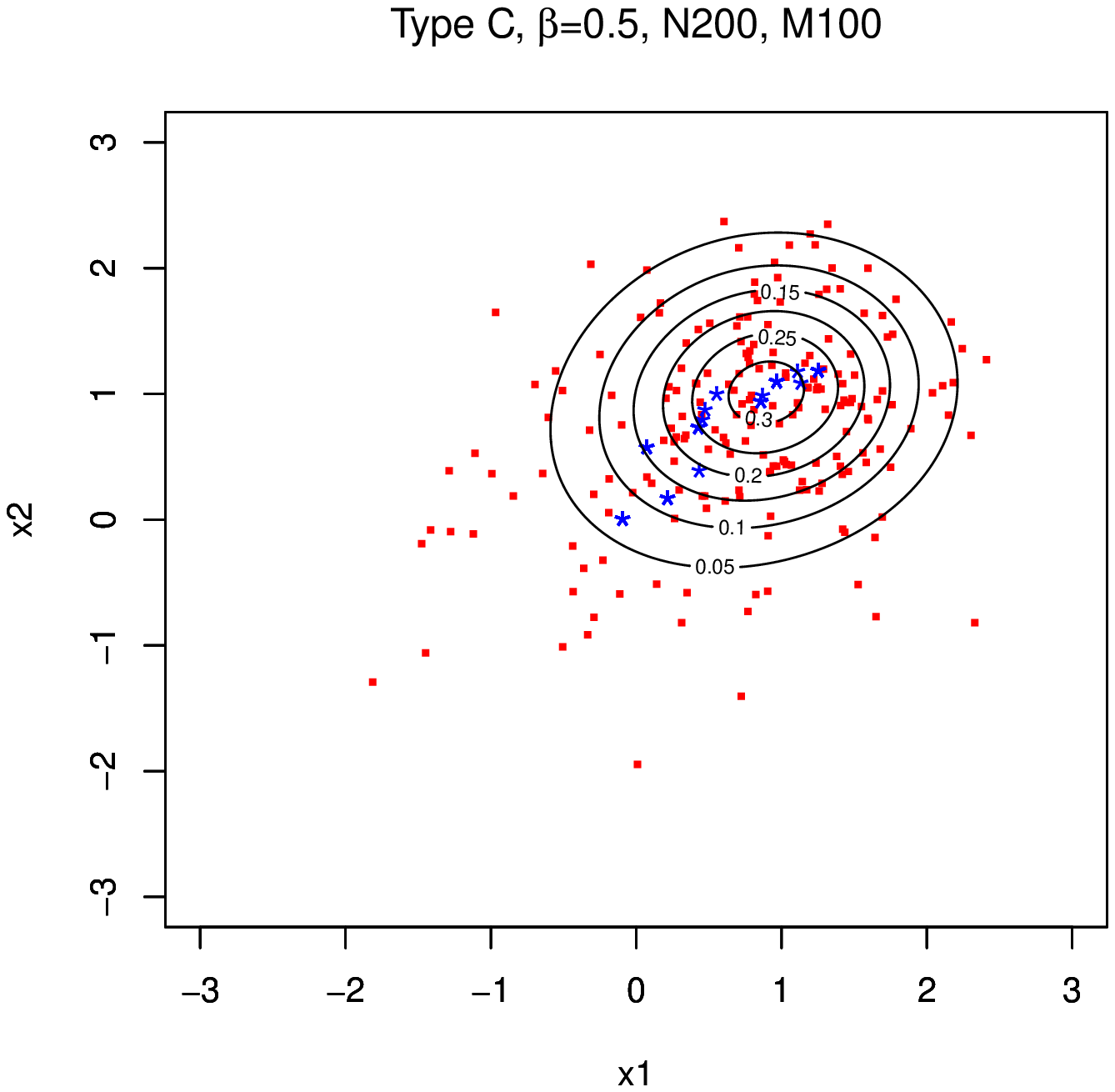}
\includegraphics*[width=0.4\linewidth]{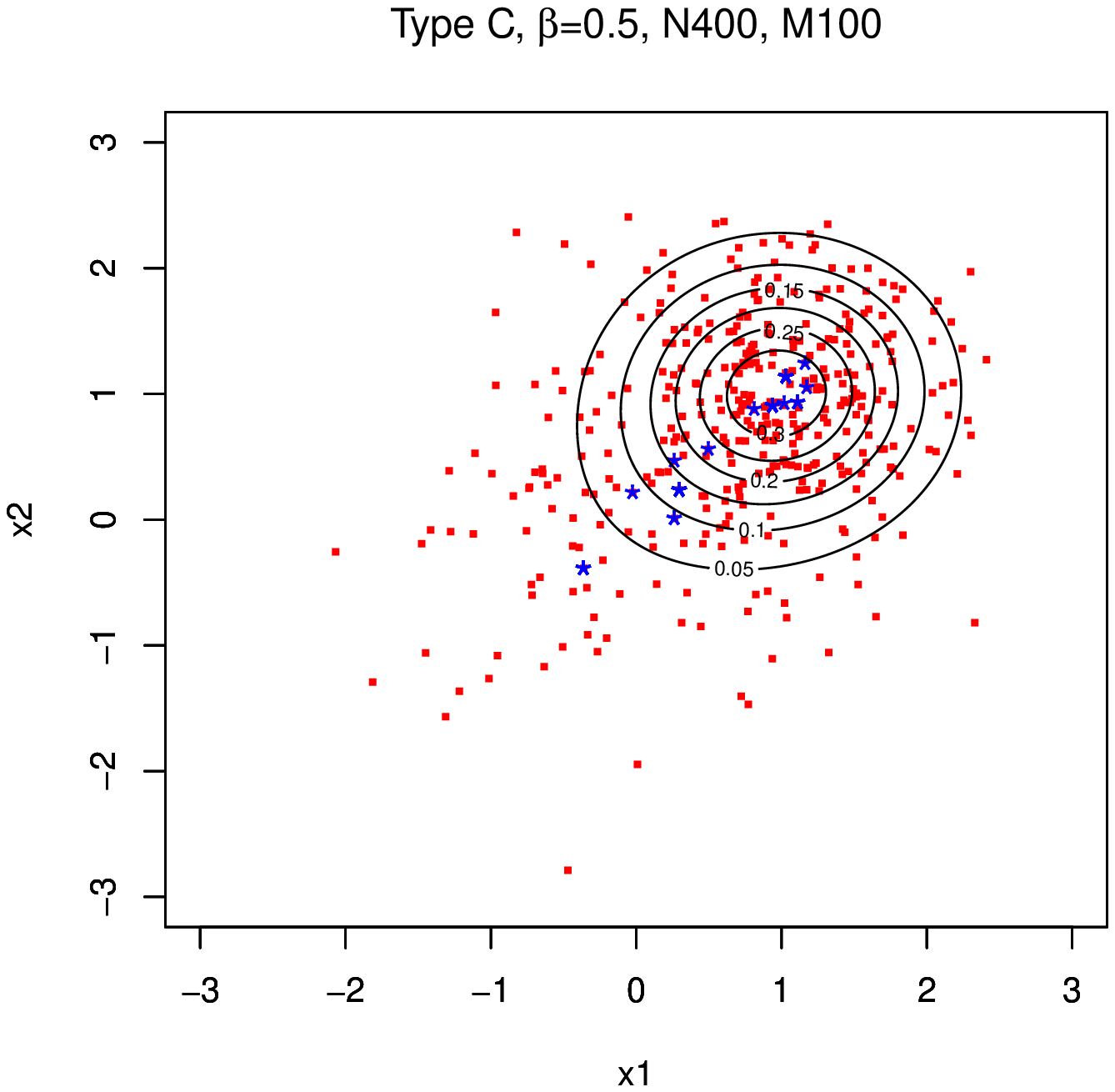}
\includegraphics*[width=0.4\linewidth]{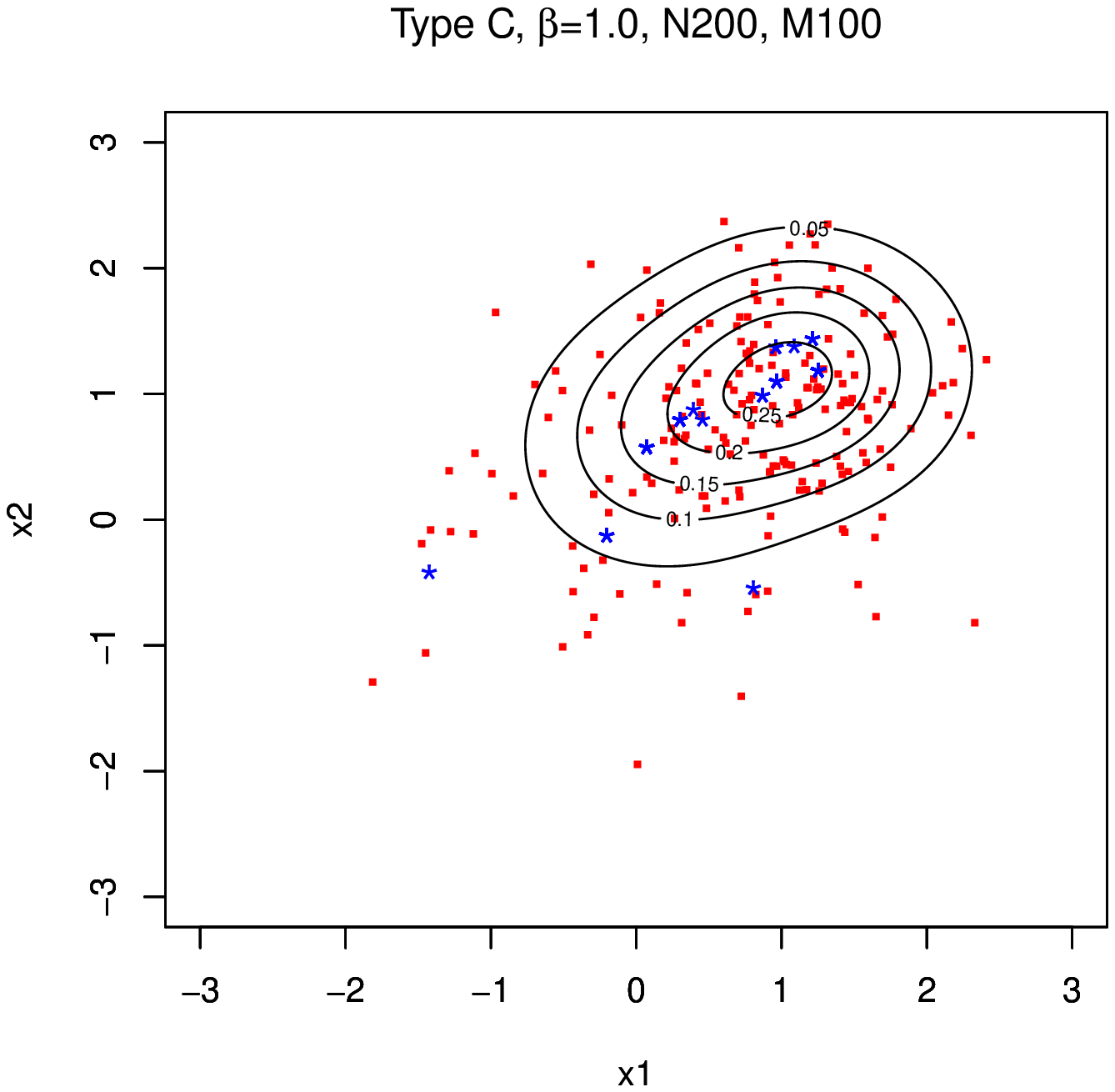}
\includegraphics*[width=0.4\linewidth]{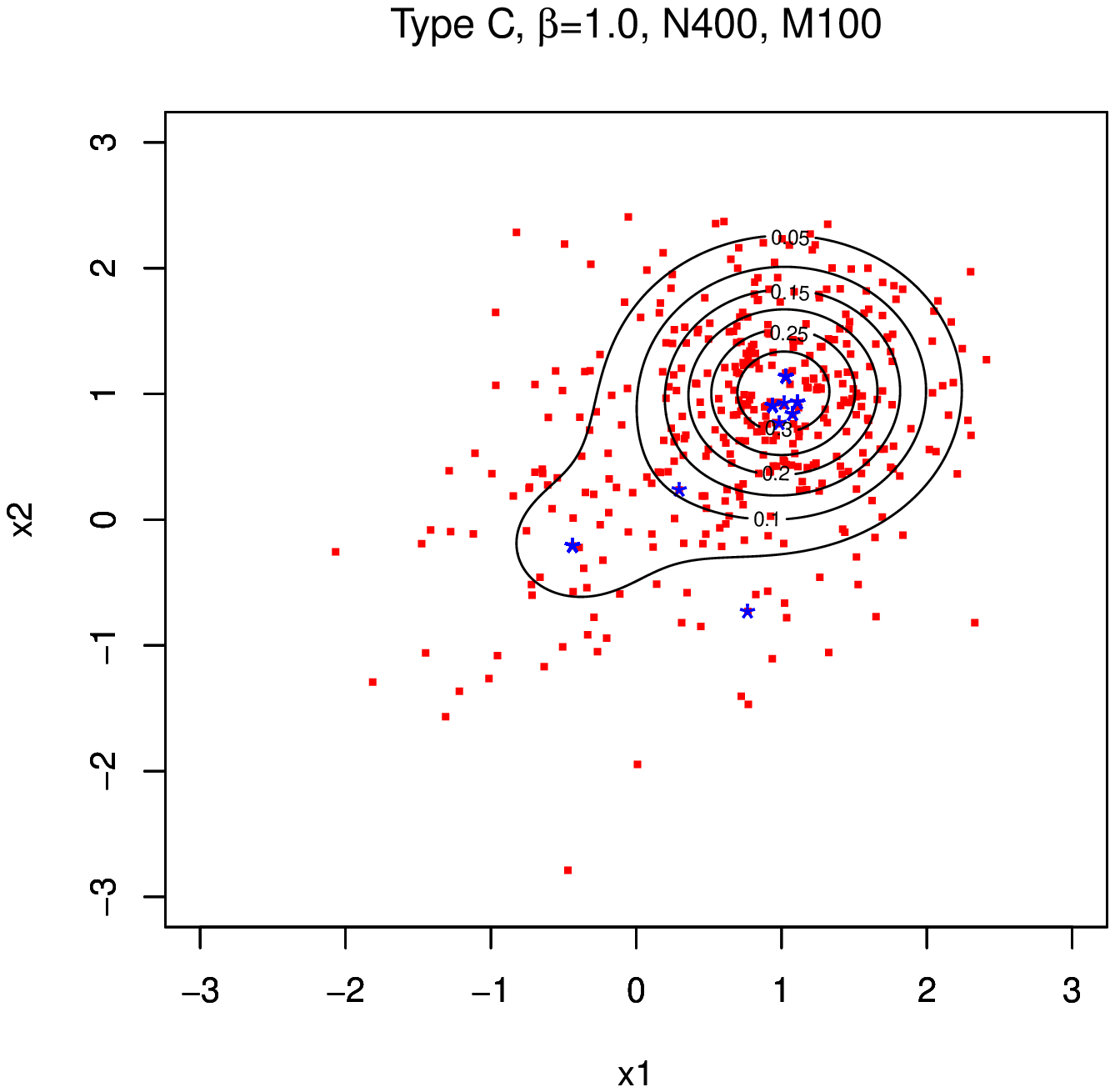}
\end{center}
\caption[]{Simulation~2: Upper: Plots of MISE vs $k$.\ \ \  Middle: Contour plots of $\widehat{f}_{0.5}$.\ \ \ Bottom: Contour plots of $\widehat{f}_{1.0}$} \label{results.C.sim2}
\end{figure}
\begin{figure}[htpb]
\begin{center}
\includegraphics*[width=0.4\linewidth]{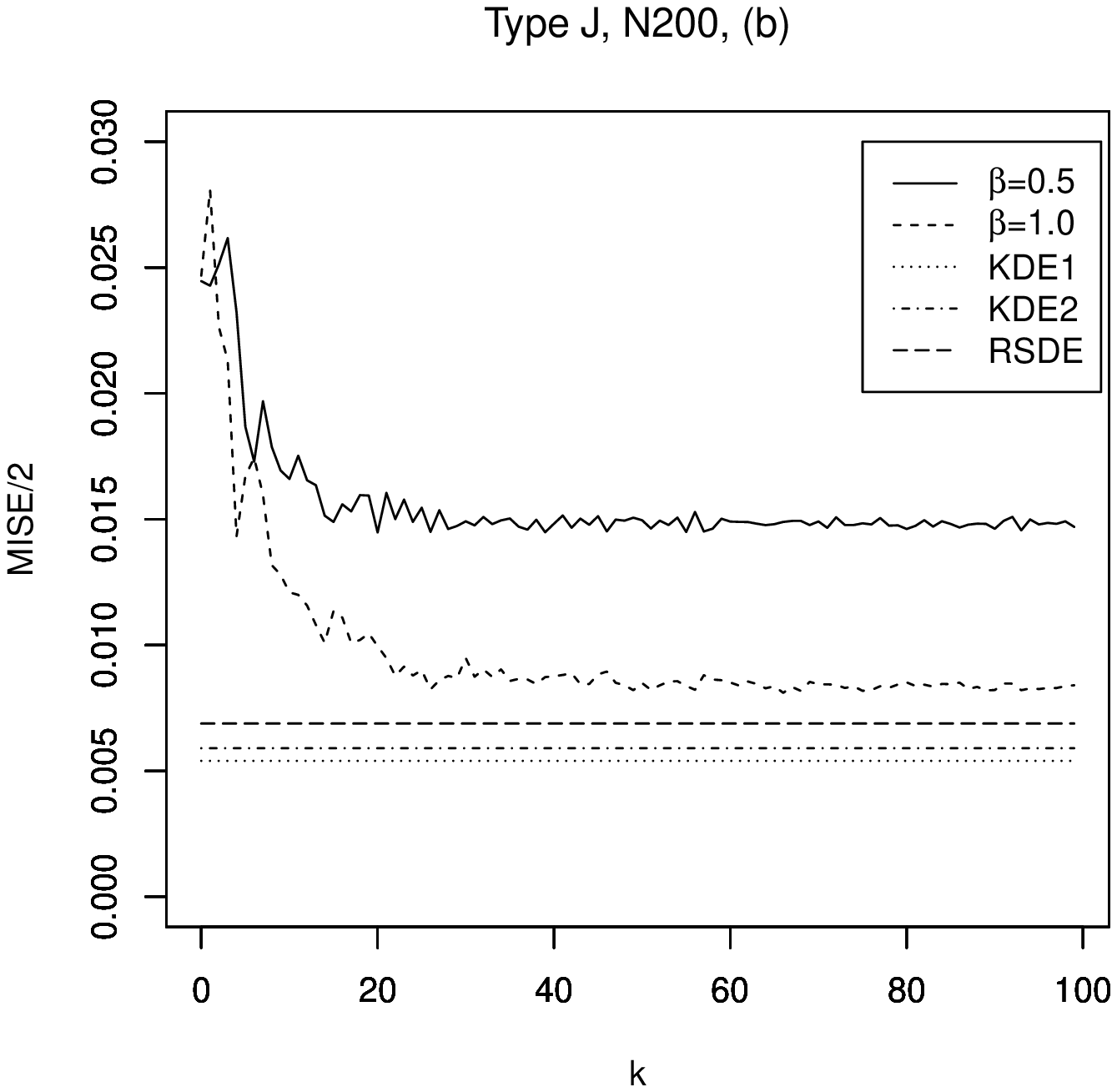}
\includegraphics*[width=0.4\linewidth]{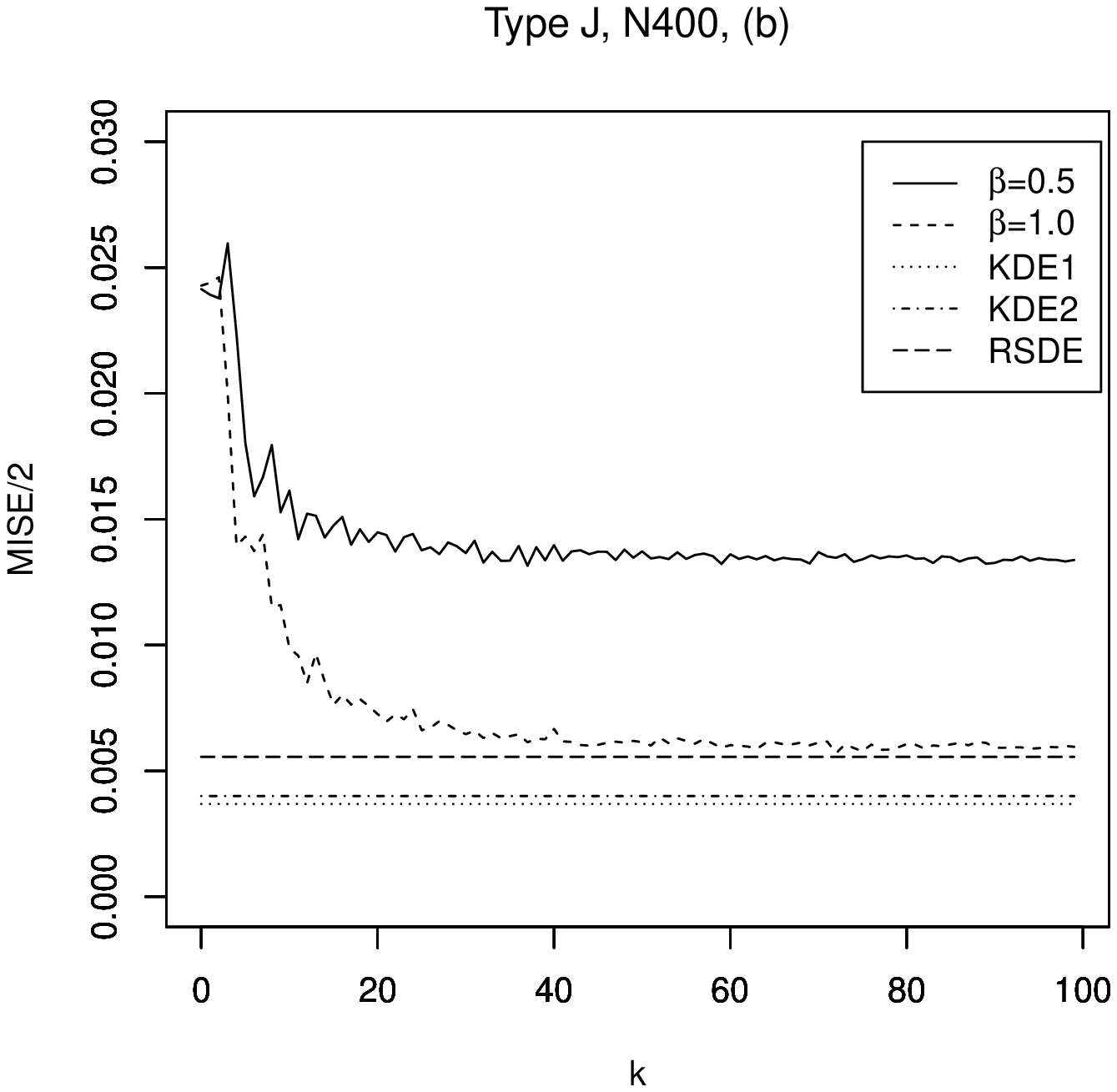}
\includegraphics*[width=0.4\linewidth]{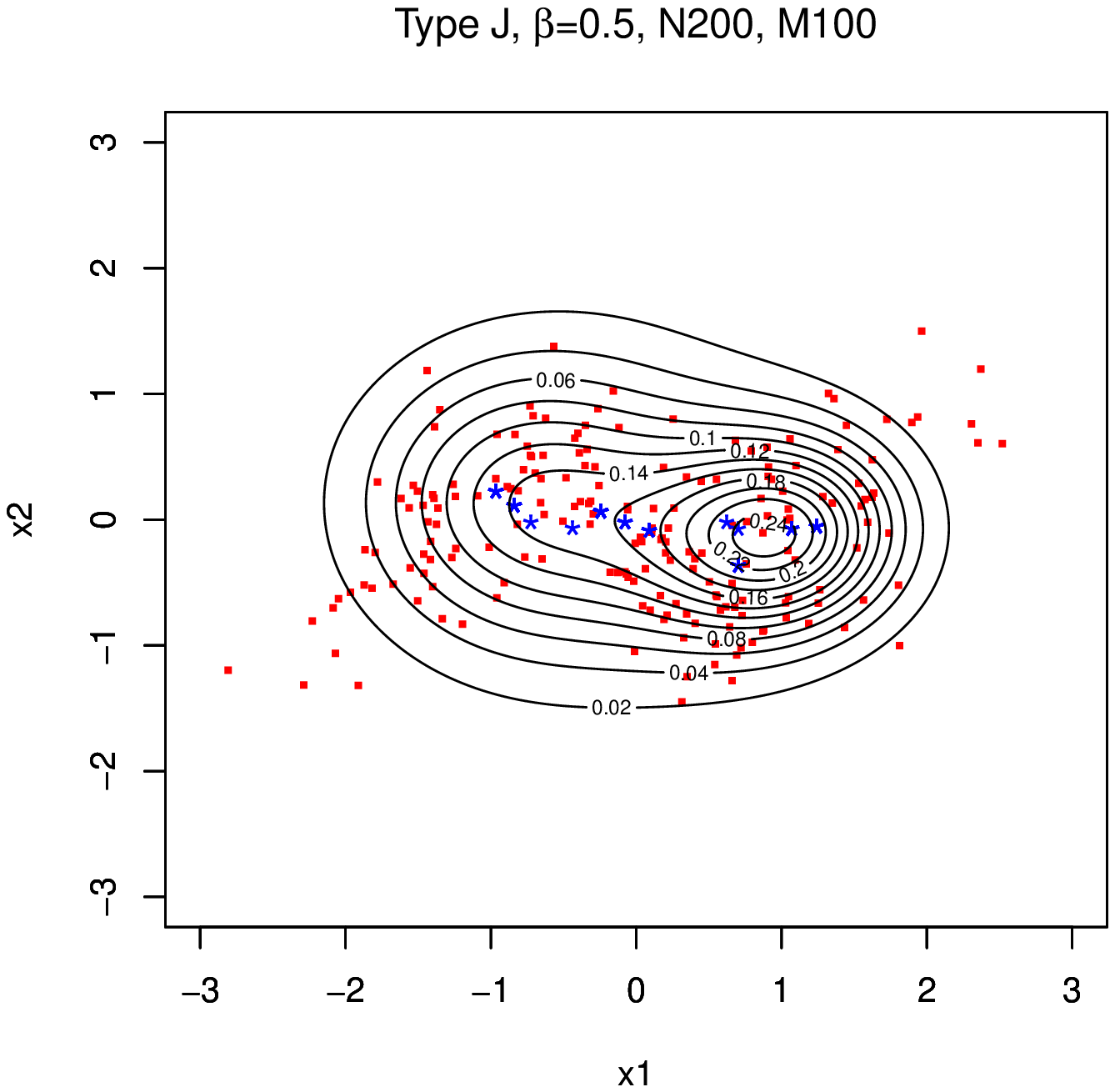}
\includegraphics*[width=0.4\linewidth]{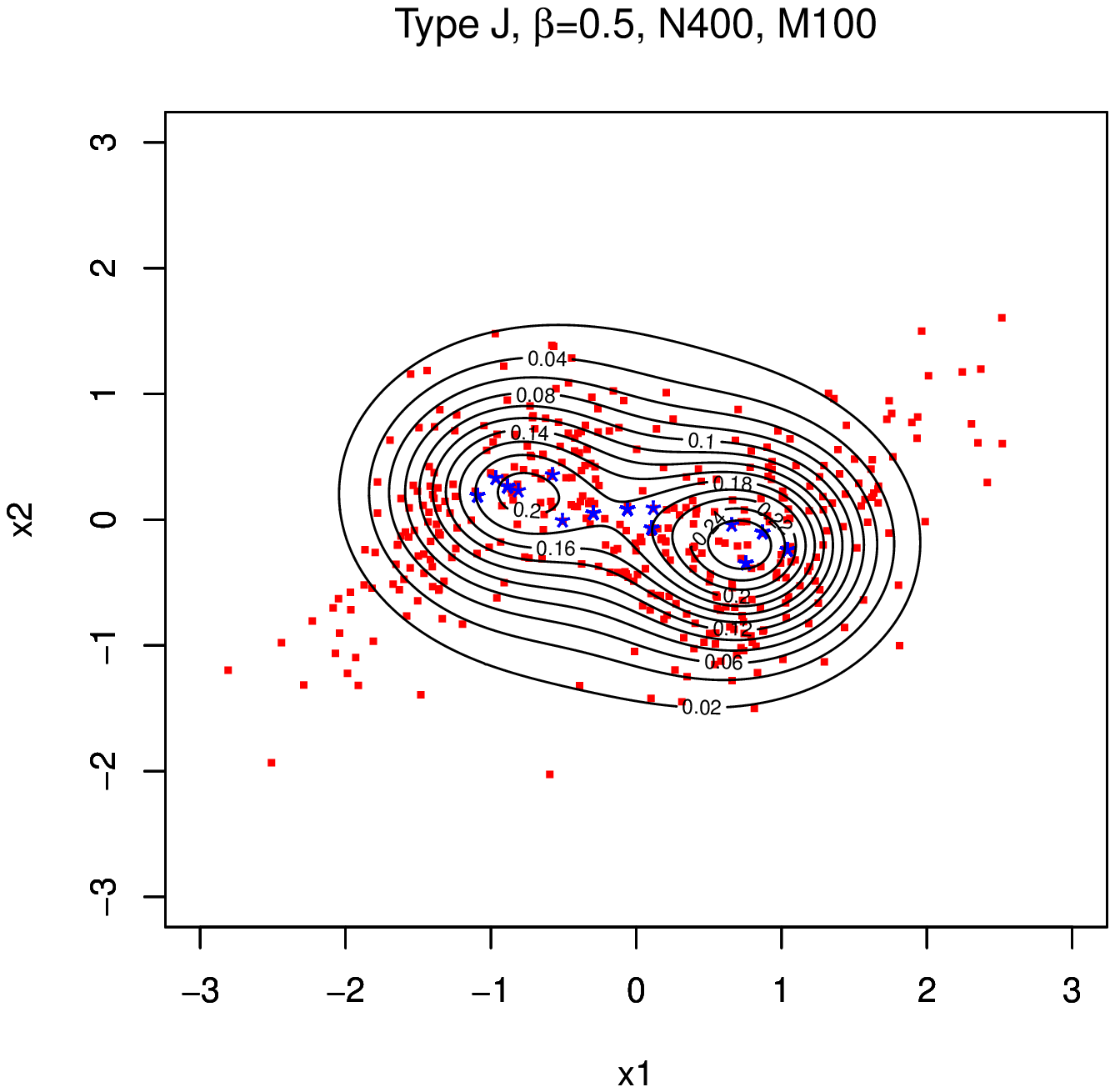}
\includegraphics*[width=0.4\linewidth]{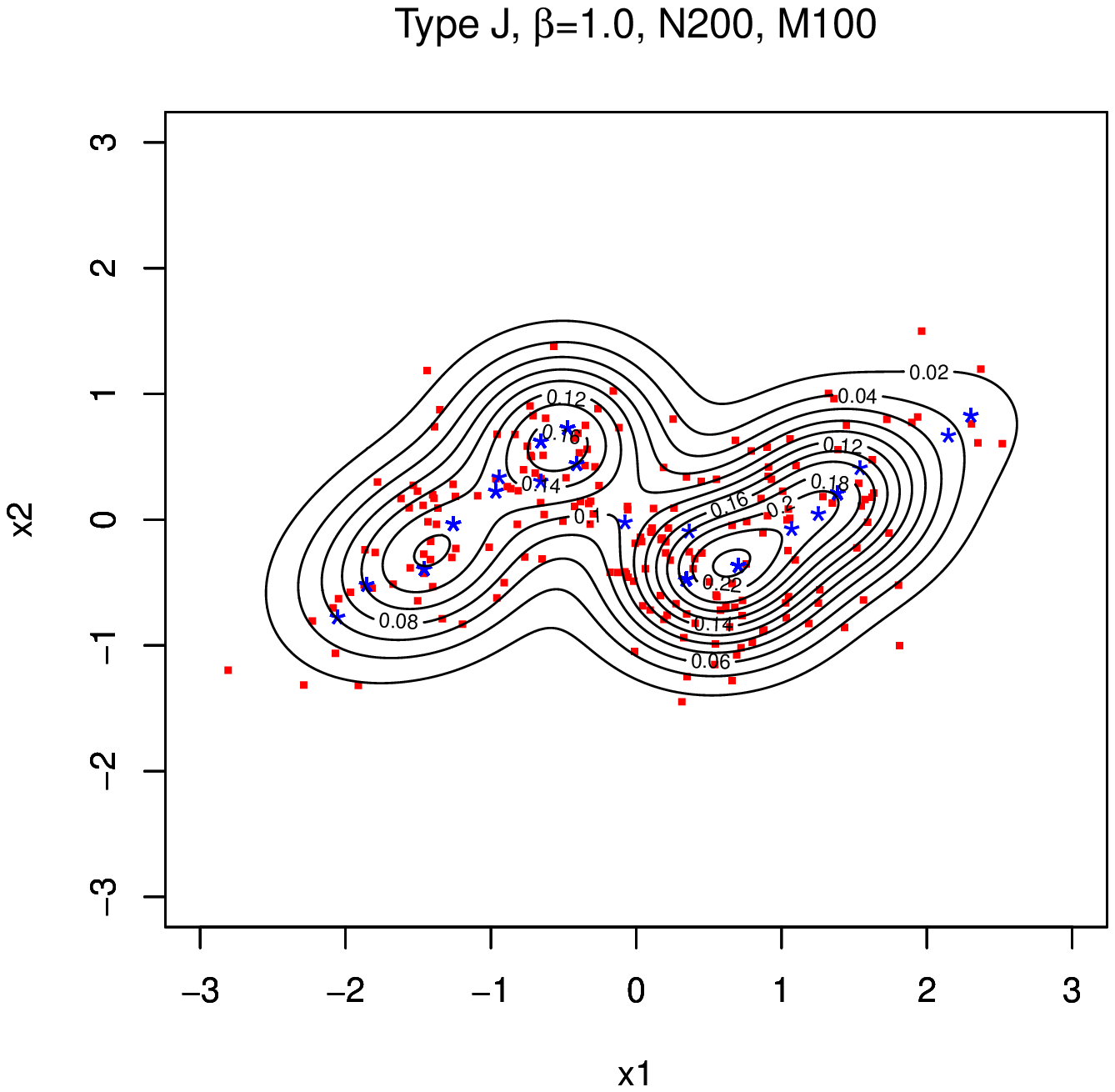}
\includegraphics*[width=0.4\linewidth]{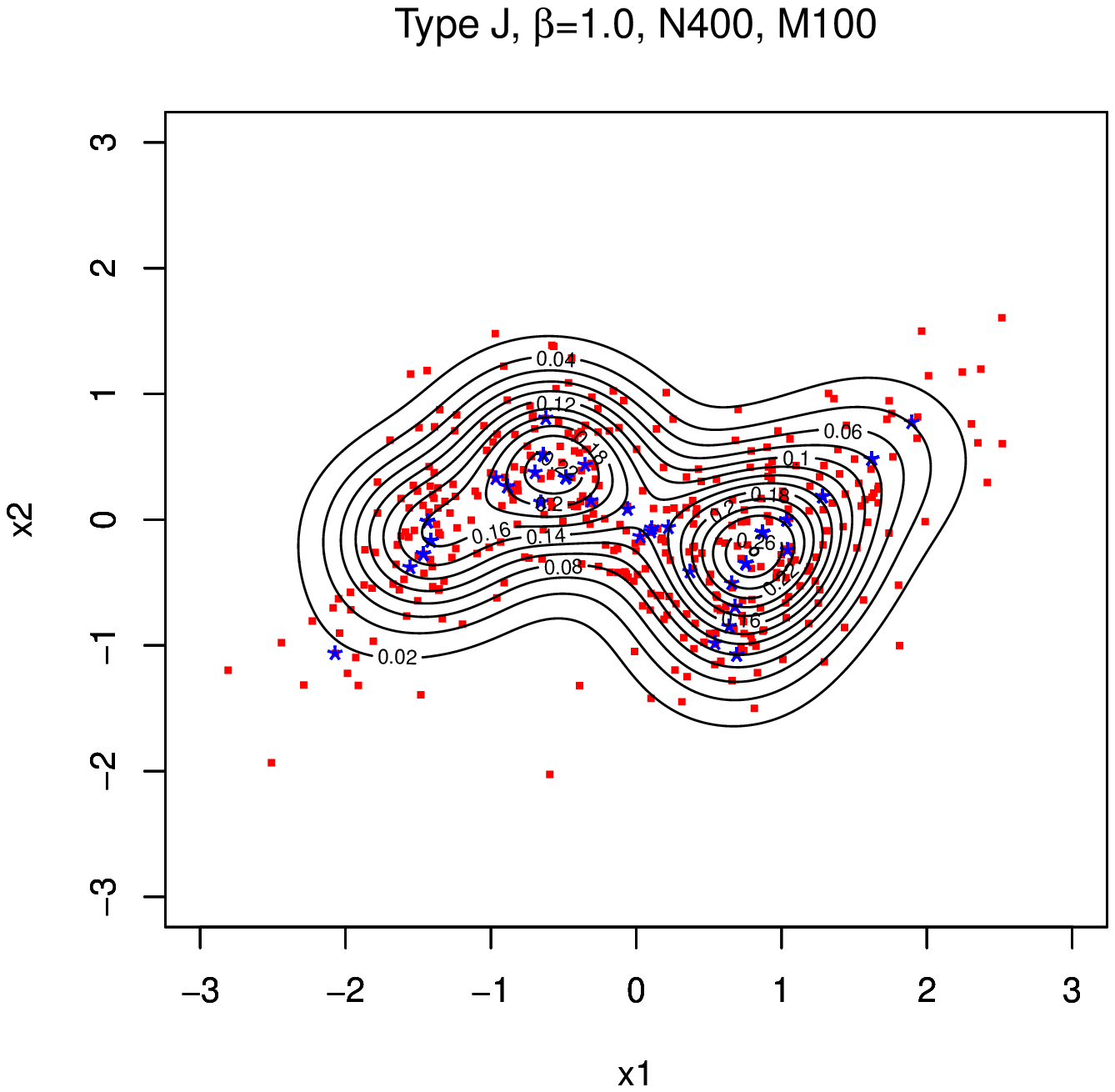}
\end{center}
\caption[]{Simulation~2: Upper: Plots of MISE vs $k$.\ \ \  Middle: Contour plots ($\beta=0.5$).\ \ \ Bottom: Contour plots ($\beta=1.0$)} \label{results.J.sim2}
\end{figure}
\begin{figure}[htpb]
\begin{center}
\includegraphics*[width=0.4\linewidth]{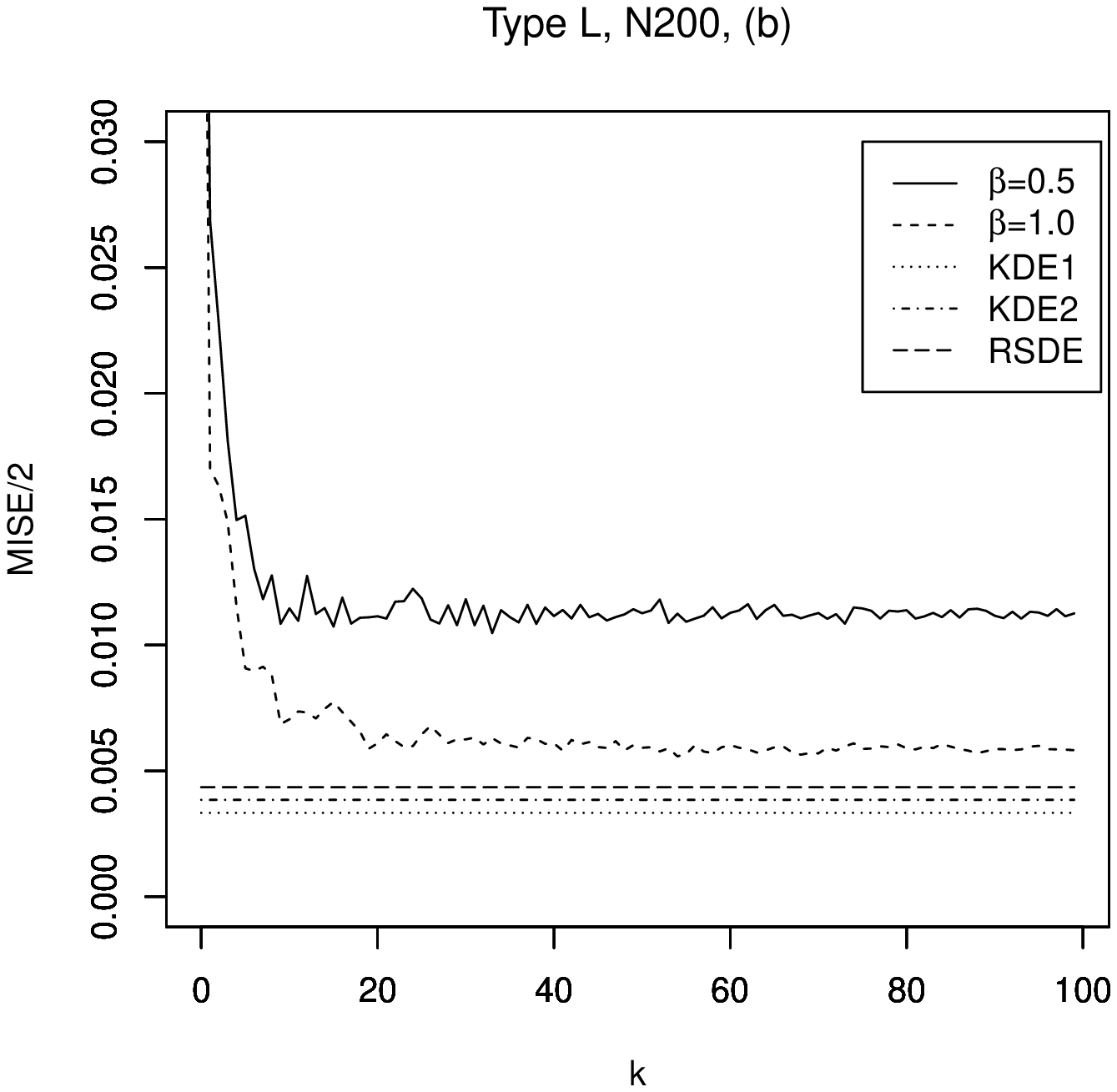}
\includegraphics*[width=0.4\linewidth]{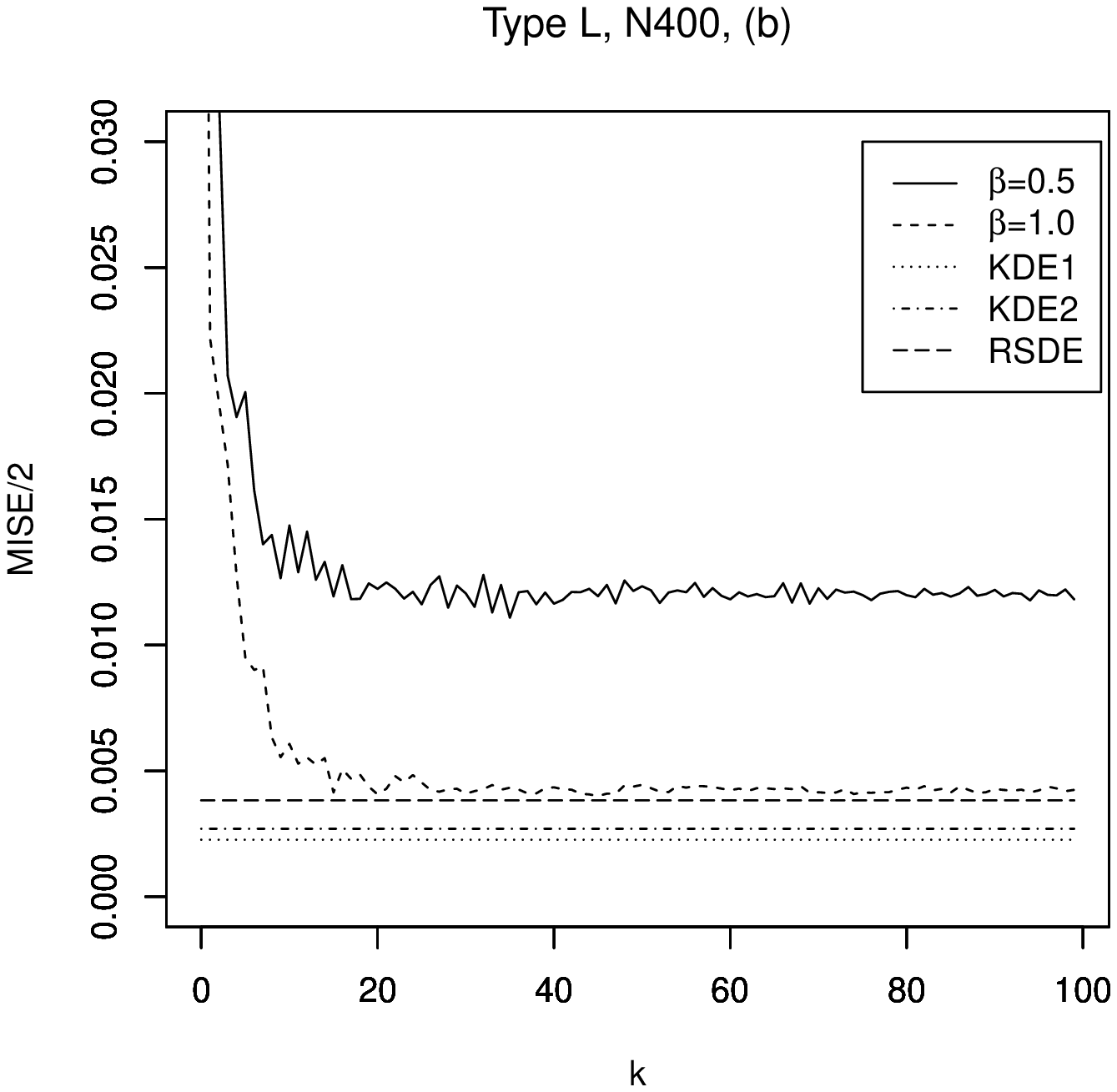}
\includegraphics*[width=0.4\linewidth]{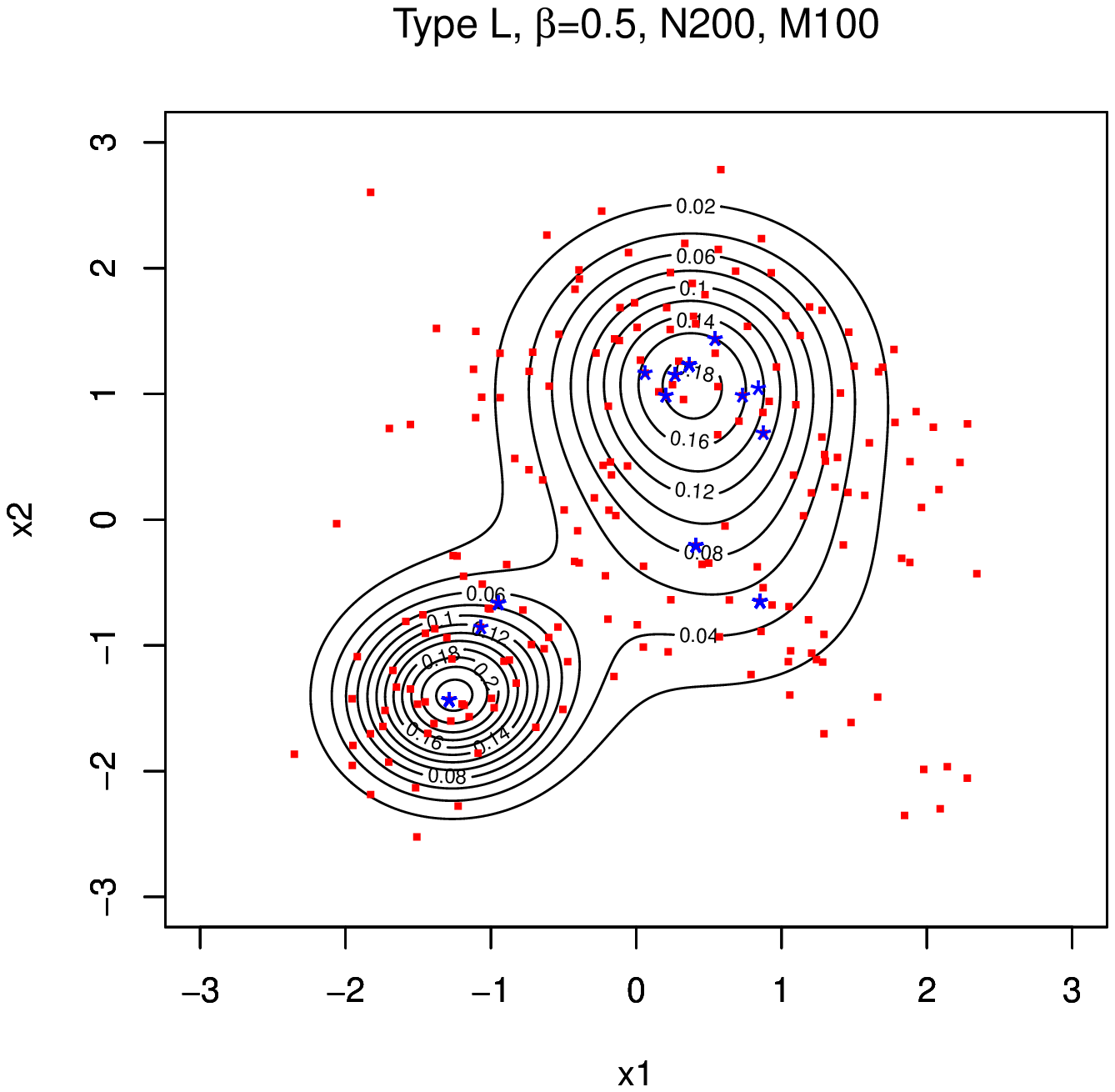}
\includegraphics*[width=0.4\linewidth]{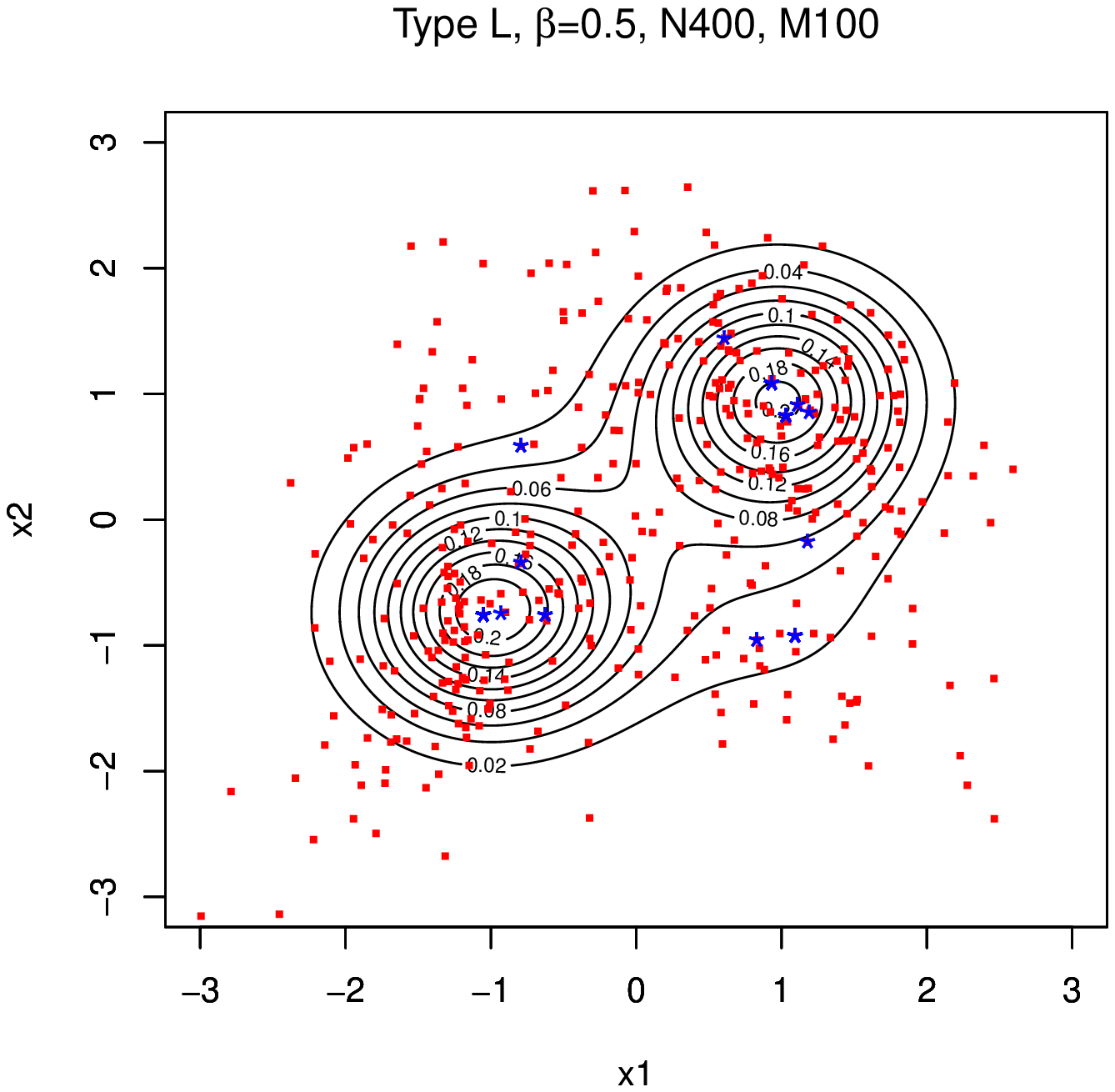}
\includegraphics*[width=0.4\linewidth]{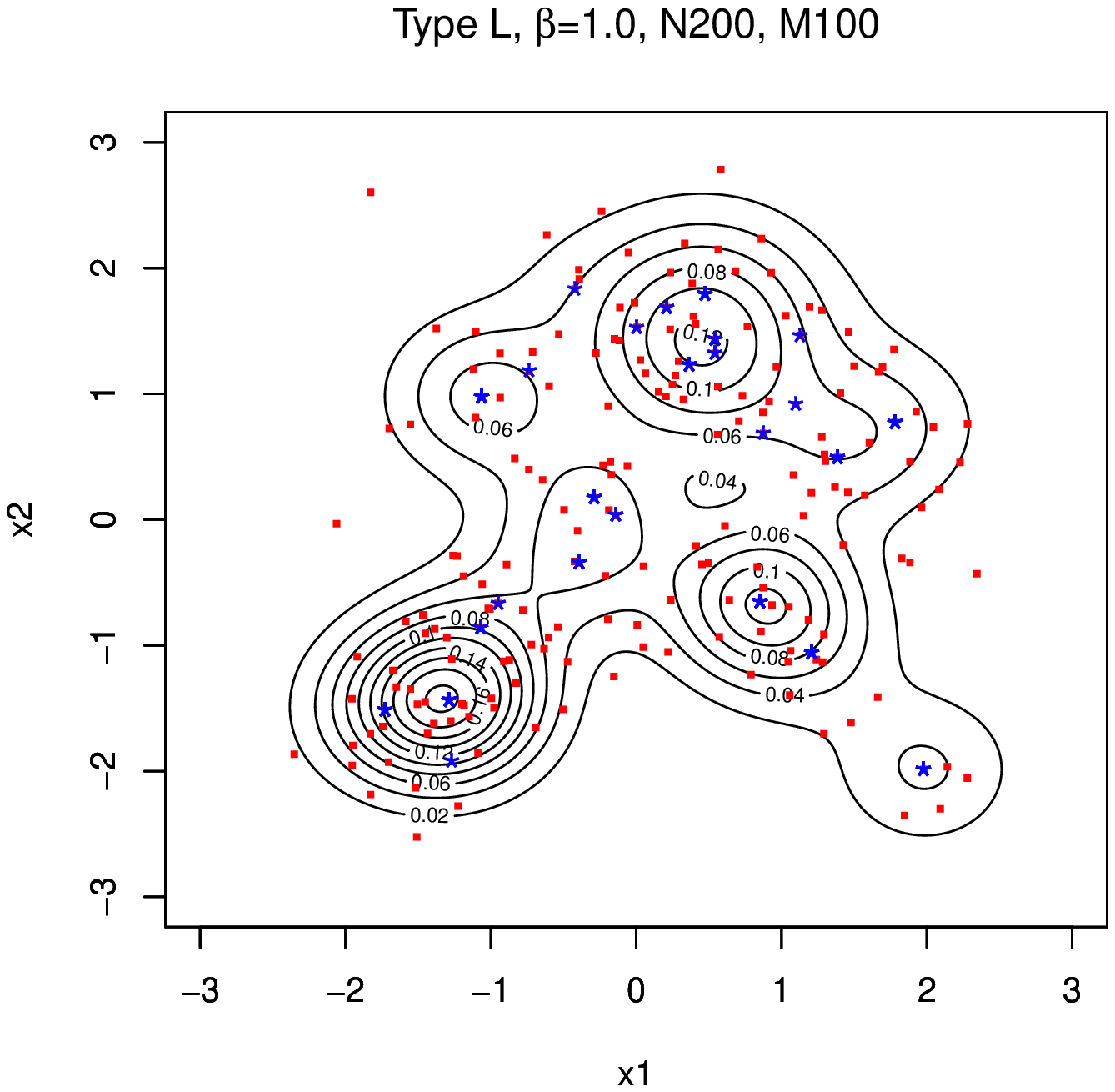}
\includegraphics*[width=0.4\linewidth]{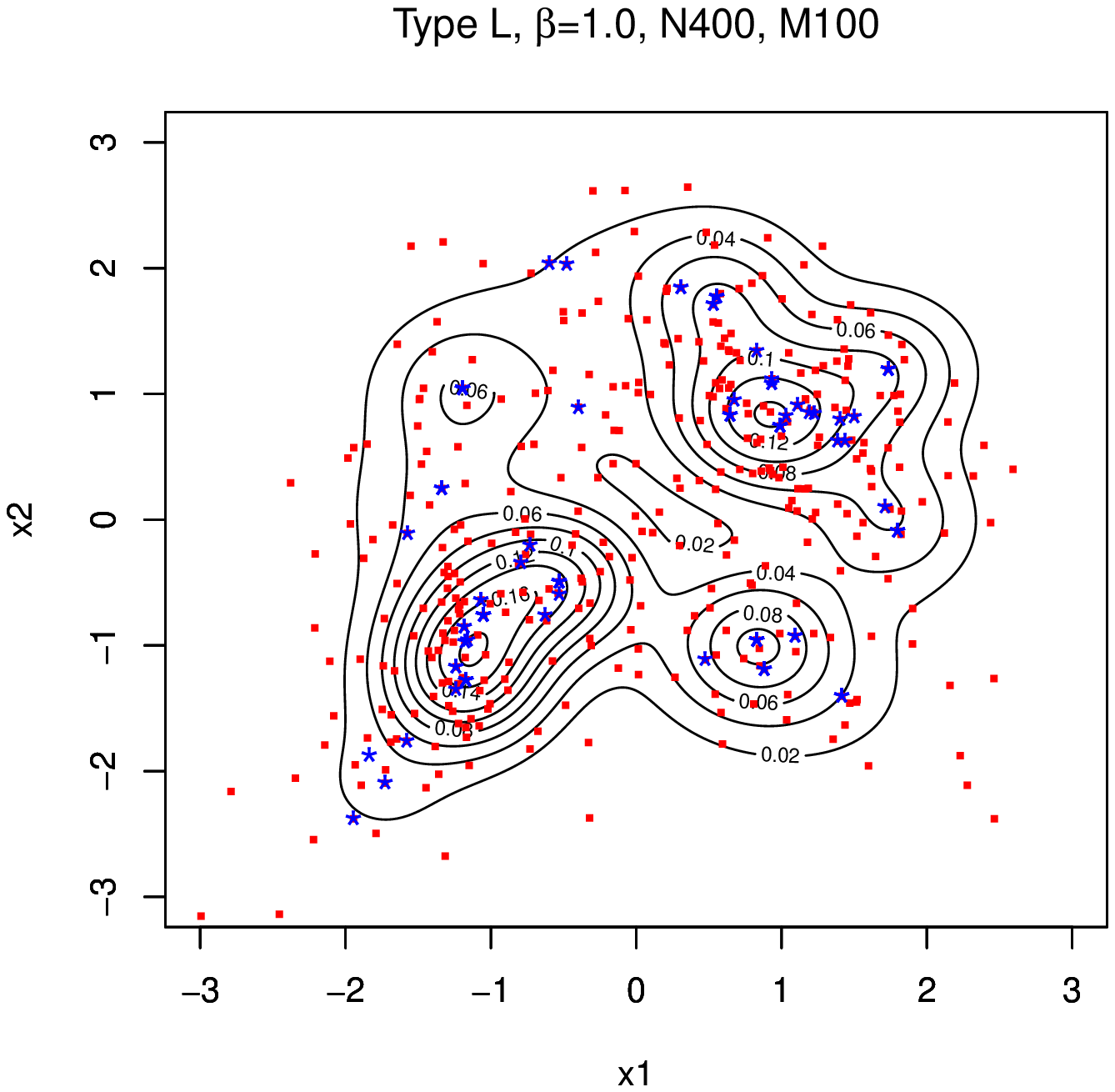}
\end{center}
\caption[]{Simulation~2: Upper: Plots of MISE vs $k$.\ \ \  Middle: Contour plots ($\beta=0.5$).\ \ \ Bottom: Contour plots ($\beta=1.0$)} \label{results.L.sim2}
\end{figure}
\clearpage
\subsection{Real data example}

We show a real data example of bivariate density estimation for our method. We use {\it{abalone data set}}, originally from Nash et al. (1994), available in the UCI Machine Learning Repository. The data set consists of the physical measurements of abalones collected in Tasmania, with each abalone being measured on eight attributes: Sex (male, female, infants), Length (mm), Diameter (mm), Height (mm), Whole weight (grams), Shucked weight (grams), Viscera weight (grams), Shell weight (grams), and Rings (integer). We choose Diameter and Viscera weight out of the eight attributes for the estimation. The sample size $N$ is $1528$ for male abalones. In the estimation, we set $m=N/2$ and $M=100$, and employ the dictionary $D_{1}$ for the cases of $\beta = 0.5$ and 1.0 each. The dictionary of bandwidths is calculated as $B_{1}=\{ h^{2} \mathbf{I}_{2}| h = 0.0294, 0.0305, 0.0320, 0.0342, 0.0384 \}$. For this data set, the DPI full bandwidth matrix is calculated as $h_{11}^{2} = 2.199 \cdot 10^{-4}$ and $h_{22}^{2} = 3.046 \cdot 10^{-4}$ for the first and the second diagonal elements respectively, and $h_{12} = 2.187 \cdot 10^{-4}$ for the non-diagonal element of the full bandwidth matrix. The univariate cross-validation bandwidth used for RSDE is calculated as $h_{cv}^{2} = 1.21 \cdot 10^{-4}$.

The values of the empirical $U$-loss at each stage of the algorithm is given in Figure~\ref{Dia.Vis.Emp.Loss}. The contour plots of the estimators $\hat{f}_{0.5}$ and $\hat{f}_{1.0}$ are given in the upper left and right panels in Figure~\ref{Dia.Vis.Contour}, respectively. The red points in the contour plots designate the data points used for the dictionary, while the blue ones are chosen by the algorithm for the estimation. The same figure also has the contour plots of the estimators, KDE with DPI full bandwidth matrix and RSDE in the lower left and right panels, respectively. From the shape of the contour plots of $\hat{f}_{0.5}$ and $\hat{f}_{1.0}$, the estimation by our algorithm appears to be working to some degree. It also seems that the shape of the contour plot $\hat{f}_{0.5}$ is more compressed vertically than that of $\hat{f}_{1.0}$. We consider this is because the case of $\beta = 0.5$ yields a robust estimation, so our algorithm chooses fewer peripheral data points in the distributional platform of $\mathbf{X}^{*}_{i}$.

We show the plots of the bandwidths selected at each stage of the algorithm in chronological order in the upper two panels in Figure~\ref{Dia.Vis.Info}, along with the frequency plots of the selected bandwidths in the lower two panels, for $\hat{f}_{0.5}$ and $\hat{f}_{1.0}$ each. The bandwidths in the dictionary are indexed in ascending order in sizes in the lower two panels of Figure~\ref{Dia.Vis.Info} for the sake of convenience. In the case of $\hat{f}_{0.5}$, we observe that the widest size of bandwidth is chosen at the initial stage. After that, the sizes of the selected bandwidths bear no relation to the progress of iteration. In the case of $\hat{f}_{1.0}$, however, it is observed that the wider sizes of bandwidths are chosen at the earlier stages and the smaller ones are chosen as the iteration progresses.

The frequency plots of the data points chosen by the algorithm are presented in Figure \ref{Dia.Vis.Freq.data.points}. The data points are indexed in the ascending order of distance from the origin for convenience. In the case of $\hat{f}_{0.5}$, 32 data points are chosen by the algorithm. The data condensation ratio is 32/1528 = 0.0209. The ratio of the actual words in $D_{1}$ chosen by the algorithm in the total number of words is 50/(5 $\times$ 764) = 0.0131. The most frequently chosen data point is $(0.5, 0.288)$, 15 times, in all. In the case of $\hat{f}_{1.0}$, however, the algorithm chooses 39 data points. The data condensation ratio is 39/1528 = 0.0255. The ratio of the number of actual words chosen by the algorithm in the total number of words is 46/(5 $\times$ 764) = 0.0120. The data point $(0.5, 0.288)$ is chosen most frequently as in the case of $\hat{f}_{0.5}$, 15 times in all. For reference, applying RSDE to this example chooses 237 data points for the estimation, yielding a larger data condensation ratio of 0.1551 than that of $\hat{f}_{0.5}$ and $\hat{f}_{1.0}$.
\begin{figure}[htpb]
\begin{center}
\includegraphics*[width=0.49\linewidth]{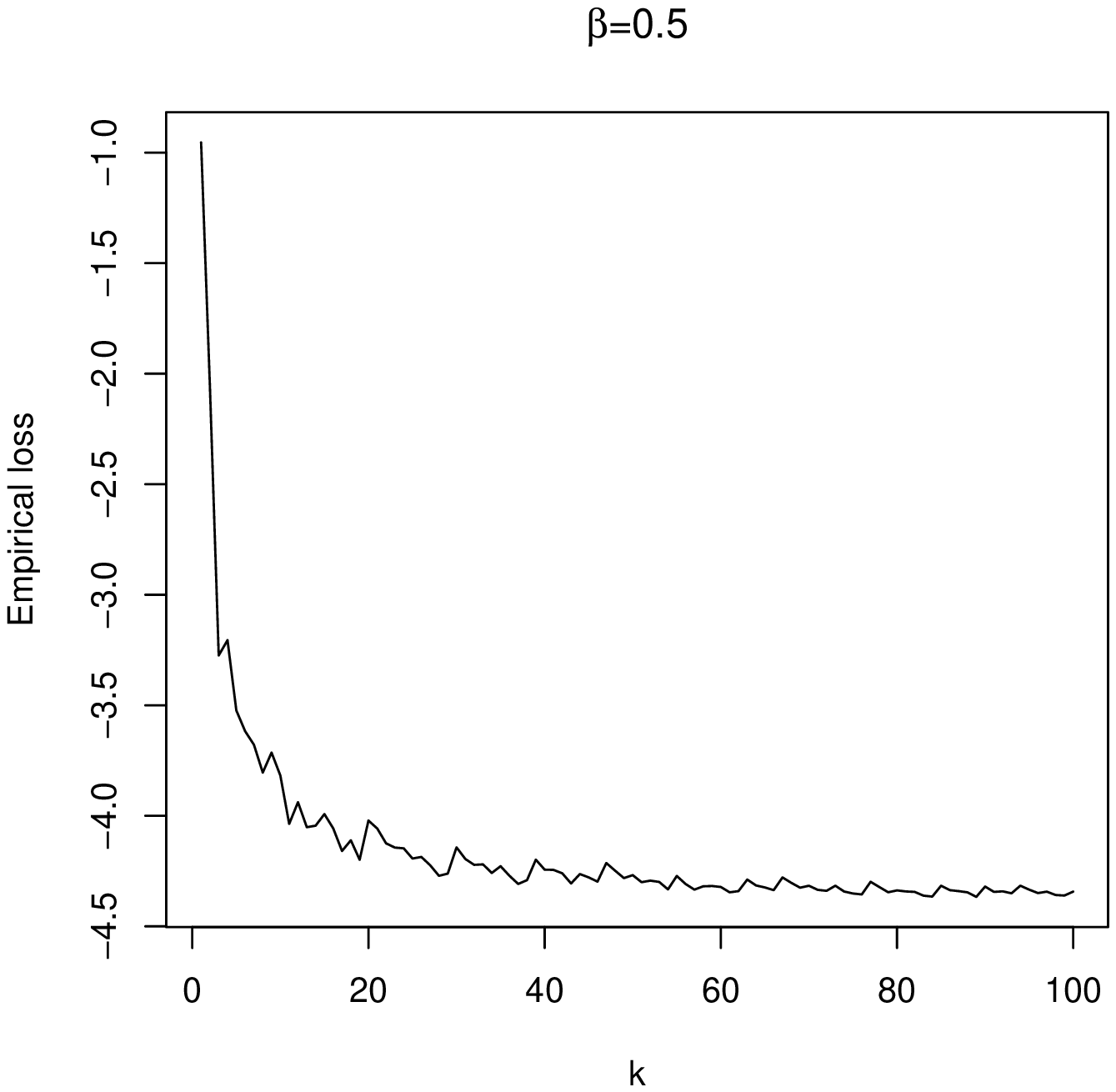}
\includegraphics*[width=0.49\linewidth]{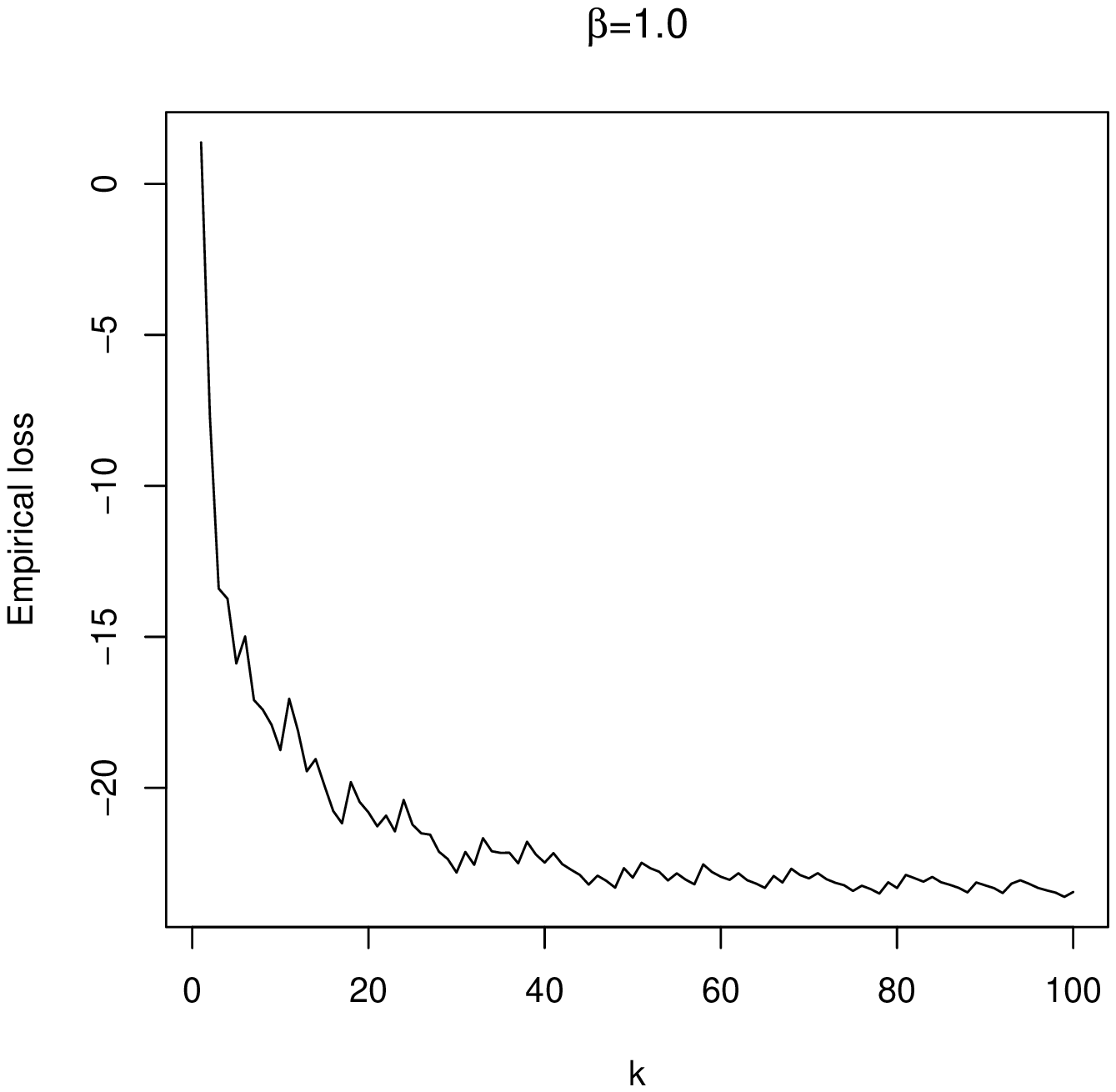}
\end{center}
\caption[]{Plots of the values of the empirical $U$-loss at each stage of the algorithm.} \label{Dia.Vis.Emp.Loss}
\end{figure}
\begin{figure}[htpb]
\begin{center}
\includegraphics*[width=0.49\linewidth]{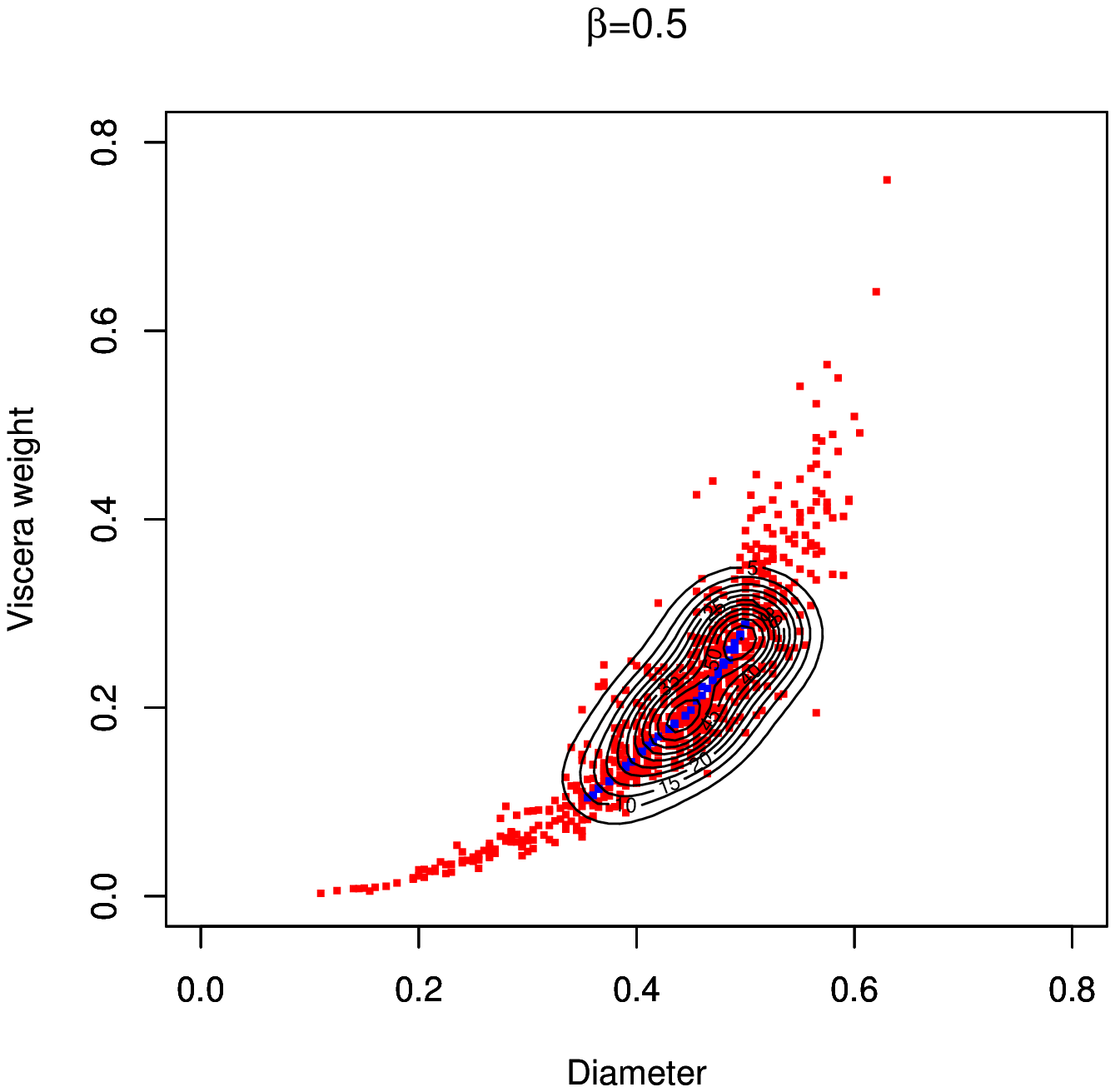}
\includegraphics*[width=0.49\linewidth]{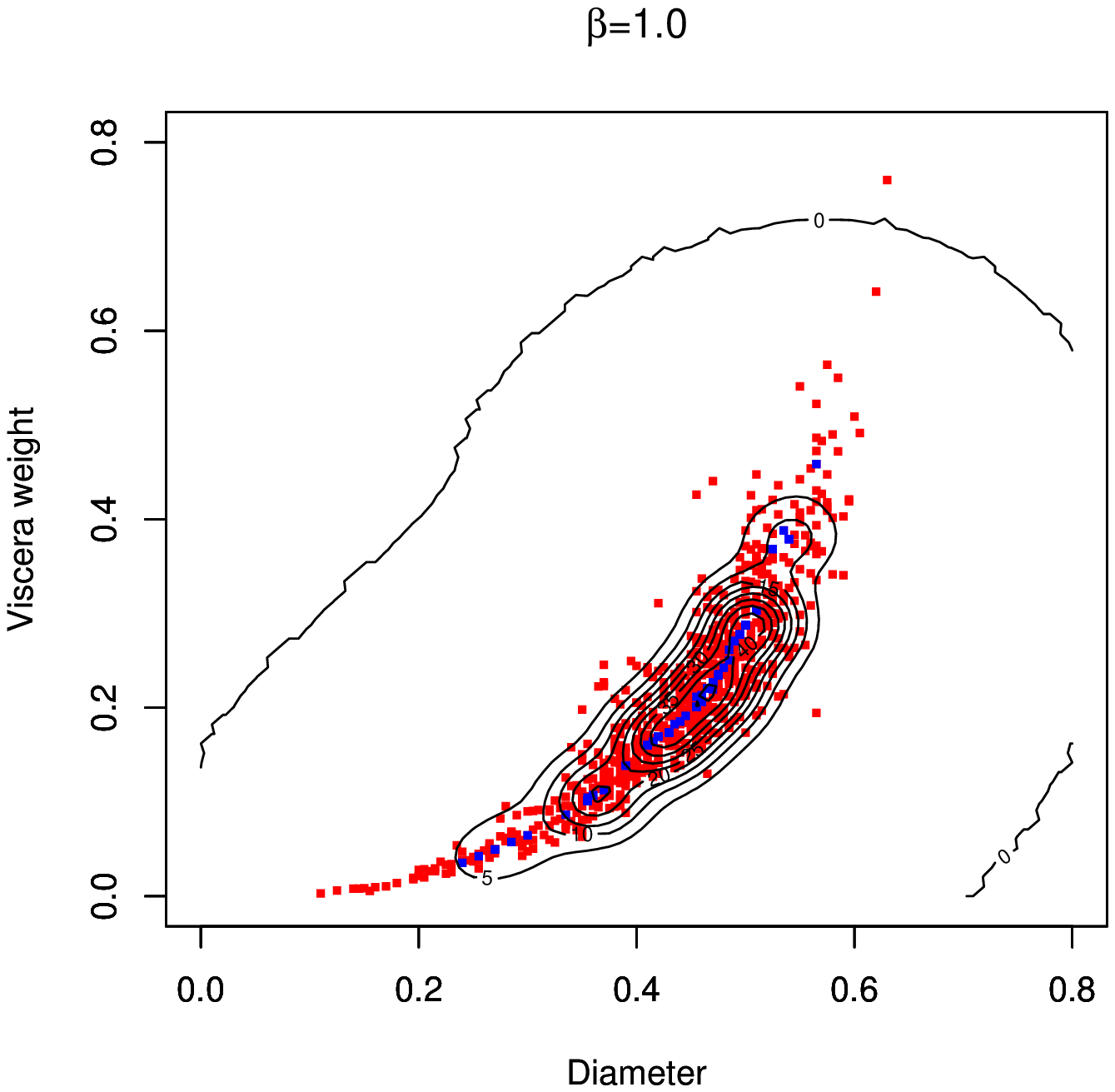}
\includegraphics*[width=0.49\linewidth]{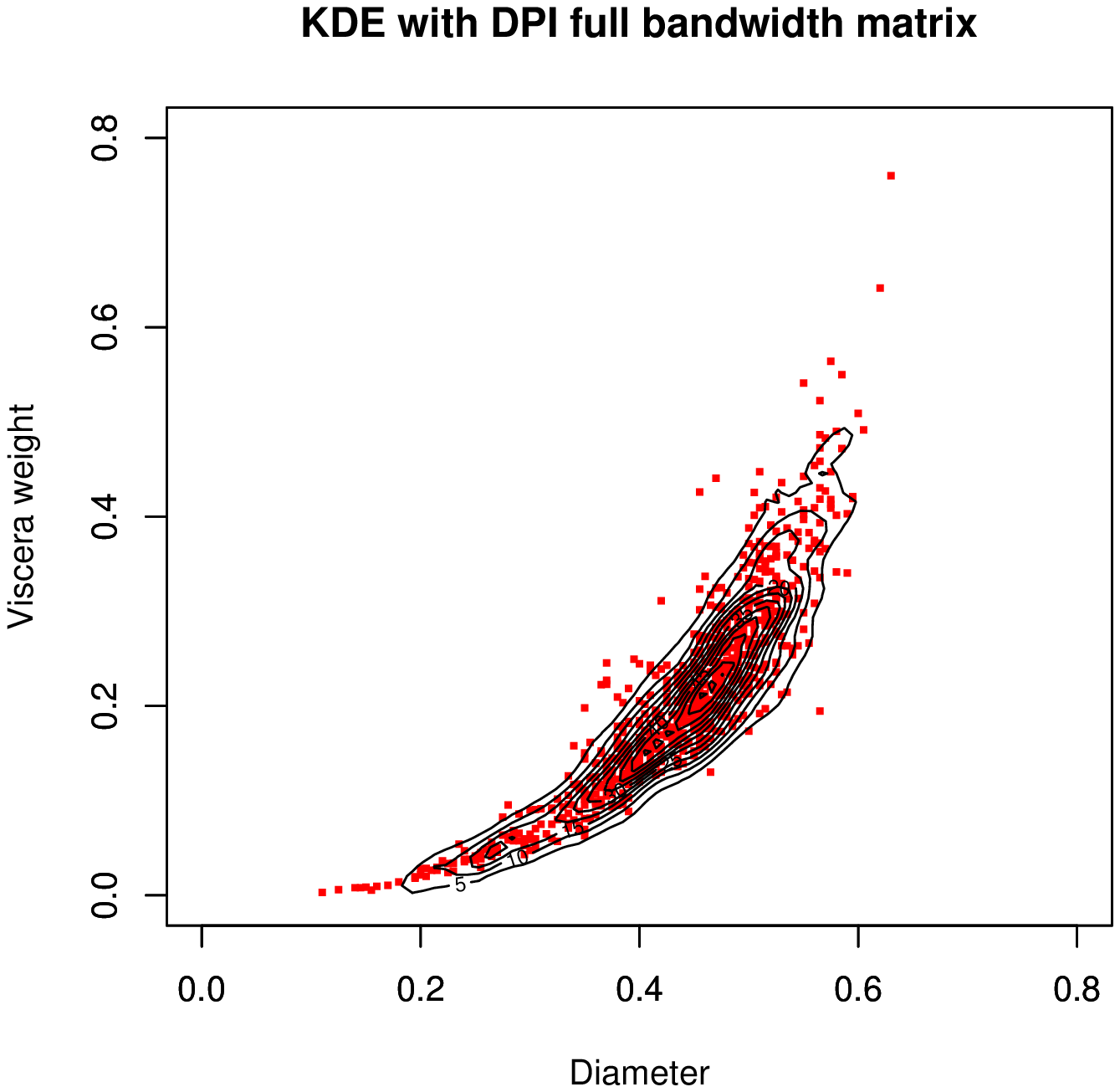}
\includegraphics*[width=0.49\linewidth]{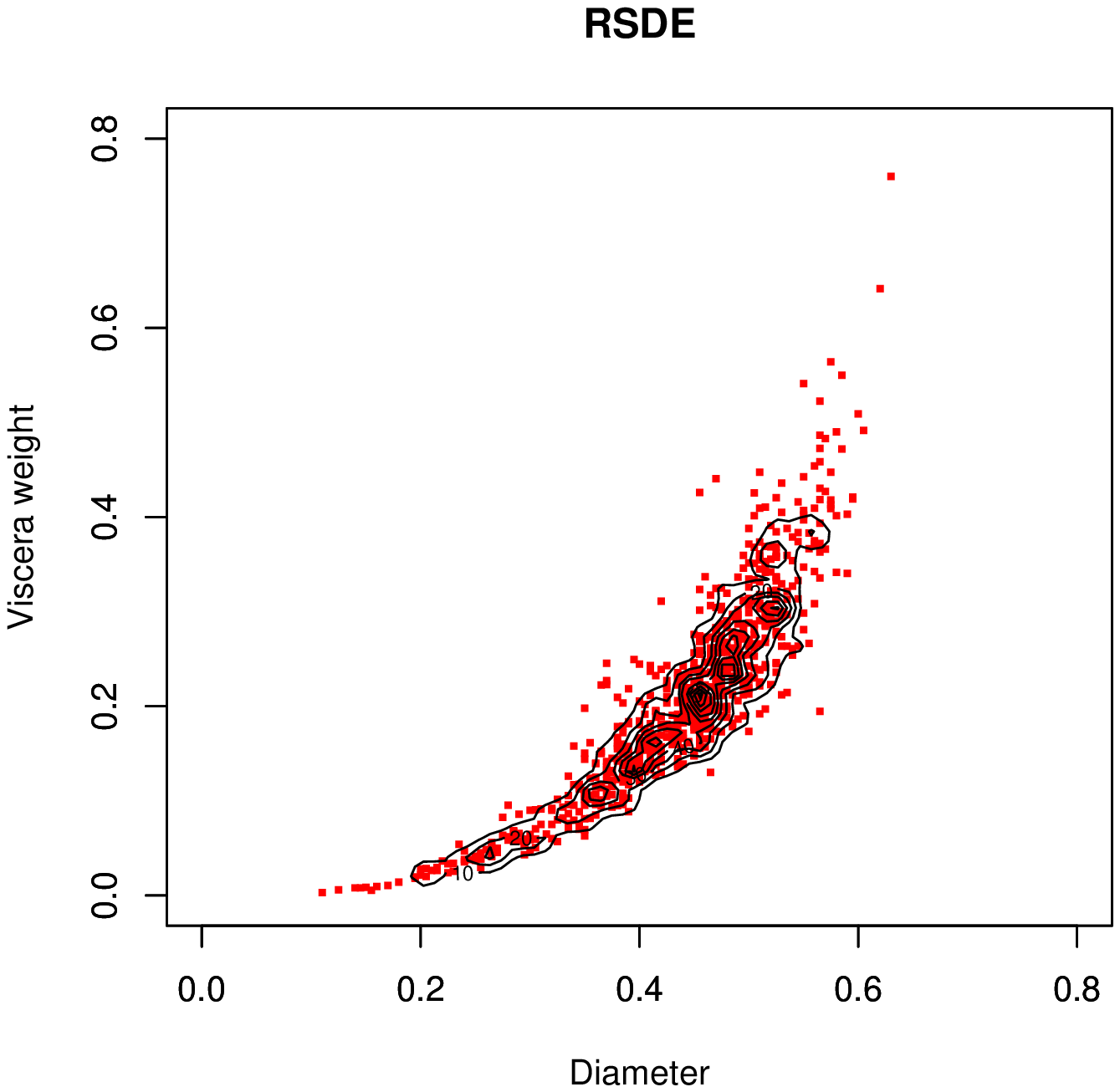}
\end{center}
\caption[]{Contour plots of the bivariate density estimators with the horizontal and vertical axes being the diameter and viscera weights in {\it{Abalone data set}} respectively. Upper left: $\hat{f}_{0.5}$. Upper right: $\hat{f}_{1.0}$. Lower left: KDE with DPI full bandwidth matrix. Lower right: RSDE.} \label{Dia.Vis.Contour}
\end{figure}
\begin{figure}[htpb]
\begin{center}
\includegraphics*[width=0.49\linewidth]{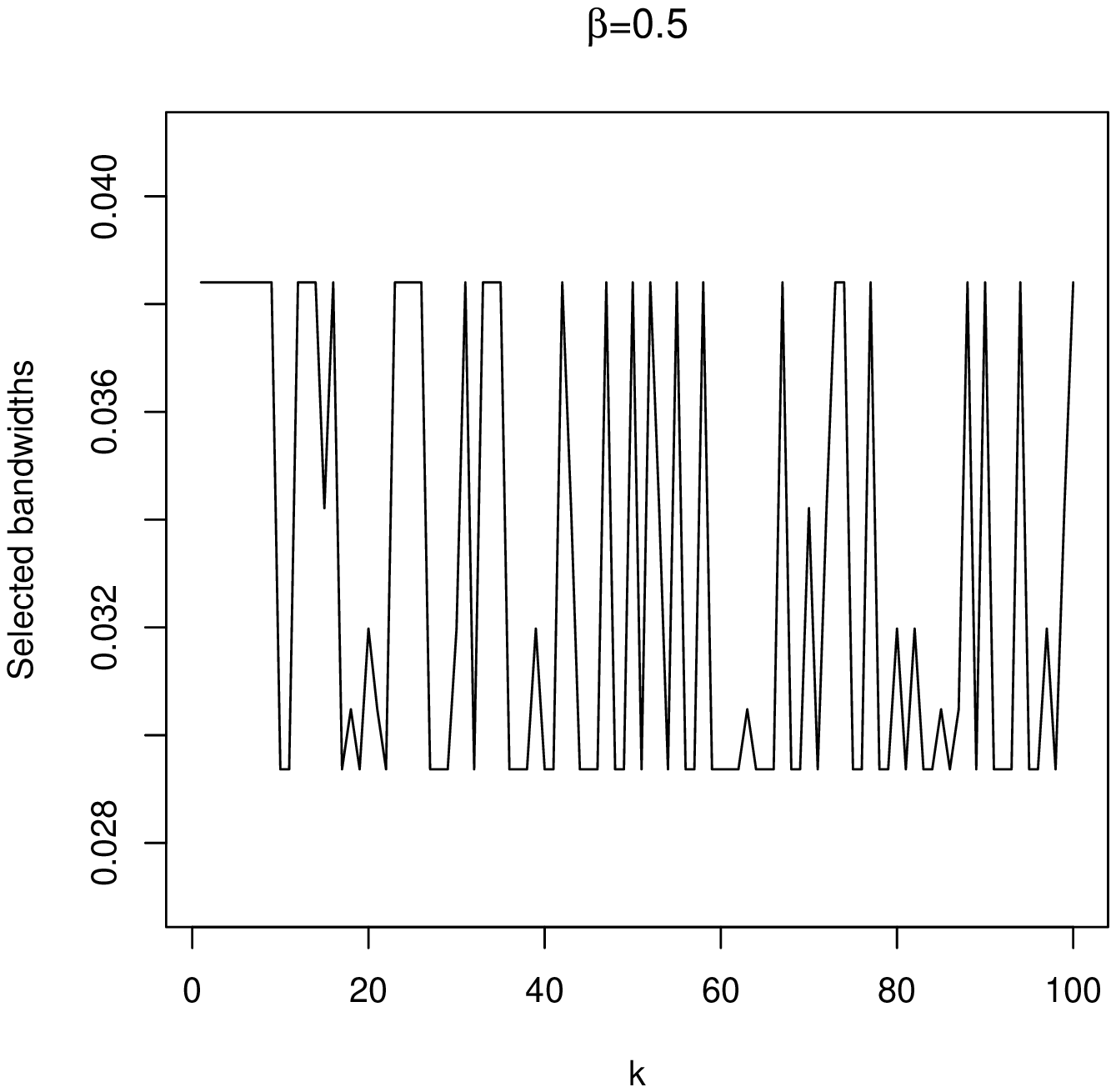}
\includegraphics*[width=0.49\linewidth]{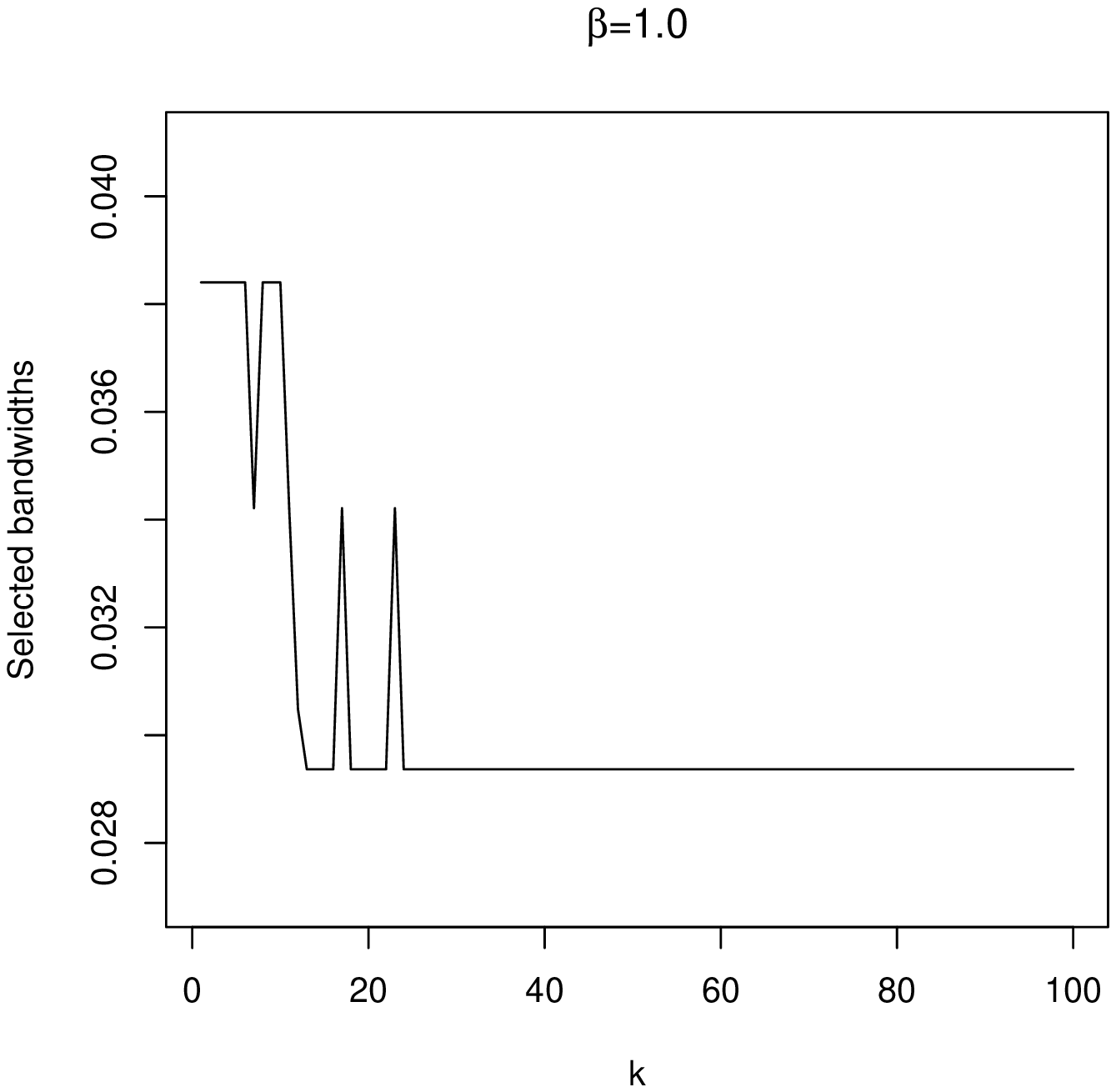}
\includegraphics*[width=0.49\linewidth]{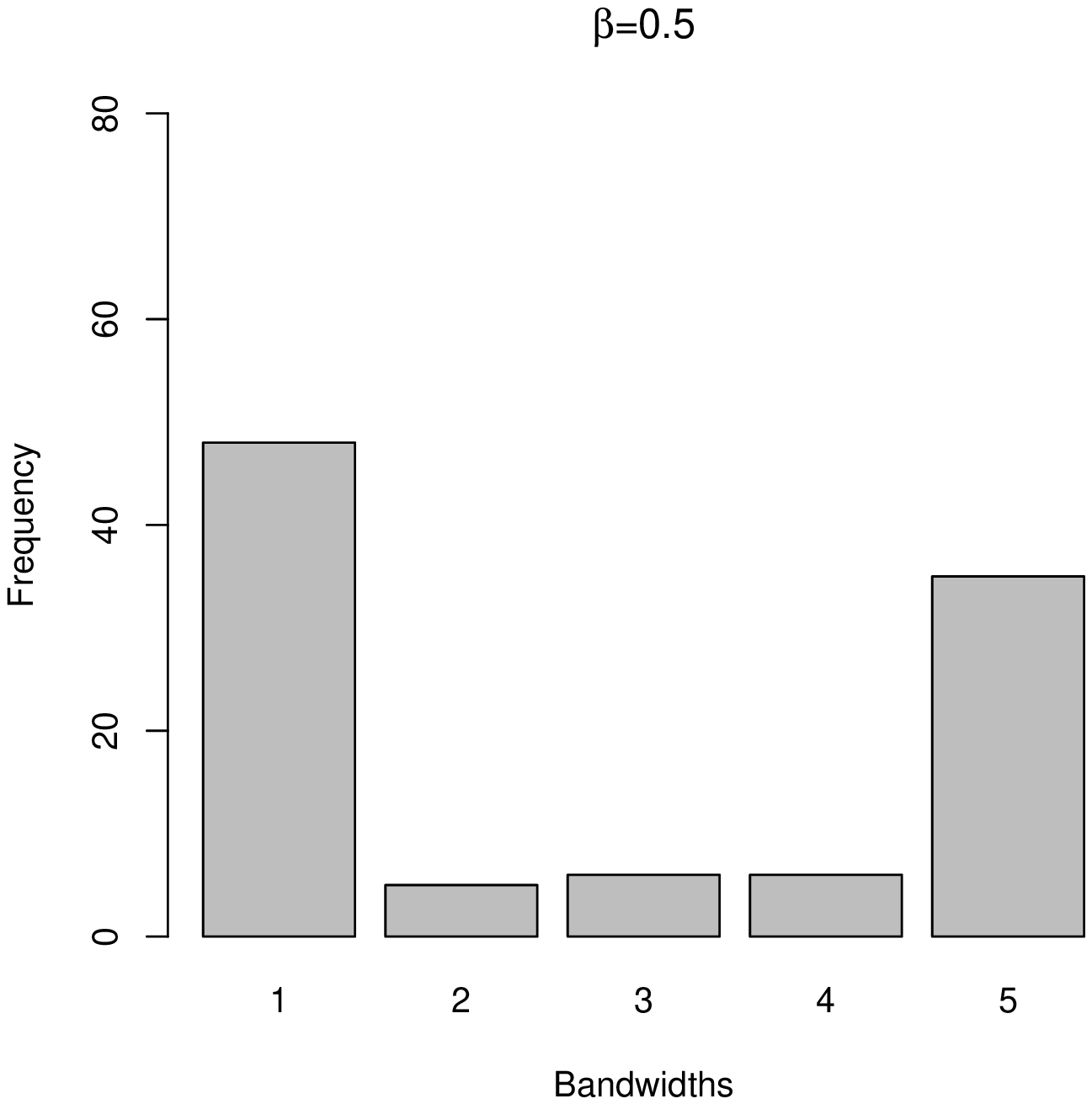}
\includegraphics*[width=0.49\linewidth]{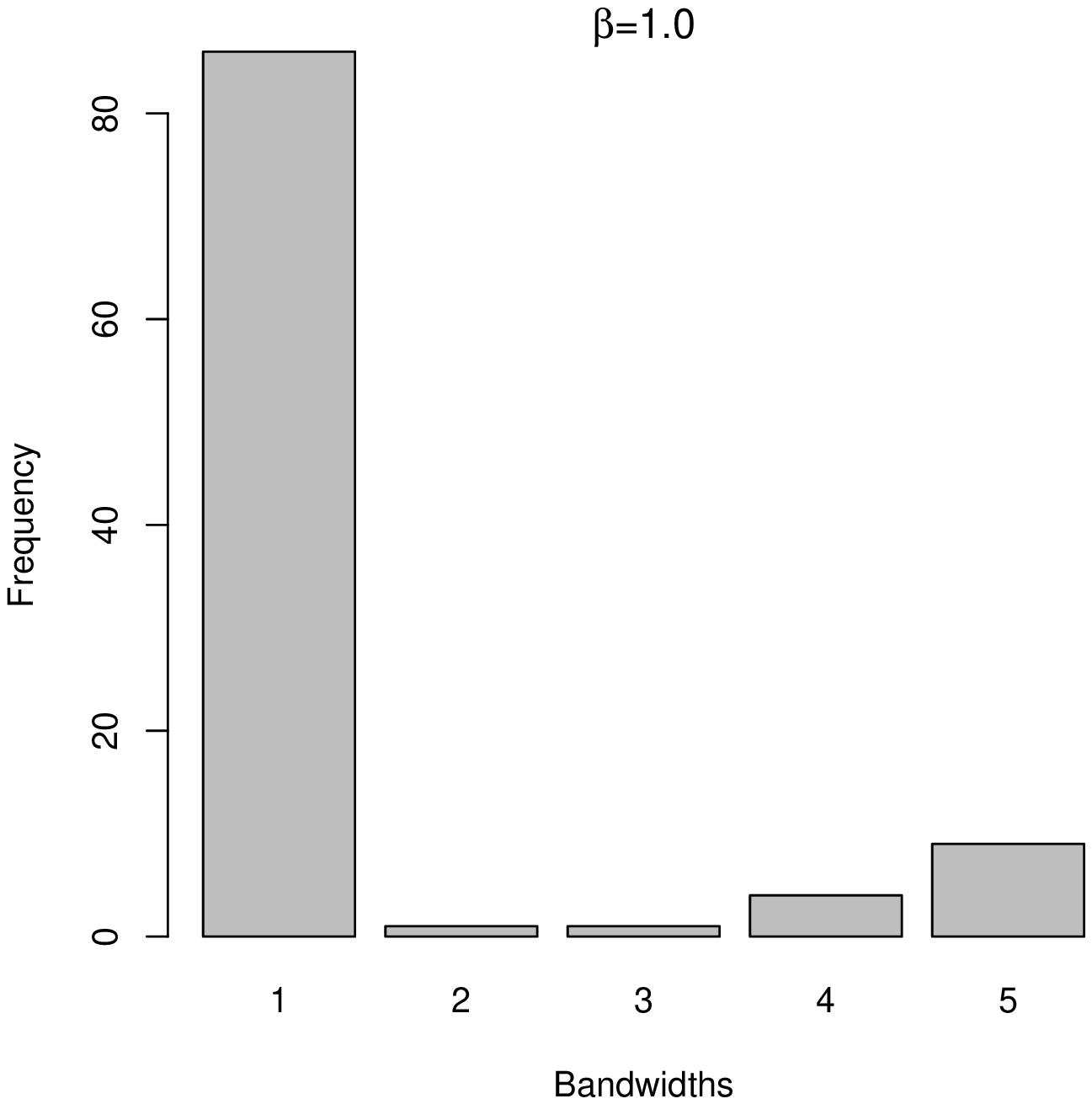}
\end{center}
\caption[]{Upper two panels: Plots of the bandwidths selected at each stage of the algorithm in chronological order. Lower two panels: Frequency of the bandwidths selected. The bandwidths in the dictionary are indexed in the ascending order of size for convenience.} \label{Dia.Vis.Freq.bandwidths.} \label{Dia.Vis.Info}
\end{figure}
\begin{figure}[htpb]
\begin{center}
\includegraphics*[width=0.49\linewidth]{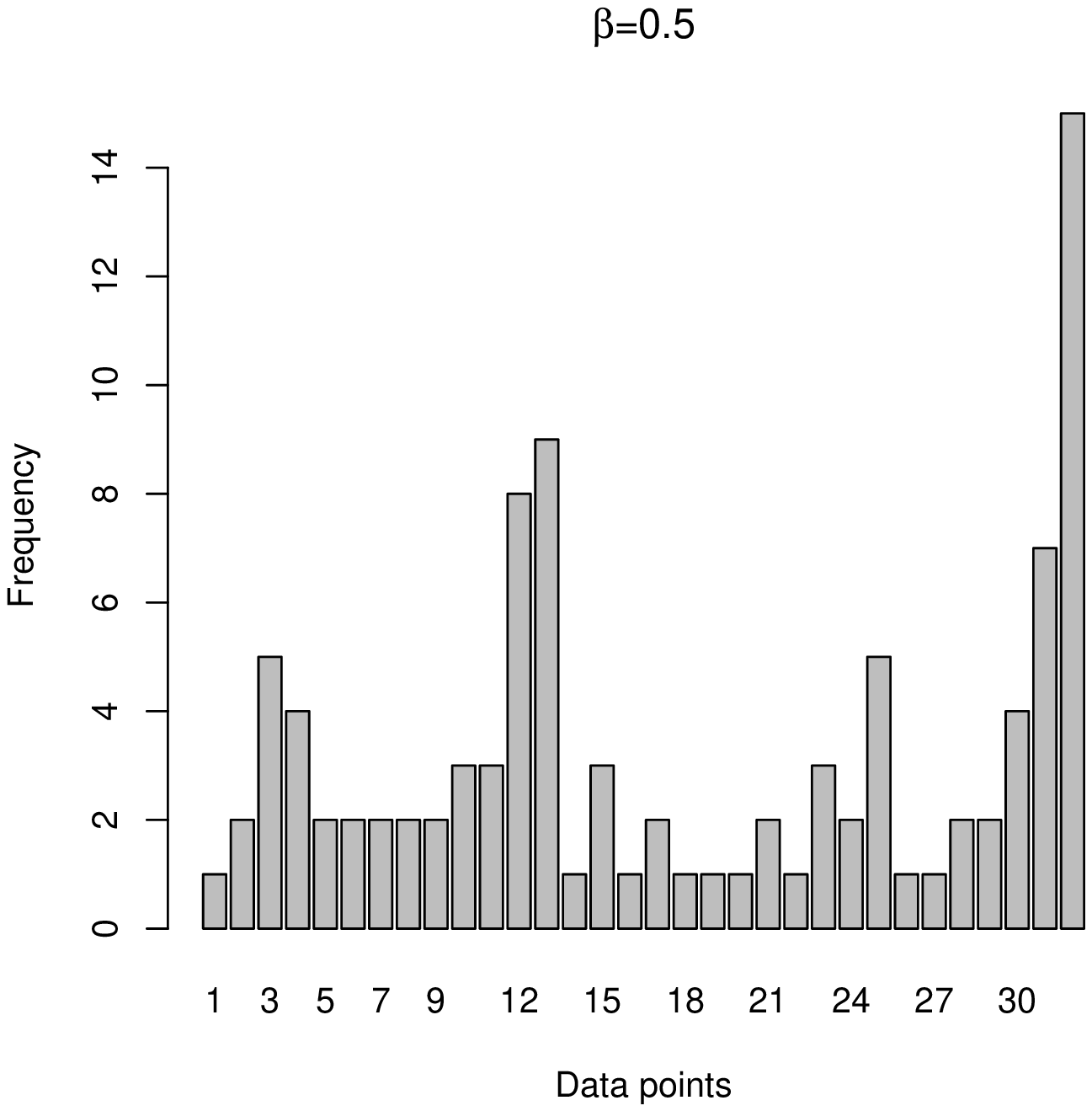}
\includegraphics*[width=0.49\linewidth]{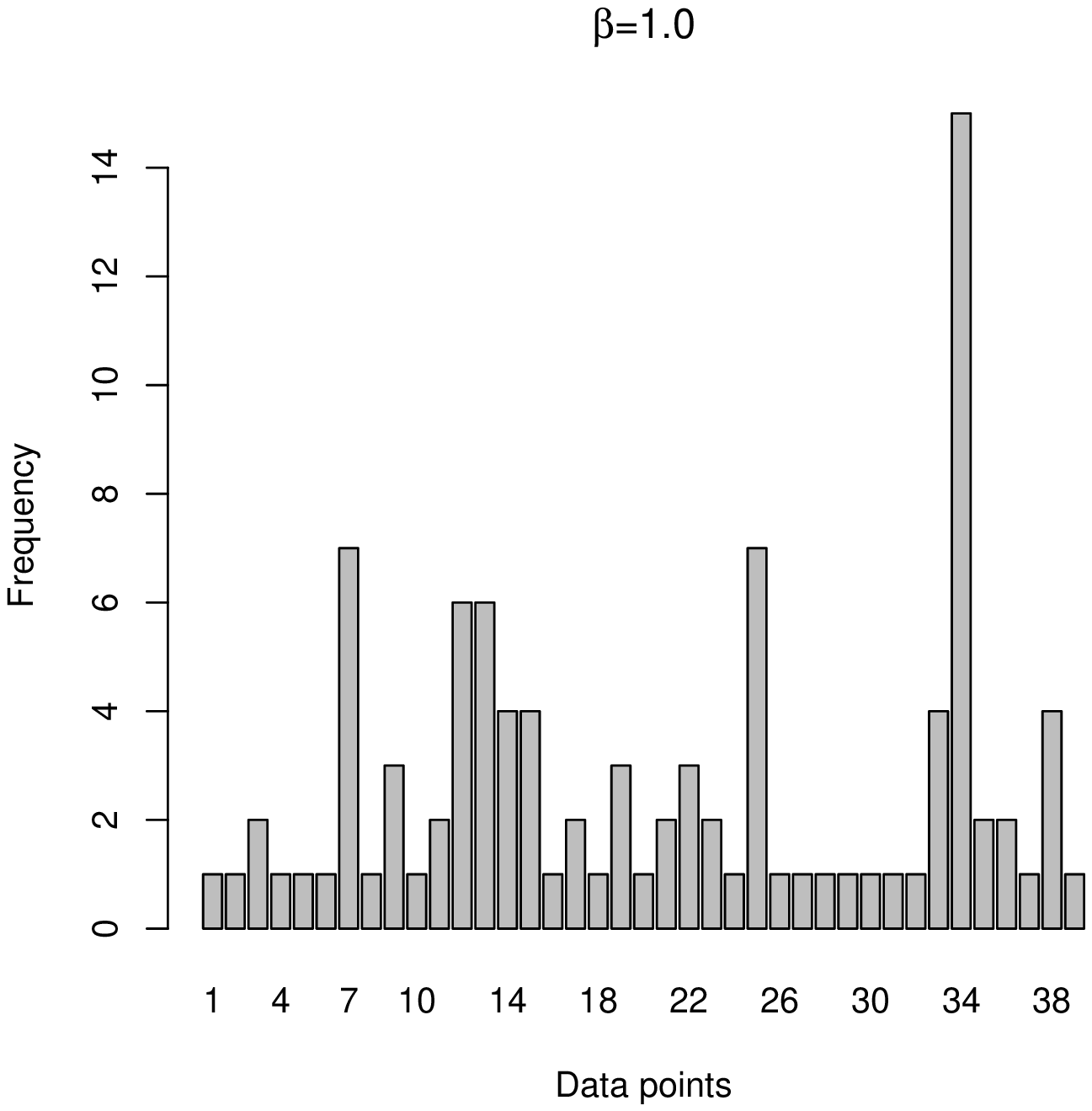}
\end{center}
\caption[]{Frequency plots of the data points selected by the algorithm. The data points are indexed in the ascending order of distance from the origin for convenience.} \label{Dia.Vis.Freq.data.points}
\end{figure}
\clearpage
\section{Discussion} \label{Discussion}
In this study, we propose an algorithm-based method for kernel density estimation. In the proposed algorithm, we first make a dictionary that consists of kernel functions with scalar bandwidth matrices. To make the dictionary, we randomly split an i.i.d. sample into two disjoint sets; we use the one for the means of the kernels and calculating the bandwidths in the dictionary and the other for calculating evaluation criterion. In calculating the bandwidths in the dictionary, we arrange the DPI diagonal bandwidth estimator as one example. Subsequently, the algorithm proceeds in a stagewise manner, choosing a new kernel from the dictionary at each stage to minimize evaluation criterion of the convex combination of the new kernel and the estimator obtained in the previous stage. For the evaluation criterion, we employ $U$-divergence. We present the non-asymptotic error bounds of the estimator in Theorems~\ref{Theorem.eb} and \ref{Theorem.eb.nrm}. The error bound shrinks as the number of iterations increases.

Our method has three advantages. First, our method can yield the density estimation, making both bandwidths and weighting parameters data-adaptive, that is, KDE, not doing both of them and RSDE doing the weighting parameters. Second, the bandwidth matrices in the dictionary are simpler and require less information on data structure than the DPI full bandwidth matrix of Duong and Hazelton(2003). Third, the proposed method obtains a sparse representation of kernel density estimation just like RSDE.

We obtain five results from the simulations in Section~\ref{Applications}. First, in terms of MISE, our proposed method can sometimes outperform KDE with DPI full and diagonal bandwidth matrices in a setting of true symmetric density. It can also outperform RSDE in terms of MISE in comparatively more cases than KDE. Second, our proposed method yields a lower data condensation ratio than RSDE. Third, although the proportion of dictionary data points in the total sample size influences the performance of the estimation, allocating fewer data points to the dictionary can improve MISE. Four, the algorithm tends to choose data points along the mountain ridges of the contour plots and uses them for density estimation. Fifth, the algorithm in the case of $0 < \beta < 1$, characterized as the mode of robust estimation, chooses fewer peripheral data points in the distributional platform, and influences the results.
\section*{Acknowledgments}
The second author gratefully acknowledges the financial support from KAKENHI 19K11851.
\clearpage
\section*{Appendix A} \label{AppendixA}

To prove Theorem~\ref{Theorem.eb}, we need the following Lemmas~\ref{L1}-\ref{L5}. Let $\tilde{f}(\mathbf{x})$ be any density estimator and let
\begin{eqnarray} \label{f_star}
f^{*}(\mathbf{x}|\mathbf{X}^{*}) = u\left( \sum_{j=1}^{N}p_{j}\xi(\phi_{j}(\mathbf{x}|\mathbf{X}^{*})) \right),
\end{eqnarray}
where $p_{j} \ge 0$, $\sum_{j=1}^{N} p_{j} = 1$, and $\phi_{j}(\mathbf{x}|\mathbf{X}^{*}) \in D$.
We consider two functions of the variable $\pi \in [0, 1]$ defined as follows:
\begin{eqnarray} \label{theta}
\lefteqn{\theta(\pi| \tilde{f}(\mathbf{x}), f^{*}(\mathbf{x}|\mathbf{X}^{*}))} \nonumber \\
&=& \sum_{j=1}^{N} p_{j} \int_{\mathbb{R}^{d}}[ U((1-\pi)\xi(\tilde{f}(\mathbf{x})) + \pi \xi(\phi_{j}(\mathbf{x}|\mathbf{X}^{*}))) - (1-\pi)U(\xi(\tilde{f}(\mathbf{x}))) - \pi U(\xi(f^{*}(\mathbf{x}|\mathbf{X}^{*}))) ]d\mathbf{x}. \nonumber
\end{eqnarray}
\begin{eqnarray} \label{eta}
\lefteqn{\eta(\pi| \tilde{f}(\mathbf{x}), f^{*}(\mathbf{x}|\mathbf{X}^{*}))} \nonumber \\
&=& \sum_{j=1}^{N} p_{j} \int_{\mathbb{R}^{d}}[ U((1-\pi)\xi(\tilde{f}(\mathbf{x})) + \pi \xi(\phi_{j}(\mathbf{x}|\mathbf{X}^{*}))) - U((1-\pi) \xi(\tilde{f}(\mathbf{x})) + \pi \xi(f^{*}(\mathbf{x}|\mathbf{X}^{*}))) ]d\mathbf{x}. \nonumber
\end{eqnarray}
Then, Lemmas~\ref{L1}-\ref{L3} are obtained as follows:\\
{\Lemma{\hspace{-2.0mm}{\bf{.}} \hspace{-2.0mm} Let $\tilde{f}(\mathbf{x})$ be a density estimator of $f(\mathbf{x})$ and let $f^{*}(\mathbf{x}|\mathbf{X}^{*})$ be the estimator given in \eqref{f_star}. For $0 \le \pi \le 1$, it then follows that
\begin{eqnarray}
\lefteqn{\sum_{j=1}^{N}p_{j} \widehat{L}_{U}(u((1-\pi)\xi(\tilde{f}(\cdot))+\pi \xi(\phi_{j}(\cdot|\mathbf{X}^{*})))) - \widehat{L}_{U}(f^{*}(\cdot|\mathbf{X}^{*}))} \nonumber \\
&=& (1-\pi)[\widehat{L}_{U}(\tilde{f}(\cdot)) - \widehat{L}_{U}(f^{*}(\cdot|\mathbf{X}^{*}))] + \theta(\pi|\tilde{f}(\mathbf{x}), f^{*}(\mathbf{x}|\mathbf{X}^{*})). \nonumber
\end{eqnarray}}\label{L1}}
{\Lemma{\hspace{-2.0mm}{\bf{.}} \hspace{-2.0mm} Let $\tilde{f}(\mathbf{x})$ be a density estimator of $f(\mathbf{x})$ and let $f^{*}(\mathbf{x}|\mathbf{X}^{*})$ be the estimator given in \eqref{f_star}. Then for any $\pi \in [0, 1]$,
\begin{eqnarray}
\theta(\pi|\tilde{f}(\mathbf{x}), f^{*}(\mathbf{x}|\mathbf{X}^{*})) &\le& \eta(\pi|\tilde{f}(\mathbf{x}), f^{*}(\mathbf{x}|\mathbf{X}^{*})) \nonumber
\end{eqnarray}}\label{L2}}
{\Lemma{\hspace{-2.0mm}{\bf{.}} \hspace{-2.0mm} Let $f^{*}(\mathbf{x}|\mathbf{X}^{*})$ be as given in \eqref{f_star} and let
\begin{eqnarray} \label{f_hat}
\tilde{f}(\mathbf{x}|\mathbf{X}^{*}) &=& u \left( \sum_{l=1}^{k} q_{l} \xi(\tilde{\phi}_{l}(\mathbf{x}|\mathbf{X}^{*})) \right) \nonumber
\end{eqnarray}
for some $\sum_{l=1}^{k} q_{l} \xi(\tilde{\phi_{l}}(\cdot|\mathbf{X}^{*})) \in co(\xi(D))$. Under Assumption~\ref{Ass1}, it follows that
\begin{eqnarray}
\eta(\pi) = \eta(\pi|\tilde{f}(\mathbf{x}|\mathbf{X}^{*}), f^{*}(\mathbf{x}|\mathbf{X}^{*})) \le \pi^{2} B_{U}(\mathbf{X}^{*})^{2} \nonumber
\end{eqnarray}
for any $\pi \in [0,1]$.}\label{L3}}\\

In Lemmas~\ref{L4}-\ref{L5}, let $W = \{ (\lambda_{\phi})_{\phi \in D} | \lambda_{\phi} \ge 0, \sum_{\phi \in D}\lambda_{\phi}=1, \#\{ \lambda_{\phi} > 0 \} < \infty \}$ and we define the following convex combination:
\begin{eqnarray}
f(\mathbf{x}, \Lambda) = f(\mathbf{x}, \Lambda, D) = u \Bigl( \sum_{\phi \in D} \lambda_{\phi}\xi(\phi(\mathbf{x}|\mathbf{X}^{*})) \Bigr), \mathbf{x} \in \mathbb{R}^{d}, \nonumber
\end{eqnarray}
where $\Lambda = (\lambda_{\phi})_{\phi \in D} \in W$. Then, we obtain Lemmas \ref{L4}-\ref{L5}.
{\Lemma{\hspace{-2.0mm}{\bf{.}} \hspace{-2.0mm} For stagewise minimization density estimator $\widehat{f}(\mathbf{x}|\mathbf{X}^{*})$, it holds under Assumption~\ref{Ass1} that
\begin{eqnarray}
\widehat{L}_{U}(\widehat{f}(\cdot|\mathbf{X}^{*})) &\le& \inf_{\Lambda \in W} \widehat{L}_{U}({f}(\cdot, \Lambda)) + \frac{\theta^2 B_{U}(\mathbf{X}^{*})^{2}}{M + (\theta -1)} + \delta. \nonumber
\end{eqnarray}}\label{L4}}
{\Lemma{\hspace{-2.0mm}{\bf{.}} \hspace{-2.0mm} For $\tau > 0$, let $\widehat{f}(\cdot|\mathbf{X}^{*}) \in \{ f(\cdot. \Lambda) | \Lambda \in W\}$ be such that
\begin{eqnarray} \label{Lemma5.assmption}
\widehat{L}_{U}(\widehat{f}(\cdot|\mathbf{X}^{*})) \le \inf_{\Lambda \in W} \widehat{L}_{U}(\widehat{f}(\cdot, \Lambda)) + \tau. \nonumber
\end{eqnarray}
Then,
\begin{eqnarray}
\lefteqn{D_{U}(f, \widehat{f}(\cdot|\mathbf{X}^{*})) \le \inf_{\Lambda \in W} \widehat{L}_{U}(\widehat{f}(\cdot, \Lambda)) + \tau + 2 \sup_{\phi \in D} |\nu_{n}(\xi(\phi(\cdot|\mathbf{X}^{*})))|,} \nonumber \\
&where& \nu_{n}(\xi(\phi(\cdot|\mathbf{X}^{*}))) = \frac{1}{n} \sum_{i=1}^{n} \xi(\phi(\mathbf{X}_{i}|\mathbf{X}^{*})) - \int_{\mathbb{R}^{d}} \xi(\phi(\mathbf{x}|\mathbf{X}^{*}))f(\mathbf{x})d\mathbf{x}. \nonumber
\end{eqnarray}}\label{L5}}
\\
If we replace the dictionary in the proofs of Klemel\"{a} (2007) and Naito and Eguchi(2013) with the one in \eqref{Dictionary}, we obtain Lemmas~\ref{L1}-\ref{L5}. Using Lemmas~\ref{L4}-\ref{L5} in conjunction with Lemmas~\ref{L1}-\ref{L3}, we obtain the non-asymptotic error bound given by Theorem~\ref{Theorem.eb}.
\section*{Appendix~B} \label{AppendixB}

We obtain Theorem~\ref{Theorem.eb.nrm} using Lemma~\ref{L6}.
{\Lemma{\hspace{-2.0mm}{\bf{.}} For any $h(\mathbf{x}|\mathbf{X}^{*}) \in co(\xi(D))$, let $g(\mathbf{x}|\mathbf{X}^{*}) = u(h(\mathbf{x}|\mathbf{X}^{*}))$ and let $g_{c}(\mathbf{x}|\mathbf{X}^{*}) = v_{g}^{-1}g(\mathbf{x}|\mathbf{X}^{*})$, where
\begin{eqnarray}
v_{g} = v_{g}(\mathbf{X}^{*}) = \int_{\mathbb{R}^{d}}g(\mathbf{x}|\mathbf{X}^{*})d\mathbf{x}. \nonumber
\end{eqnarray}
Under Assumption~\ref{Ass2}, we have
\begin{eqnarray}
\lefteqn{D_{U}(f(\cdot), g_{c}(\cdot|\mathbf{X}^{*}))} \nonumber \\
& & \le D_{U}(f(\cdot), g(\cdot|\mathbf{X}^{*})) + C_{U}^{-1} \bigl|1-v_{g}(\mathbf{X}^{*})^{-1} \bigr| \int_{\mathbb{R}^{d}} \bigl|g_{c}(\mathbf{x}|\mathbf{X}^{*}) -f(\mathbf{x})\bigr| g(\mathbf{x}|\mathbf{X}^{*})^{1-\alpha} d\mathbf{x}. \nonumber
\end{eqnarray}}\label{L6}}

If we replace the dictionary in the proofs of Klemel\"{a}~(2007) and Naito and Eguchi~(2013) with the one in \eqref{Dictionary}, we obtain Lemma~6.
\section*{Appendix~C} \label{AppendixC}

Using convex property of $U(t) = \exp(t)$, we can verify for the KL divergence that
\begin{eqnarray}
\lefteqn{\Psi(\delta, \Phi|\mathbf{X}^{*})} \nonumber \\
&=& \int_{\mathbb{R}^{d}} \exp \left( (1-\delta) \sum_{m=1}^{T}q_{m} \log \tilde{\phi}_{m}(\mathbf{x}|\mathbf{X}^{*}) + \delta \log \phi(\mathbf{x}|\mathbf{X}^{*}) \right)  \{ \log \phi(\mathbf{x}|\mathbf{X}^{*}) - \log \bar{\phi}(\mathbf{x}|\mathbf{X}^{*}) \}^{2} d\mathbf{x} \nonumber \\
&\le& \int_{\mathbb{R}^{d}} \left( (1-\delta) \sum_{m=1}^{T}q_{m} \tilde{\phi}_{m}(\mathbf{x}|\mathbf{X}^{*}) + \delta \phi(\mathbf{x}|\mathbf{X}^{*}) \right)  \{ \log \phi(\mathbf{x}|\mathbf{X}^{*}) - \log \bar{\phi}(\mathbf{x}|\mathbf{X}^{*}) \}^{2} d\mathbf{x} \nonumber \\
&\le& (1-\delta) \sum_{m=1}^{T}q_{m} \int_{\mathbb{R}^{d}} \tilde{\phi}(\mathbf{x}|\mathbf{X}^{*}) \{ \log \phi(\mathbf{x}|\mathbf{X}^{*}) - \log \bar{\phi}(\mathbf{x}|\mathbf{X}^{*}) \}^{2} d\mathbf{x} \nonumber \\
& & + \delta \int_{\mathbb{R}^{d}} \phi(\mathbf{x}|\mathbf{X}^{*}) \{ \log \phi(\mathbf{x}|\mathbf{X}^{*}) - \log \bar{\phi}(\mathbf{x}|\mathbf{X}^{*}) \}^{2} d\mathbf{x} \nonumber \\
&\le& (1-\delta)\sum_{m=1}^{T}q_{m}B_{KL}(\mathbf{X}^{*})^{2} + \delta B_{KL}(\mathbf{X}^{*})^{2} \nonumber \\
&=& B_{KL}(\mathbf{X}^{*})^{2}. \nonumber
\end{eqnarray}

We evaluate the constant $B_{KL}(\mathbf{X}^{*})^{2}$. Let $h_{a}$, $h_{b}$ and $h_{c}$ be three different scalar bandwidths, which are not random variables by assumption. Let $\mathbf{X}_{i}^{*}$, $\mathbf{X}_{j}^{*}$ and $\mathbf{X}_{k}^{*}$ respectively be the means of the words. In what follows, we denote the density of the $d$-dimensional multivariate normal distribution $N_{d}(\mathbf{X}_{i}^{*}, h_{a}^{2}\mathbf{I})$ to be $\phi_{a}(\mathbf{x}|\mathbf{X}_{i}^{*})$. We also denote the density of the $d$-dimensional standard normal distribution to be $\phi(\cdot)$. For notational convenience, we define $h_{ab} \equiv h_{a}/h_{b}$ and $h_{ac} \equiv h_{a}/h_{c}$. Then, we obtain
\begin{eqnarray} \label{J}
J &=& \int_{\mathbb{R}^{d}} \phi_{a}(\mathbf{x}|\mathbf{X}_{i}^{*}) \{ \log{\phi_{b}(\mathbf{x}|\mathbf{X}_{j}^{*})} - \log{\phi_{c}(\mathbf{x}|\mathbf{X}_{k}^{*})} \}^{2} d\mathbf{x} \nonumber \\
&=& \int_{\mathbb{R}^{d}} \phi_{a}(\mathbf{x}|\mathbf{X}_{i}^{*}) \Bigl\{ \frac{1}{2h_{c}^{2}} \| \mathbf{x} - \mathbf{X}_{k}^{*} \|^{2} - \frac{1}{2h_{b}^{2}}\| \mathbf{x} - \mathbf{X}_{j}^{*} \|^{2} + d\log{h_{cb}} \Bigr\}^{2} d\mathbf{x} \nonumber \\
&=& \int_{\mathbb{R}^{d}} \phi_{a}(\mathbf{x}|\mathbf{X}_{i}^{*}) \Biggl[ \frac{h_{ac}^{2} - h_{ab}^{2}}{2} \cdot \frac{\| \mathbf{x} - \mathbf{X}_{i}^{*} \|^{2}}{h_{a}^{2}} + \frac{(\mathbf{x}-\mathbf{X}_{i}^{*})^{T}}{h_{a}} \Bigl\{ h_{ac}^{2} \frac{(\mathbf{X}_{i}^{*} - \mathbf{X}_{k}^{*})}{h_{a}} - h_{ab}^{2} \frac{(\mathbf{X}_{i}^{*} - \mathbf{X}_{j}^{*})}{h_{a}} \Bigr\} \nonumber \\
& & + \frac{\| \mathbf{X}_{i}^{*} - \mathbf{X}_{k}^{*} \|^{2}}{2h_{c}^{2}} - \frac{\| \mathbf{X}_{i}^{*} - \mathbf{X}_{j}^{*} \|^{2}}{2h_{b}^{2}} + d \log h_{cb} \Biggr]^{2} d\mathbf{x} \nonumber \\
&=&  \int_{\mathbb{R}^{d}} \phi(\mathbf{t}) \Bigl\{ C_{1}(a, b, c) \| \mathbf{t} \|^{2} + \mathbf{t}^{T} \mathbf{C}_{2}(a, b, c; i,j,k) + C_{3}(b, c; i,j,k) \Bigr\}^{2} d\mathbf{t}, \nonumber
\end{eqnarray}
where we obtain the last equation by change of variable $\mathbf{t} = \mathbf{x} - \mathbf{X}_{i}^{*}$ and define to be
\begin{eqnarray}
C_{1} &\equiv& C_{1}(a, b, c) = \frac{h_{ac}^{2} - h_{ab}^{2}}{2} \ \ \ \in \mathbb{R}, \nonumber \\
\mathbf{C}_{2} &\equiv& \mathbf{C}_{2}(a, b, c; i,j,k) = h_{ac}^{2} \frac{(\mathbf{X}_{i}^{*} - \mathbf{X}_{k}^{*})}{h_{a}} - h_{ab}^{2} \frac{(\mathbf{X}_{i}^{*} - \mathbf{X}_{j}^{*})}{h_{a}} \ \ \ \in \mathbb{R}^{d}, \nonumber \\
C_{3} &\equiv& {C}_{3}(b, c; i,j,k) = \frac{\| \mathbf{X}_{i}^{*} - \mathbf{X}_{k}^{*} \|^{2}}{2h_{c}^{2}} - \frac{\| \mathbf{X}_{i}^{*} - \mathbf{X}_{j}^{*} \|^{2}}{2h_{b}^{2}} + d \log h_{cb} \ \ \ \in \mathbb{R}. \nonumber
\end{eqnarray}
The symbol $\| \mathbf{x} \|$ means $(\mathbf{x}^{T}\mathbf{x})^{1/2}$. Using the fact that the odd order moments of normal distribution are zero, and Theorem 11.22 in Schott (2017, p.480), we obtain
\begin{eqnarray} \label{J}
J &=& \int_{\mathbb{R}^{d}} \phi(\mathbf{t}) \Bigl\{ C_{1}^2 \| \mathbf{t} \|^{4} + (\mathbf{C}_{2}^{T}\mathbf{t})(\mathbf{t}^{T} \mathbf{C}_{2}) + C_{3}^{2} + 2 C_{1} \|\mathbf{t} \|^{2}(\mathbf{t}^{T} \mathbf{C}_{2}) + 2C_{3}\mathbf{t}^{T}\mathbf{C}_{2} + 2C_{1}C_{3} \|\mathbf{t}\|^{2} \Bigr\} d\mathbf{t} \nonumber \\
&=& d(d+2) C_{1}^{2} + \| \mathbf{C}_{2} \|^{2} + 2d C_{1}C_{3} + C_{3}^{2} \nonumber \\
&=& 2d C_{1}^{2} + \| \mathbf{C}_{2} \|^{2} + (d C_{1} + C_{3})^{2}.
\end{eqnarray}

We define
\begin{eqnarray}
h_{R} &=& \frac{h_{max}}{h_{min}}, \nonumber
\end{eqnarray}
where $h_{min}$ and $h_{max}$ are the minimum and the maximum bandwidths in the dictionary, respectively. We also define
\begin{eqnarray}
R^{2} &=& \max_{i \neq j} \{ \| \mathbf{X}_{i}^{*} - \mathbf{X}_{j}^{*} \|^{2} \}. \nonumber
\end{eqnarray}
Then, we obtain
\begin{eqnarray} \label{C1}
|C_{1}| &=& \Biggl| \frac{h_{ac}^{2}-h_{ab}^{2}}{2} \Biggr| \nonumber \\
&\le& \frac{h_{ac}^{2}+h_{ab}^{2}}{2} \nonumber \\
&\le& \frac{h_{R}^{2}+h_{R}^{2}}{2} \nonumber \\
&=& h_{R}^{2},
\end{eqnarray}
\begin{eqnarray} \label{C2}
\| \mathbf{C}_{2}\|^{2} &=& \Biggl \| h_{ac}^{2} \frac{(\mathbf{X}_{i}^{*} - \mathbf{X}_{k}^{*})}{h_{a}} - h_{ab}^{2} \frac{(\mathbf{X}_{i}^{*} - \mathbf{X}_{j}^{*})}{h_{a}} \Biggr \|^{2} \nonumber \\
&\le& \frac{h_{ac}^{4}}{h_{a}^{2}} \| \mathbf{X}_{i}^{*} - \mathbf{X}_{k}^{*}\|^{2} + \frac{h_{ab}^{4}}{h_{a}^{2}} \|\mathbf{X}_{i}^{*} - \mathbf{X}_{j}^{*}\|^{2}  + 2 \frac{h_{ac}^{2}h_{ab}^{2}}{h_{a}^{2}} \|\mathbf{X}_{i}^{*} - \mathbf{X}_{k}^{*}\| \cdot \| \mathbf{X}_{i}^{*} - \mathbf{X}_{j}^{*} \| \nonumber \\
&\le& \frac{h_{ac}^{4}}{h_{a}^{2}}R^{2} + \frac{h_{ab}^{4}}{h_{a}^{2}}R^{2} + 2 \frac{h_{ac}^{2}h_{ab}^{2}}{h_{a}^{2}} R^{2} \nonumber \\
&=& \Biggl( \frac{h_{ac}^{2} + h_{ab}^{2}}{h_{a}} \Biggr)^{2} R^{2} \nonumber \\
&\le& 4 \Biggl( \frac{h_{R}^{2}}{h_{min}} \Biggr)^{2} R^{2},
\end{eqnarray}
and
\begin{eqnarray} \label{C3}
|C_{3}| &=& \Biggl| \frac{\|\mathbf{X}_{i}^{*} - \mathbf{X}_{k}^{*}\|^{2}}{2h_{c}^{2}} - \frac{\|\mathbf{X}_{i}^{*} - \mathbf{X}_{j}^{*}\|^{2}}{2h_{b}^{2}} + d \log h_{cb} \Biggr| \nonumber \\
&\le& \frac{1}{2h_{min}^{2}} \|\mathbf{X}_{i}^{*} - \mathbf{X}_{k}^{*}\|^{2} + \frac{1}{2h_{min}^{2}} \|\mathbf{X}_{i}^{*} - \mathbf{X}_{j}^{*}\|^{2} + d \log h_{R} \nonumber \\
&\le& \frac{1}{2h_{min}^{2}} R^{2} + \frac{1}{2h_{min}^{2}} R^{2} + d \log h_{R} \nonumber \\
&=& \frac{R^{2}}{h_{min}^{2}} + d \log h_{R}.
\end{eqnarray}
Therefore, using \eqref{C1}, \eqref{C2}, and \eqref{C3} in \eqref{J}, we obtain the upper bound
\begin{eqnarray} \label{upb.BKL}
J &\le& B_{KL}(\mathbf{X}^{*})^{2} \nonumber \\
&\le& 2d h_{R}^{4} + 4 \Biggl( \frac{h_{R}^{2}}{h_{min}} \Biggr)^{2} R^{2} + \Biggl \{ dh_{R}^{2} + \frac{R^{2}}{h_{min}^{2}} + d \log h_{R} \Biggr \}^2.
\end{eqnarray}
\hfill $\Box$\\
\section*{Appendix~D} \label{AppendixD}

We prove that the finiteness of $E_{\mathbf{X}^{*}}[B_{U}(\mathbf{X}^{*})^{2}]$ in \eqref{error.bound} is ensured if the fourth moment of $\mathbf{X}_{i}^{*}$ is assumed in the case of KL divergence as described in Remark~\ref{Rmk.3}. Considering the expectation of the right-hand side of \eqref{upb.BKL}, we need to evaluate
\begin{eqnarray} \label{fourth.moment}
\lefteqn{E_{\mathbf{X}^{*}}\Bigl[ (R^{2})^2 \Bigr]} \nonumber \\
&=& E_{\mathbf{X}^{*}} \Bigl[ \max_{i \neq j} (\| \mathbf{X}_{i}^{*} - \mathbf{X}_{j}^{*}\|^{2})^2 \Bigr] \nonumber \\
&\le& E_{\mathbf{X}^{*}} \Bigl[ \max_{i \neq j} (\| \mathbf{X}_{i}^{*} \|^{2} + \| \mathbf{X}_{j}^{*} \|^{2} + 2 |\mathbf{X}_{i}^{*T} \mathbf{X}_{j}^{*}| )^2 \Bigr] \nonumber \\
&\le& E_{\mathbf{X}^{*}} \Bigl[ (\max_{i} \| \mathbf{X}_{i}^{*} \|^{2} + \max_{j} \| \mathbf{X}_{j}^{*} \|^{2} + 2 \max_{i} \|\mathbf{X}_{i}^{*}\| \max_{j} \|\mathbf{X}_{j}^{*}\| )^2 \Bigr] \nonumber \\
&=& 16 E_{\mathbf{X}^{*}} \Bigl[\max_{i} \| \mathbf{X}_{i}^{*} \|^{4} \Bigr]. \nonumber
\end{eqnarray}
Furthermore, we obtain
\begin{eqnarray} \label{fourth.moment}
\lefteqn{E_{\mathbf{X}^{*}} \Bigl[ \max_{i} \| \mathbf{X}_{i}^{*} \|^{4} \Bigr]} \nonumber \\
&\le& \sum_{i=1}^{m} E_{\mathbf{X}^{*}} \Bigl[ \| \mathbf{X}_{i}^{*} \|^{4} \Bigr] \nonumber \\
&\le& m \cdot \max_{i} E_{\mathbf{X}^{*}} \Bigl[ \| \mathbf{X}_{i}^{*} \|^{4} \Bigr]. \nonumber
\end{eqnarray}
Hence, it suffices to assume $E_{\mathbf{X}^{*}} \Bigl[ \| \mathbf{X}_{i}^{*} \|^{4} \Bigr] < \infty$. \hfill $\Box$\\\\

\end{document}